\documentclass[botnum, fleqn]{unmeethesis}

\usepackage{mathrsfs}
\usepackage{graphicx}
\usepackage{pdfpages}
\usepackage{float}
\usepackage{listings}
\usepackage{amsmath}

\usepackage{amsfonts}
\usepackage{amssymb}
\usepackage{color}
\definecolor{codegreen}{rgb}{0,0.6,0}
\definecolor{codegray}{rgb}{0.5,0.5,0.5}
\definecolor{codepurple}{rgb}{0.58,0,0.82}
\definecolor{backcolour}{rgb}{0.95,0.95,0.92}

\lstdefinestyle{mystyle}{
	backgroundcolor=\color{backcolour},   
	commentstyle=\color{codegreen},
	keywordstyle=\color{magenta},
	numberstyle=\tiny\color{codegray},
	stringstyle=\color{codepurple},
	basicstyle=\footnotesize,
	breakatwhitespace=false,         
	breaklines=true,                 
	captionpos=b,                    
	keepspaces=true,                 
	numbers=left,                    
	numbersep=5pt,                  
	showspaces=false,                
	showstringspaces=false,
	showtabs=false,                  
	tabsize=2
}

\usepackage{algorithm}
\usepackage{algorithmic}

\begin{document}




\frontmatter

\includepdf[pages=-]{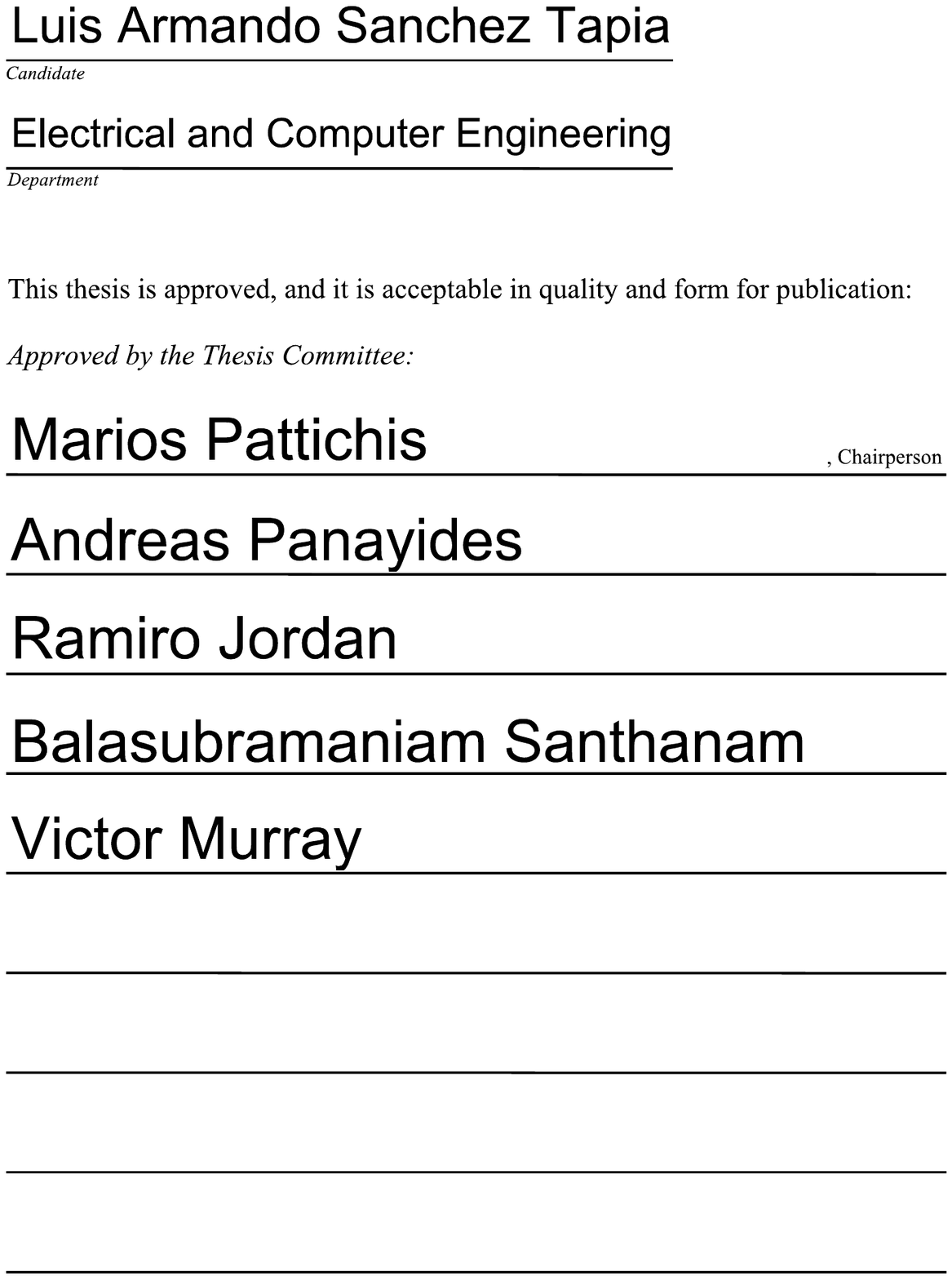}



\title{The Importance of the Instantaneous \\ Phase in Detecting Faces \\ with Convolutional Neural Networks}

\author{Luis Sanchez Tapia}

\degreesubject{M.S., Computer Engineering}

\degree{Master of Science \\ Computer Engineering}

\documenttype{Thesis}

\previousdegrees{B.S., Universidad de Ingenieria y Tecnologia, 2019}

\date{July, \thisyear}

\maketitle


\begin{dedication}
	
\setcounter{page}{3}
	To my parents, Gaby Tapia Campos and Luis Sanchez Hurtado. To Mom, without your constant support and love I would not have made it all the way here. To Dad, thanks for inspiring me and for your words in the key moments. Sol and Pancho too.  
	
	To my family in New York, Jenny, Martin, Renato and Adrianna. You make me feel just like home since I first arrived and every time I go visit you. 
	
	To my amazing godparents in Miami, Lula and Jose. You care a lot about me. I know I can count on you both for anything. This is for you. 
	
	To my grandparents. To Papa Guillermo, who introduced me to music and what engineering is about. To Mami, who showed me how much someone can love their family. 

\end{dedication}

\begin{acknowledgments}
   \vspace{1.1in}
   
   I would first like to thank my advisor Marios Pattichis for all his patience in training another hardware student. His guidance and ideas were crucial for this thesis. It is inspiring to work with him and I truly look forward to learning from him, how to conduct good research and be a good leader.
   
   I would like to thank my committee members for allowing me to present my work and share their valuable feedback. I do want to become a good PhD student.
   
   I would like to thank my amazing lab mates. They became my family far from home. Thanks to Venkatesh Jatla for addressing my doubts since I started grad school and cooking nice food. Thanks to Wenjing Shi for being such a good co-worker and better friend. Thanks to Gangadharan Esakki for all the jokes and dance, Miguel hombrados for really interesting conversations, Zhen Yu for welcoming us to her house and Phuong Tran for keeping us in shape with sports. 
   
   I would also like to acknowledge the AOLME team, for the assistantship and the opportunity of working there. I learned a lot about working with people from different disciplines.
   
   I would like to give a strong thanks to Victor Murray, who had a key role for me to become part of this school, for introducing me to his former advisor, now my advisor. Also I would like to thank the people in UTEC, for their support, long before coming to UNM. Finally, for my friends from Lima: Jose Maria Munoz, Juan Llanos, Luis Montano y Giancarlo Patino, for all of those hours of studying that brought me here. Without the friends from Lima, it would have been so much harder. 
   
   This material is based upon work supported by the National Science Foundation under Grant No. 1613637 and Grant No. CNS-1422031. Any opinions or findings of this thesis reflect the views of the author. They do not necessarily reflect the views of NSF.
   
\end{acknowledgments}

\maketitleabstract 

\begin{abstract}
	Convolutional Neural Networks (CNN) have provided new and accurate methods
	   for processing digital images and videos.
	Yet, training CNNs is extremely demanding in terms of computational resources. 
	Also, for specific applications, 
	   the standard use of transfer learning also tends to require far
	   more resources than what may be needed.
	Furthermore, the final systems tend to operate as black boxes that
	   are difficult to interpret. 
	
	The current thesis considers the problem of detecting faces from the AOLME video dataset.
	The AOLME dataset consists of a large video collection of group interactions 
	that are recorded in unconstrained classroom environments.
	For the thesis, still image frames were extracted at every minute 
	   from 18 24-minute videos.
    Then, each video frame was divided into $ 9 \times 5 $ blocks with $ 50 \times 50$ pixels each. For each of the 19440 blocks, the percentage of face pixels was set as 
    ground truth. Face detection was then defined as a regression problem for determining the 
    face pixel percentage for each block. For testing different methods, 12 videos were used for training and validation. The remaining 6 videos were used for testing.
    
	The thesis examines the impact of using the instantaneous phase for the 
	    AOLME block-based face detection application.
	For comparison, the thesis compares the use of the Frequency Modulation
	    image based on the instantaneous phase, the use of the instantaneous amplitude,
	    and the original gray scale image.
    To generate the FM and AM inputs, the thesis uses dominant component analysis
        that aims to decrease the training overhead while maintaining
        interpretability.
			
	The results indicate that the use of the FM image yielded about the same
	    performance as the MobileNet V2 architecture (AUC of 0.78 vs 0.79), 
	    with vastly reduced training times.
	Training was 7x faster for an Intel Xeon with a GTX 1080 based desktop 
	    and 11x faster on a laptop with Intel i5 with a GTX 1050.
	Furthermore, the proposed architecture trains 123x less parameters
	    than what is needed for MobileNet V2.
	The FM-based neural network architecture uses a single convolutional layer.
	In comparison, the full LeNet-5 on the same image block using the original image
	   could not be trained for 
	   face detection (AUC of 0.5).      
\clearpage 
\end{abstract}

\tableofcontents
\listoffigures
\listoftables

\mainmatter

\chapter{Introduction}
Image representations are of extreme relevance in digital image processing methods and applications. From Fourier representations to Amplitude-Modulation Frequency-Modulation (AM-FM) representations, there is a great need to effectively describe image content.
The current research aims to highlight the importance of using the dominant frequency modulation component in the application of face detection with Convolutional Neural Networks.

CNNs have been widely used in image processing applications in recent years. Their ability of learning patterns make them suitable for complex problems where there is not a defined procedure to attempt a solution. One of this hardest image processing problems is the detection of faces in unconstrained environments, in which the face image can vary significantly based on the camera imaging geometry. At the same time, the training process of modern deep learning methods requires a massive amount of data to learn from. The dataset then, is apt to be modified according to the necessities of such imaging conditions.

\section{Motivation}

The main motivation of this thesis is to explore the relevance of instantaneous phase representations of images as a key part of face detection methods using Convolutional Neural Networks. Motivated by earlier research in the importance of the phase information \cite{oppenheim} and the development of recent AM-FM representations \cite{multiscale}, the importance of the instantaneous phase in images needs to be investigated in the context of the current explosive use of CNNs.

The initial motivation of this research was to use machine learning methods to analyze videos collected from the Advancing Out-of-school Learning in Mathematics and Engineering (AOLME) \cite{aolme1}  after-school program. Despite having almost the same camera position and the same group every video, kids go around the table and the lighting conditions changes significantly. This makes this dataset suitable for studying real-life unconstrained environments as it resembles the actual hard problems machine learning methods are facing.

The thesis is focused on the development of multiscale AM-FM demodulation method that can be implemented using a small number of fixed-point digital filters. Ultimately, the goal is to support fast training for large-scale video datasets.

\section{Thesis Statement}

The thesis of this research is that the use of AM-FM techniques can provide efficient methods for deep learning methods. To validate the thesis claims, the proposed approach is compared against the use of deep learning methods on raw images.


\section{Contributions}

The contributions include:

\begin{itemize}

\item A low parameter Gabor filterbank to estimate the dominant FM component of each input image. It consist of 8-directions using a rotated ellipsoidal set of gaussians and gabor filters using  $ 11 \times 11 $ coefficients.  

\item  For implementing fixed-point designs the thesis provides a Simulated Annealing approach. The approach is demonstrated on the design of the Hilbert filter using 8-bit fixed-point arithmetic.

\item A hybrid Neural Net architecture based on both single and multi region regression to implement face detection using dominant FM components. The proposed approach allows for fast training for face detection while the original LeNet-5 architecture cannot be trained on the raw image data from the same region, and the use of modern MobileNet V2 architectures require significantly more training to give slightly improved results.

\end{itemize}

\section{Thesis Overview}

Chapter 2 provides background in basic AM-FM demodulation methods and relevant Deep-Learning architectures. Chapter 2 also includes the discussion of what was used to generate the ground truth on the AOLME dataset. Chapter 3 describes the low-parameter AM-FM demodulation approach and the low-parameter regression architecture for detecting faces in independent image blocks as well as by combining results from different blocks. Chapter 4 gives a summary of the results from the different approaches. Chapter 5 gives the conclusion and a summary of future work.

\chapter{Background}

The first section describes the source of the data and the generation of the ground truth dataset used in this research. The following section provides background in AM-FM methods relevant to the current thesis. The last section describes two deep learning (LeNet-5 and MobileNet V2). 

\section{AOLME Dataset} \label{aolmeDataset}

The Advancing Out-of-school Learning in Mathematics and Engineering (AOLME) project \cite{aolme1} is an after school program implemented by the department of Electrical and Computer Engineering and the department of Language, Literacy and Sociocultural Studies. AOLME generated a large amount of video data that includes around 2000 hours of groups interactions, monitor data and screen recordings. Each AOLME video camera recorded how a small group of middle school students interacted with each other and their facilitator. For each student, AOLME collected a video each week, up to 11 weeks, for a total of 20 hours. Video are characterized by strong illuminations variations and the students moving around within their group or to join other groups. 

A sample of still images was used for the purpose of establishing ground truth for face detection. The image samples were selected to reflect the variations in the data.

The AOLME dataset has also been used in related prior research. In \cite{cody1}, the authors used a scalable and distributed architecture in the cloud to analyze cropped videos to determine (a) writing versus non-writing and (b) typing versus non-typing. In contrast, in \cite{callie} the author worked in detecting hand movements in full videos.

One recent example of video analysis to detect human interactions can be found in \cite{abby1}. In this research, color-based segmentation was applied to identify potential regions of interest. Then, context-based rules are applied to further filter the regions of interest of the interaction. Motion vectors were extracted to determine the interaction. Then, K-nearest neighborhood classifiers and deep neural networks were used to classify the activity. 

Another example of research using the AOLME dataset is concerned with Human Attention Detection \cite{wenjingshi}. This research uses single frames to extract AM-FM features to identify regions of interest. This is followed by face detection and back of the head detection, and further processing to determine where students are looking. The methods were extended in \cite{shi2} and \cite{shi3}.

From the AOLME video collection, the group interactions videos were selected to create the dataset used in this research. Each session comprises of 4 video clips of 24 minutes each. The training set consists of 12 video clips extracted from 12 sessions. The test set consists of 6 video clips extracted from 6 different sessions.

In order to have diversity of faces, one of those video clips were taken from a total of 18 sessions. 12 videos were used to create the training set and 6 for the testing set, so there is no chance of mixing these 2 sets. Then the process is divided into generating the image data, X\_train and X\_test 3D arrays, and the overlap percentages for the y\_train and y\_test 1D arrays.

\begin{figure}	[bt]
	\includegraphics[width=\columnwidth]{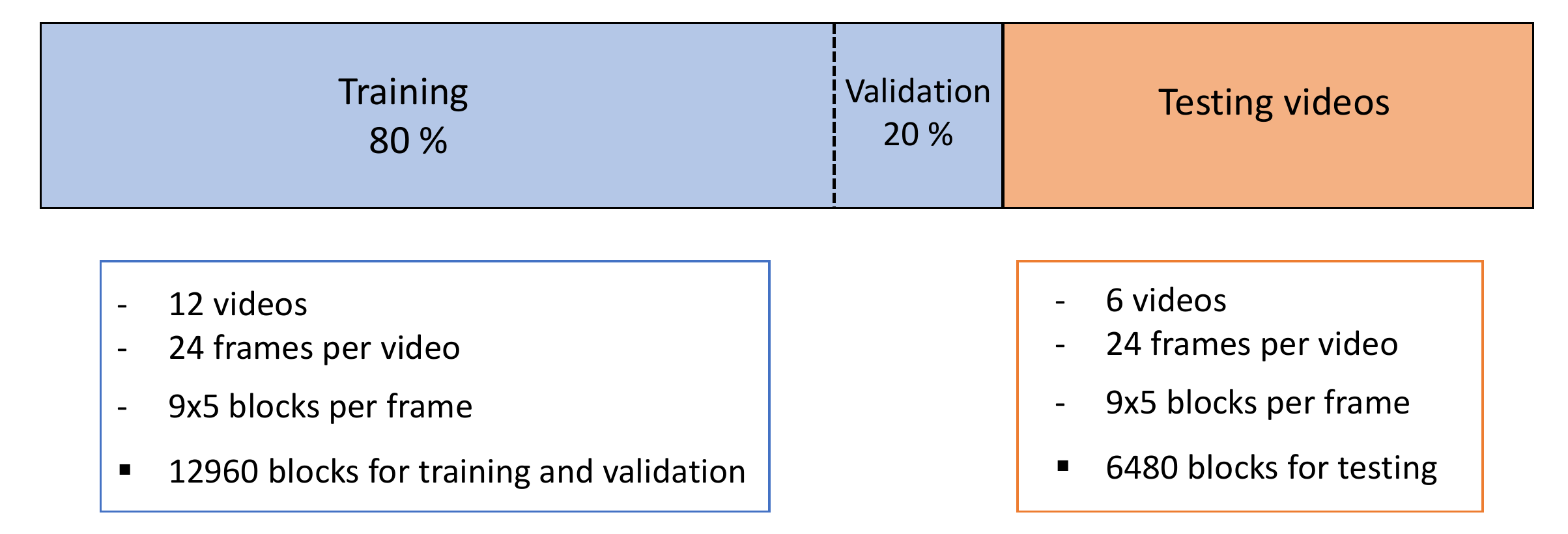}
	\caption{Training, validation, and testing video frames.}
	\label{fig:videosDataset}
\end{figure}

One frame is extracted for each minute in each video clip, resulting in 24 frames per video. After this, a decimation is used to shrink the frame to half the original resolution: $ 240 \times 429 $. Each frame is zero padded to extend the size to the nearest multiple of 5 for the rows and 9 for the columns, giving a total of 45 blocks of the same size: $ 50 \times 50 $ pixels. Figure \ref{fig:videosDataset} gives a summary of the training, validation and testing datasets. 

\begin{figure}	[bt]
	\includegraphics[width=\columnwidth]{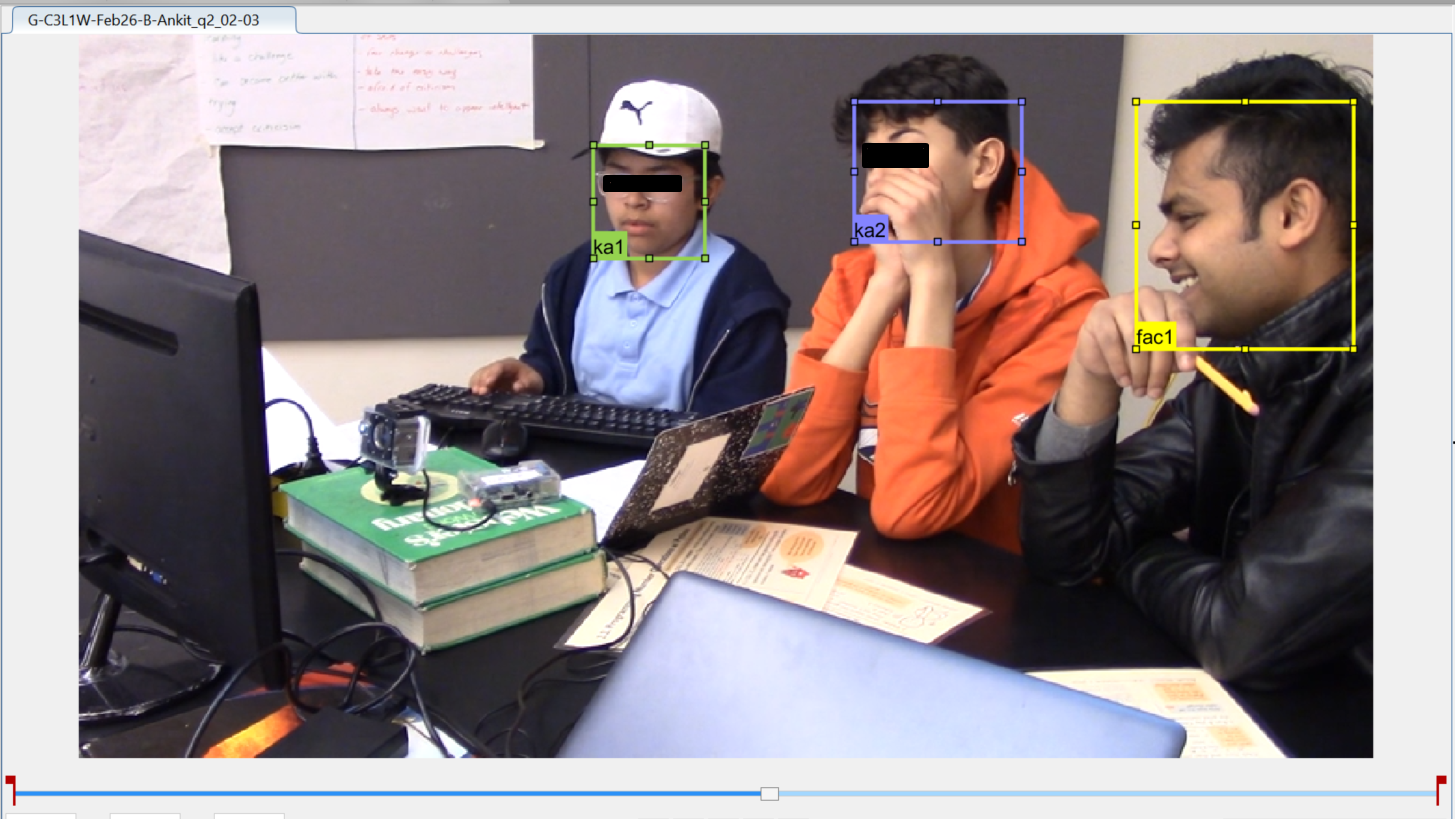}
	\caption{Process for generating ground truth for face locations.}
	\label{fig:Matlab_gt}
\end{figure}

We next describe the process of generating ground truth for each block. Each block is assigned the percentage of face pixels that lie within the block. The shrank frames from the previous process are loaded into a video labeler software. Here, each face is manually selected by a rectangle and assigned a tag for each person. This process is shown in figure \ref{fig:Matlab_gt}.

\begin{figure}	[bt]
	\includegraphics[width=0.9\columnwidth]{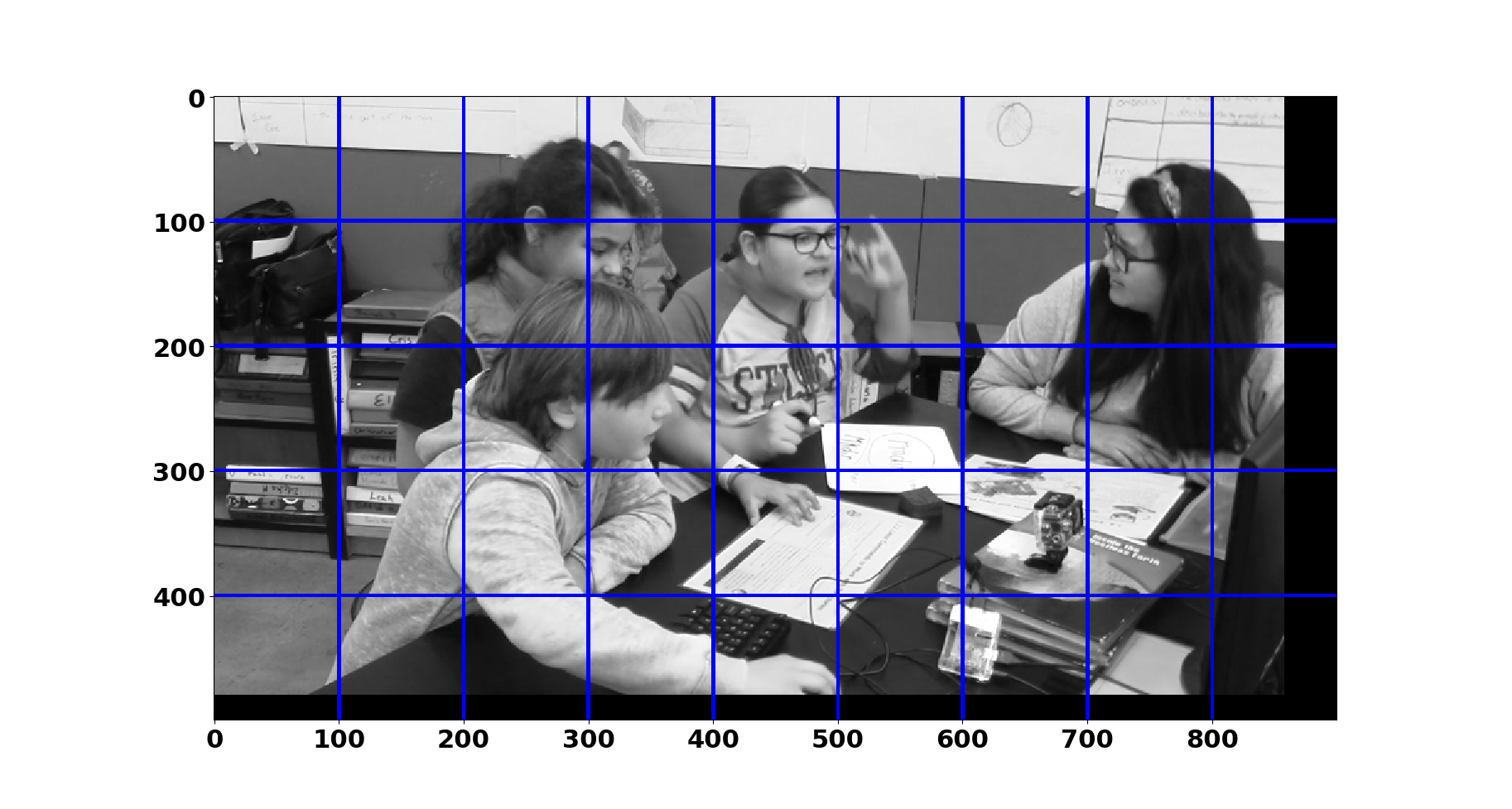}
	\caption{Example of zero padded frame divided into 9 by 5 blocks of $50 \times 50$ pixels.}
	\label{fig:GT_masks_1}
\end{figure}

\begin{figure}	[bt]
	\includegraphics[width=0.9\columnwidth]{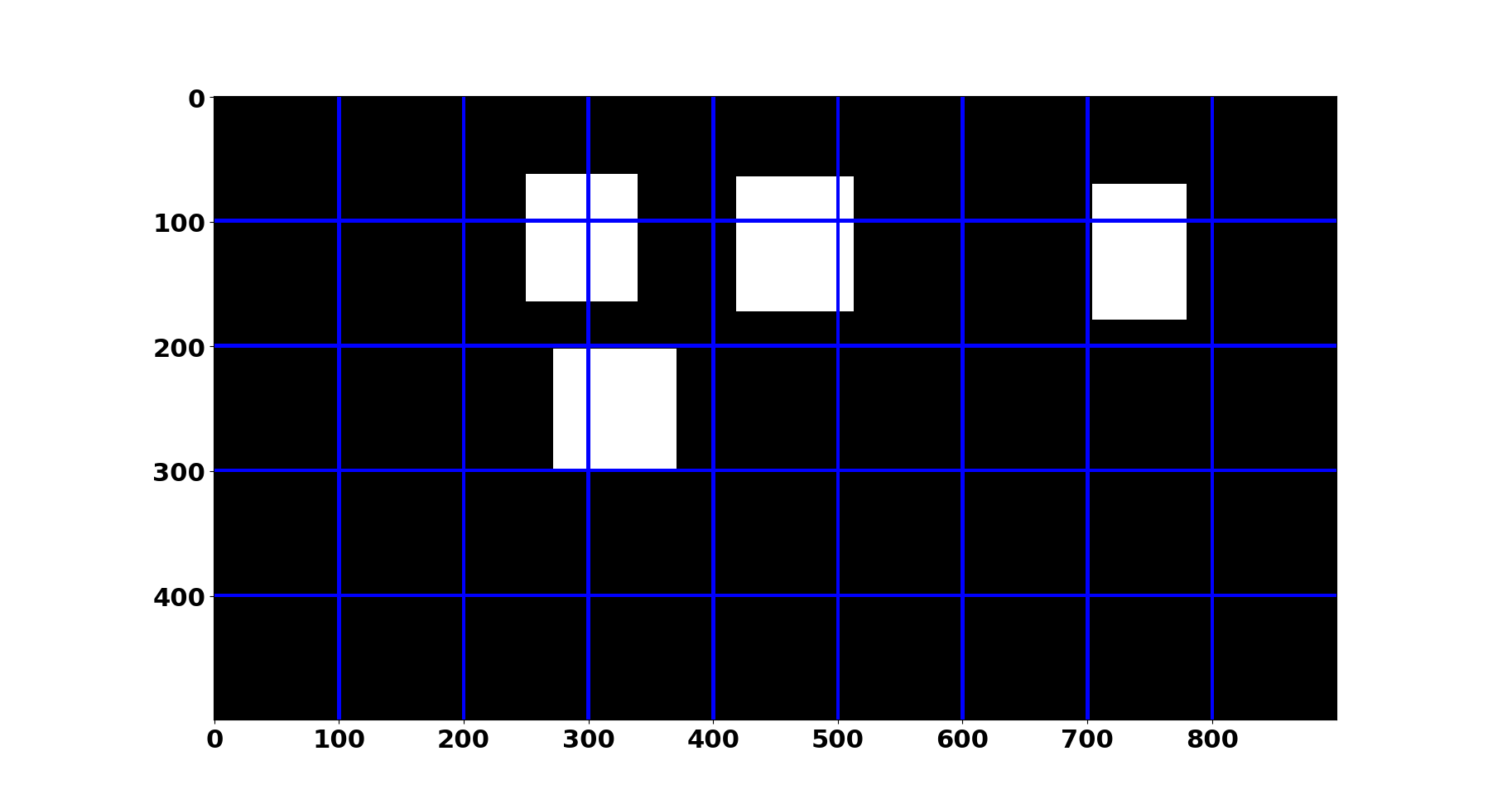}
	\caption{Face rectangles from ground truth overlapped with the block image. Refer to figure \ref{fig:GT_masks_1} for the original image. }
	\label{fig:GT_masks_2}
\end{figure}

\begin{figure}	[bt]
	\includegraphics[width=\columnwidth]{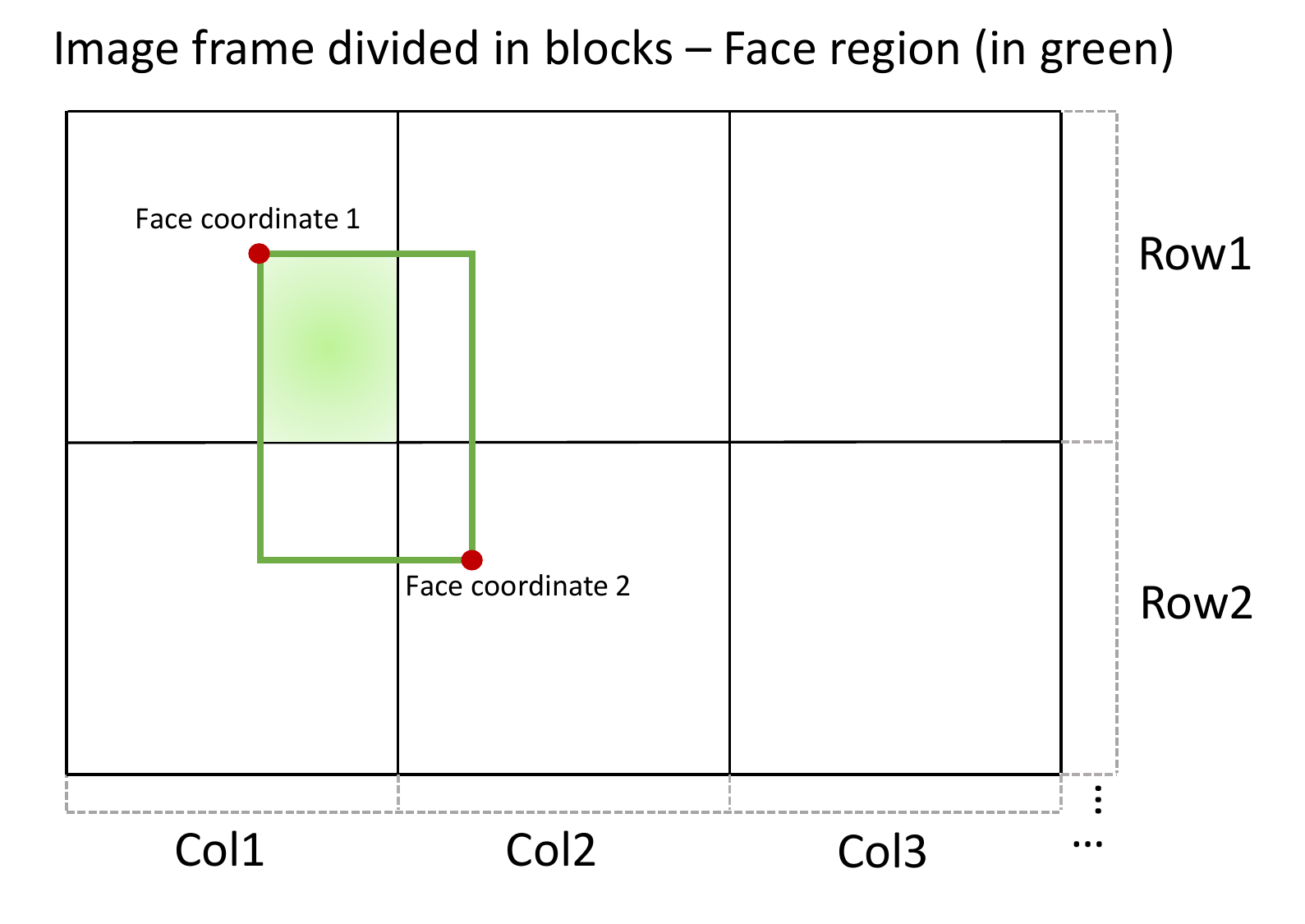}
	\caption{ Overlap ratio calculation based on the percentage of the face pixels that lie within each predefined block.	}
	\label{fig:GT_frames_face}
\end{figure}

\begin{figure}	[bt]
	\includegraphics[width=\columnwidth]{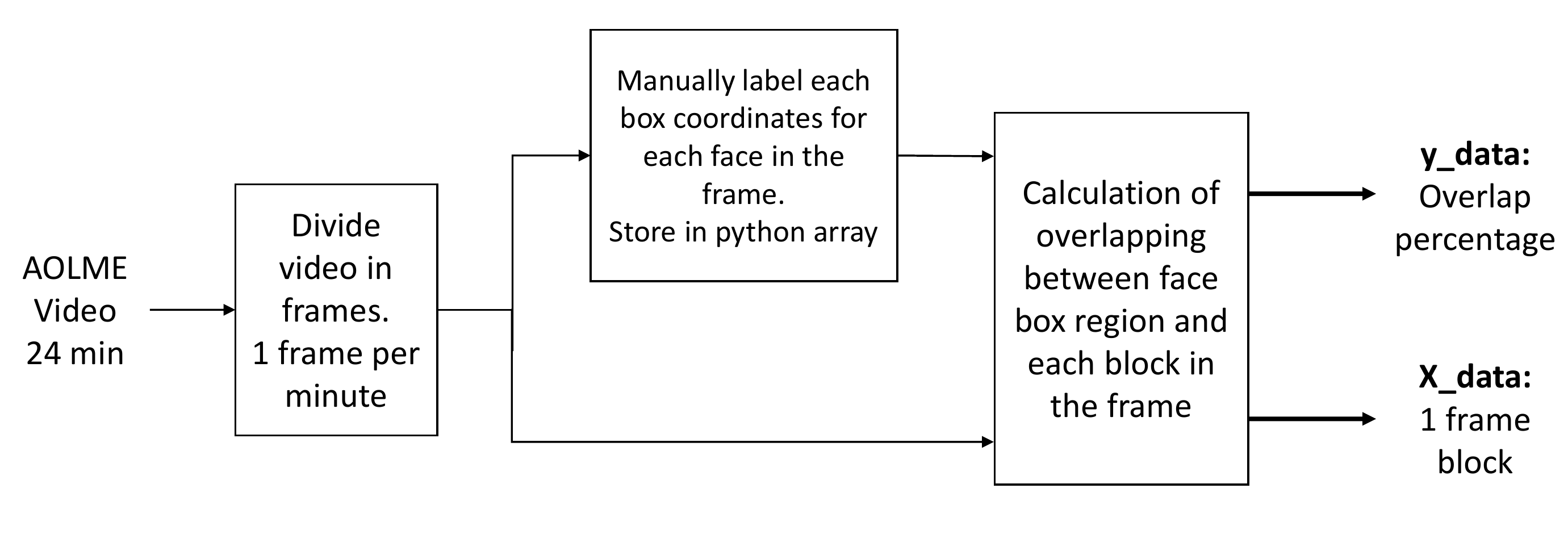}
	\caption{ Summary of ground truth process for assigning the overlap percentage for each block. X\_data corresponds to the array of all the blocks (in order) within each frame. Y\_data is a 1D array of all the overlapping percentages for each of the blocks in X\_data.}
	\label{fig:gt_blockDiagram}
\end{figure}

The software exports an array with the coordinates of the location of each of the faces. The next step is to allocate these coordinates into the 9 by 5 grid used in the image data generation. Figures \ref{fig:GT_masks_1} and \ref{fig:GT_masks_2} presents an example of the face rectangles being allocated into the blocks. For each of the blocks, the amount of pixels from the face rectangle are counted, then divided by the block area. The same process is repeated for all of the blocks. Figure \ref{fig:GT_frames_face} shows the area of the face rectangle over the first block in green. 
After all of the blocks have been assigned an overlap value, the data is stored in a 1D array to be used by the machine learning methods. A summary of both process is depicted in figure \ref{fig:gt_blockDiagram}.

\section{AM-FM Demodulation}

Let a 2D image $ I(x,y)  $ be a function of spatial coordinates $ x $ and $ y $. $ I(x,y)  $ can be represented as linear combination of AM-FM components: 
\begin{equation}
I(x,y)=\sum_{n=1}^K A_{n}(x,y) \cos\psi_{n}(x,y) \label{eq:1}	
\end{equation}
where the $ A_{n} $ represents the Instantaneous Amplitude (IA) components and  $\psi_{n}$ represent the Instantaneous Phase (IP) components.

The amplitude modulated components $ A_{n}(x,y) $ are non-negative. In addition to the IA, the Frequency modulated components $ \cos\psi_{n}(x,y) $ capture the fast changes in spatial variability among the image intensity values \cite{multiscale}. The Instantaneous Frequency (IF) components are defined as the gradient of the phase functions:

\begin{equation}
\nabla \psi_{n}(x,y) = \left( \frac{\partial \psi_{n}}{\partial x} (x,y) , \frac{\partial \psi_{n}}{\partial y} (x,y)  \right). \label{eq:2}	
\end{equation}

In what follows, we will describe Dominant Component Analysis (DCA) for estimating a single AM-FM component. First we compute the analytic image using:
\begin{equation}
I_{AS}(x,y) = I(x,y) + \jmath \mathscr{H} \{ I(x,y)  \}  \label{eq:3}	
\end{equation}
where $ \mathscr{H} $ denotes the 1D Hilbert operatos applied along each row. After filtering the analytic image through a collection of band pass filters, we obtain:
\begin{equation}
I_{n}(x,y) \approx A_{n}(x,y) \exp [\jmath \psi_{n}(x,y) ]  \label{eq:4}	
\end{equation}
where $ n $ represents the index of each filter. The IA and IP can be estimated by using:
\begin{align}
 A_{n}(x,y) &=  | I_{n_AS}(x,y) |  \label{eq:5} \\
 \intertext{and}
 \psi_{n}(x,y) &=  \arctan\left[ \frac{\operatorname{Im}(I_{n}(x,y))}{\operatorname{Re}(I_{n}(x,y))} \right]. \label{eq:6}	
\end{align}

\begin{figure}	[bt]
	\includegraphics[width=\columnwidth]{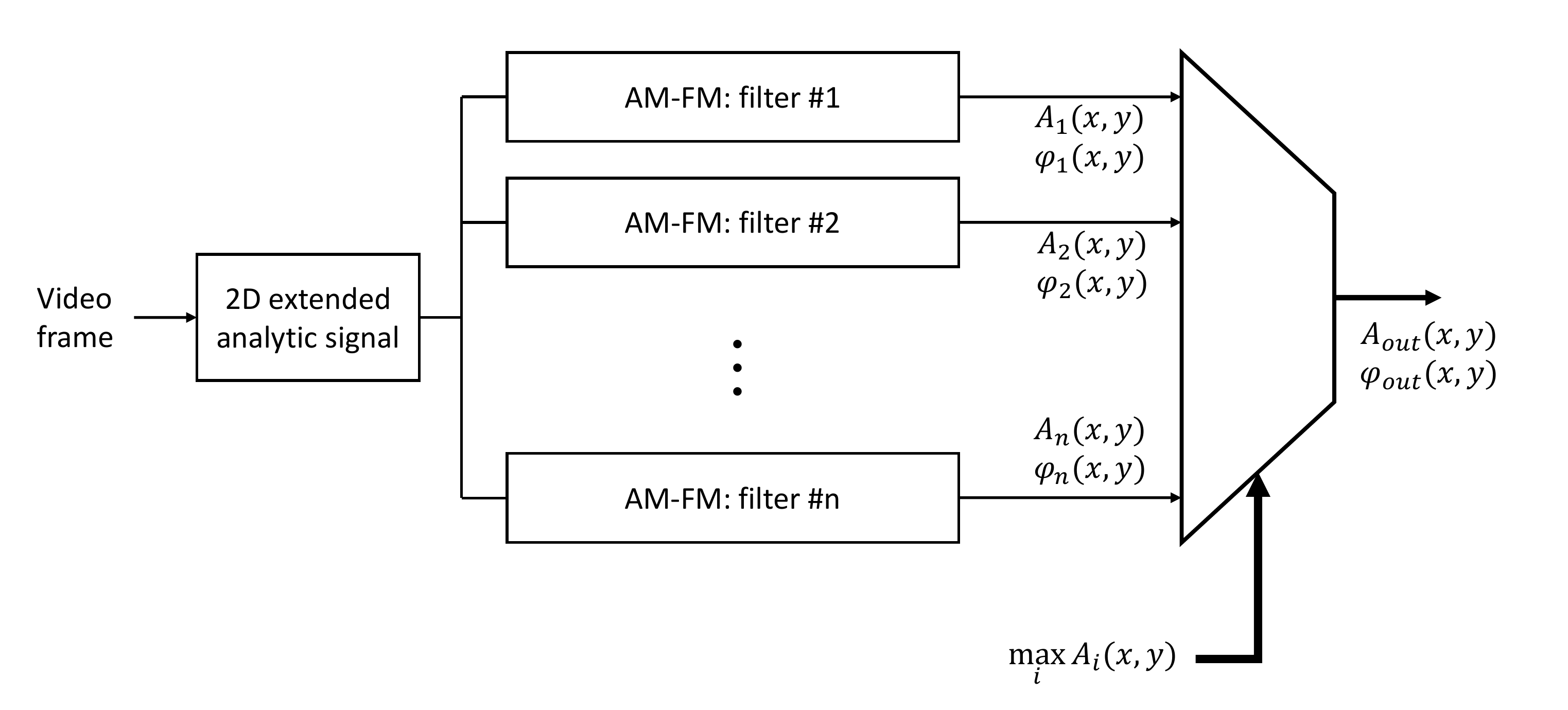}
	\caption{Dominant Component Analysis.}
	\label{fig:amfmfilterbank}
\end{figure}

Dominant Component Analysis (DCA) extracts one Instantaneous Amplitude (IA) and one Instantaneous Phase (IP) for each filterbank. DCA consists of selecting estimates from the channel with the largest IA, the dominant component, from all the filter outputs for each pixel. Subsequently, the IP is obtained from the selected pixel using equation \eqref{eq:6}. DCA is summarized in figure \ref{fig:amfmfilterbank}.

According to the focus of this research, in \cite{image_analysis1}, the authors used the phase modulation components to model tree growth. Additional references for the AM-FM theory were used from \cite{medical8}.
Also, regarding the Gabor filterbank with DCA, AM-FM methods has been used as a feature extraction for head and hair detection in group interactions, refer to \cite{shi2} and \cite{shi3}.

We also provide a summary of related AM-FM research.
Early applications of AM-FM models for image analysis include alternative fingerprint representations for fingerprint classification \cite{finger1}. 
Also, Foveated Video Compression using a nonuniform filtering scheme is presented in \cite{compression1}. An extension of this research presents a framework for assessing the quality of foveated images and video
streams \cite{quality1}. 
Orthogonal AM-FM transforms can be interpreted as a permutation of signal samples followed by the regular
DFT, in \cite{compression2} permutations are applied to compact broadband signal.  
In \cite{multi2}, the authors described the development of FM transforms based
on the use of permutations and the 2D FFT or 2D DCT.

Biomedical applications in computer aided diagnosis (CAD) had use multiscale AM-FM methods, examples of this methods are presented in the rest of the paragraph. Multiscale AM-FM to identify repetitive structures in \cite{medical1}.
Optimal thresholding of the IA components to generate candidate regions as in \cite{medical3}.
Feature extraction from fundus images to characterize normal and pathologic structures in \cite{medical2}, \cite{medical7} and \cite{medical4}.
Feature extraction from ultrasound images for despeckle filtering in \cite{ultrasound1}, an extension of this research also uses multiscale feature extraction to determine texture differences between classification groups can be found in \cite{ultrasound2}.
In \cite{medical5}, authors used AM-FM features that provided additional information from MRI images, resulting in better classification results when using the combination of the low-scale IA and IF magnitude with the medium-scale IA.
In \cite{medical6}, authors shown AM-FM features along SVM classification methods obtained 78\% in classifying neuromuscular disorders from surface electromyographic (SEMG) signals.

\section{Low complexity CNN architectures for large video dataset training}

There is no truly formal scientific understanding on how a Neural network works. Yet, there is no doubt on their effectiveness in revealing patterns and predicting or classifying data. Convolutional Deep learning methods have greatly advanced, becoming an essential part of many successful image and video processing systems.

Specific to the current research, there are two Convolutional Neural Networks (CNN) architectures of interest: (a) the LeNet-5 by Yann LeCun \cite{lecun98} and (b) MobileNetV2 as presented in \cite{mobilenetv2}. These networks represent two extremes in the development of these techniques. LeNet-5 represents of the earliest examples of convolutional neural networks associated with the deep learning revolution. MobileNet V2 represents the state of the art in low complexity neural nets with low power consumption for embedded devices.

\subsection{LeNet-5}

\begin{figure}	[bt]
	\includegraphics[width=\columnwidth]{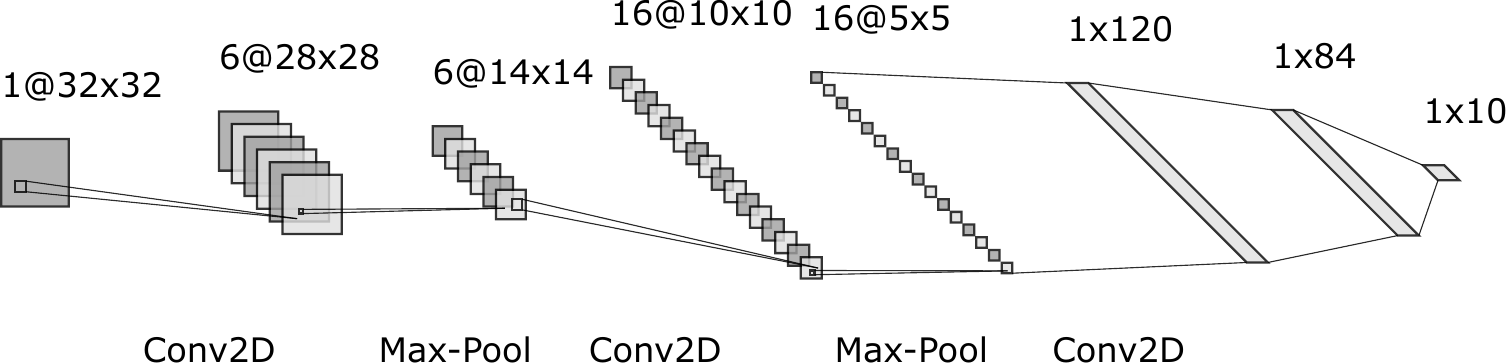}
	\caption{Representation of the LeNet-5 architecture and their layers.}
	\label{fig:lenetBasic}
\end{figure}

 Figure \ref{fig:lenetBasic} depicts the general structure of the LeNet-5. LeNet-5 consists of: (i) convolutional layers followed by a pooling layer and (ii) fully connected layers. For our application, we modified the activations functions to enable better training. Refer to table \ref{tab:lenetTable} for details.

\begin{table}[bt]
	\centering
	
	\begin{center}
		\begin{tabular}{||c c c c c c c||} 
			\hline
			Layer & Type & Maps & Size & Kernel Size & Stride & Activation \\ [0.5ex] 
			\hline\hline
			Output & Fully Connected & - & 10 & - & - & RBF \\ 
			\hline
			F6 & Fully Connected & - & 84  & - & - & tanh \\
			\hline
			C5 & Convolution & 120 & 1x1  & 5x5 & 1 & tanh \\
			\hline
			S4 & Avg Pooling & 16 & 5x5  & 2x2 & 2 & tanh \\
			\hline
			C3 & Convolution & 16 & 10x10  & 5x5 & 1 & tanh \\
			\hline
			S2 & Avg Pooling & 6 & 14x14  & 2x2 & 2 & tanh \\
			\hline
			C1 & Convolution & 6 & 28x28  & 5x5 & 1 & tanh \\
			\hline
			In & Input & 1 & 32x32 & - & - & - \\ [1ex] 
			\hline
		\end{tabular}
	\end{center}
	
	\caption{Architecture of LeNet-5 \cite{bookcnn}.}
	\label{tab:lenetTable}
\end{table}

\subsection{MobileNetV2 Architecture}

MobileNetV2 comprises the state of the art in CNN for embedded devices, featuring  a novel Bottlenecks Residual block that comprises the convolution layer stage, providing low power consumption with different quantization capabilities \cite{mobilenetv2}. This network outperforms the simple LeNet-5 when using regular RGB or grayscale data \ref{tab:MobileGeneralTable}. This network was chosen because it was already optimized to operate fast and in a quantization framework, hence allowing fixed-point hardware implementations. 

The general architecture is described in table \ref{tab:MobileGeneralTable}. The Bottleneck layer mentioned in the table makes reference to the inverted residual convolution layer, where a low dimensional input is expanded to extract features and then compressed again.
\begin{table}[bt]
	\centering
	
	\begin{center}
		\begin{tabular}{||c c c c c c c||} 
			\hline
			Layer & Type        &  Size        & t & c  & n & s \\ [0.5ex] 
			\hline\hline
			In    & conv2D      & 224x224x3    & - & 32 & 1 & 2  \\ 
			\hline
			B1	  & bottleneck  &  112x112x32  & 1 & 16 & 1 & 1     \\
			\hline
			B2	  & bottleneck  &  112x112x16  & 6 & 24 & 2 & 2      \\			
			\hline
			B3	  & bottleneck  &  56x56x24    & 6 & 32 & 3 & 2      \\
			\hline
			B4    & bottleneck  &  28x28x32    & 6 & 64 & 4 & 2      \\
			\hline
			B5	  & bottleneck  &  14x14x64    & 6 & 96 & 3 & 1      \\
			\hline
			B6	  & bottleneck  &  14x14x96    & 6 & 160 & 3 & 2     \\
			\hline
			B7	  & bottleneck  &  7x7x160     & 6 & 320 & 1 & 1    \\
			\hline
			C8	  & conv2D 1x1  &  7x7x320     & - & 1280 & 1 & 1   \\			
			\hline
			S9	  & Avg Pool    &  7x7x1280    & - & -    & 1 & 7    \\			
			\hline
			C10  & conv2D 1x1  &  7x7x1280    & - & k    & - & -    \\ [1ex]
			\hline
		\end{tabular}
	\end{center}
	
	\caption{MobileNet V2 architecture \cite{mobilenetv2}. Refer to \cite{mobilenetv2} for a description of bottleneck. The top row of the table represents the input layer. C20 represents the output layer. }
	\label{tab:MobileGeneralTable}
\end{table}

MobileNet V2 has been trained with ImageNet for image classification and the COCO dataset for object detection. For ImageNet, 3.4 million parameters were trained using 300 million Multiply-Adds. For object detection, training included a SSD prediction layer \cite{ssd} in which the convolutional layers used separable convolutions to reduce the number of parameters and computational cost. The paramaters for training the COCO dataset were 4.3 million, requiring a total of 0.8 million Multiply-Adds. 

\chapter{Methodology}
In the following chapter the methods for preprocessing the dataset will be detailed. According to the AM-FM theory presented in the previous chapter, the first step will consist in the generation of the analytic image, using the Hilbert Transform. Then, the real plus imaginary component will be processed in a Gabor filterbank with DCA to obtain an Instantaneous Amplitude and Instantaneous Phase representation. In the next chapters, the method for using these two representations as an alternative for the original image will be discussed.  

\section{System Overview}

\begin{figure}[bt]
	\centering
	\includegraphics[width=\columnwidth]{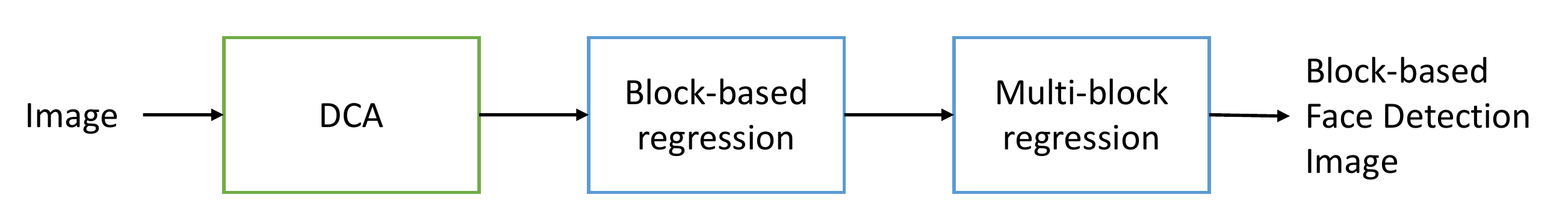}
	\caption{ System overview for block face detection.}
	\label{fig:systemsummary}
\end{figure}

In this section, a general description of the system workflow is presented. The frames are processed by Dominant Component Analysis (see figure \ref{fig:amfmfilterbank}). Then, the output FM components are fed to the Single-Block regression system to produce block predictions. Last, the block predictions are arranged into frames to enter the Multi-Block regression systems to produce a block-based detection image. A summary of the process is shown in figure \ref{fig:systemsummary}.

\section{Fixed-Point Design of Hilbert Transform using Simulated Annealing}

Fixed-point realizations of digital filters can enable faster processing of large datasets. Fixed-point filters can also be implemented efficiently on FPGAs. To reduce computational complexity, this section describes the use of a Simulated Annealing method for determining the optimal 8-bit filtering coefficients without sacrificing accuracy.

\begin{figure}[bt]
	\centering
	\includegraphics[width=\columnwidth]{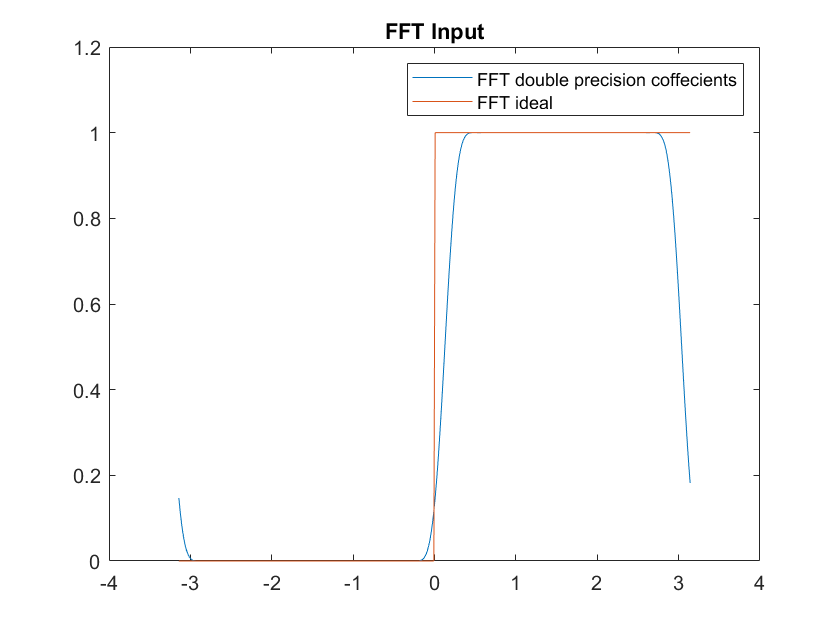}
	\caption{Fast Fourier Transform Magnitude representation of a double-precision floating point filter against the ideal frequency response (in red).}
	\label{fig:Double_precision}
\end{figure}

The Hilbert Transform filter (in blue) depicted in figure \ref{fig:Double_precision} was designed using Frequency domain sampling. Then, a Kaiser Window method is applied using a 51 double-precision floating point coefficients. An odd number of coefficients was used so that the zeroth sample represents the zeroth sample of the filter, allowing the filter to be linear phase. Also, this forces a half of sample delay in the filter, which should be taken into account for further testing of the linear phase. In color red is shown the ideal magnitude frequency response of the Hilbert Filter. Here, all the negative frequencies are reduced to zero, except for the double precision coefficients at the transition and edge bands. Later, we will use non-zero transition bands to allow efficient fixed-point implementations. 

\begin{figure}[bt]
	\centering
	\includegraphics[width=\columnwidth]{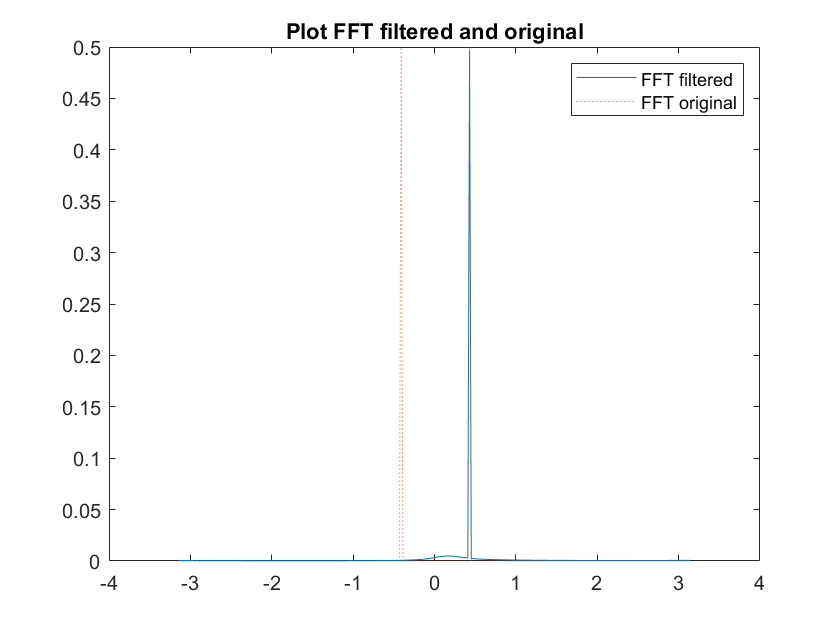}
	\caption{Magnitude plot for input sine and filtered sine. The negative sine peak is filtered out at the output signal.}
	\label{fig:FIR_sine_mag}
\end{figure}

The testing of the FIR filter had two steps. First, we need to check that the negatives frequencies are successfully wiped out. For doing this, a simple experiment is designed, where a sinusoidal function is input with the expected result being that there would be only one magnitude peak in the positive part of the spectrum, while zero everywhere else. The figure \ref{fig:FIR_sine_mag} clearly shows the result of the filtered sine, in blue, against the original sine function magnitude plot in red.

\begin{figure}[bt]
	\centering
	\includegraphics[width=\columnwidth]{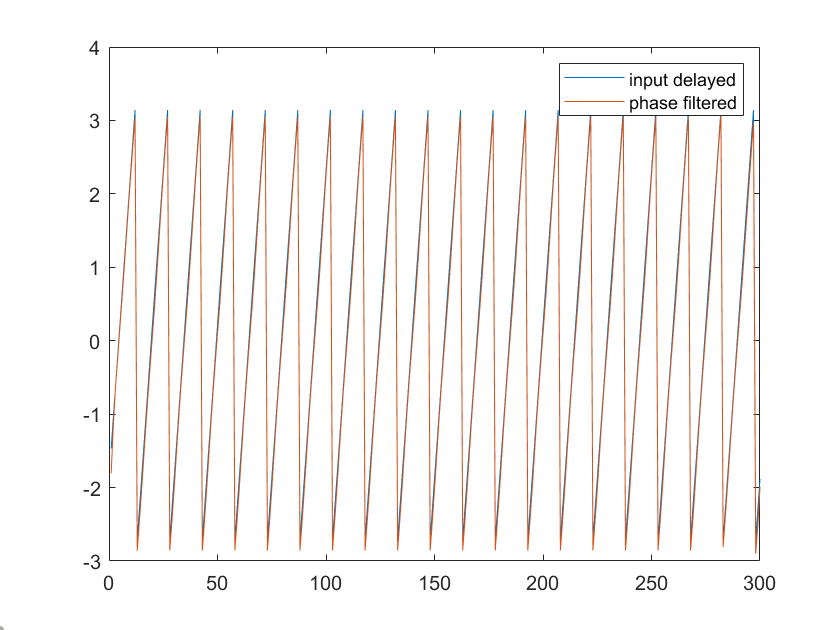}
	\caption{Phase from input sine against filtered sine.}
	\label{fig:FIR_phase}
\end{figure}

Second, the linearity of the phase should be tested. To test this, we use a sinusoidal function given by $ \cos (2*pi/N*(u*x1)) $, where $ u = 20 $, $ N=300 $, $ x_{1} $ is a 1D vector of samples from  0 to $ N $ .Then, the phase is delayed by 4.5 samples to accommodate for the processing delay of the filter.This delayed phase is the one that will be compared against the filtered phase. Later, using with the 51 fixed-point coefficients along a 1D convolution function, the sinusoidal function and filtered phase is extracted using the arc tangent function. The filtered phase is depicted in figure \ref{fig:FIR_phase} in red color, while the delayed phase is in blue. As it can be seen, the matching is almost identical, confirming the linearity of the phase.

We used Simulated Annealing to design the Hilbert Filters (see fig. \ref{fig:SA_code}). Simulated Annealing (SA) is a stochastic optimization method to find the global maximum over a certain objective function. SA has the characteristic of accepting steps that do not necessarily resolve in immediate improvement of the objective function, thus allowing the method to escape from a local minimum according to a certain probability calculated at each iteration. This probability is compared to a random number to determine whether it would be a jump to another set of input vector. Initially the probability of a jump is higher, and it gets lowered for later jumps.

Here, a neighborhood is defined in terms of the vector coefficients. Vector $ y $ is calculated by adding/subtracting a  $ delta\_step $ to 1 coefficient in $ x $. $ Delta\_step $ depends on the number of bits used for quantization. From the theory \cite{ross}, it is required for the neighborhood to be completely reachable from every point by using the $delta\_steps$ defined previously.

The objective function to be maximized is the negative MSE between the zero-padded FFT magnitude of the input vector and the ideal FFT desired :

\begin{equation}
f_{x} = {\tt mse(abs(fft(x)),abs(fft(ideal)))}. \label{eq:7}
\end{equation}

\begin{figure}[bt]
	\centering
	\includegraphics[width=\columnwidth]{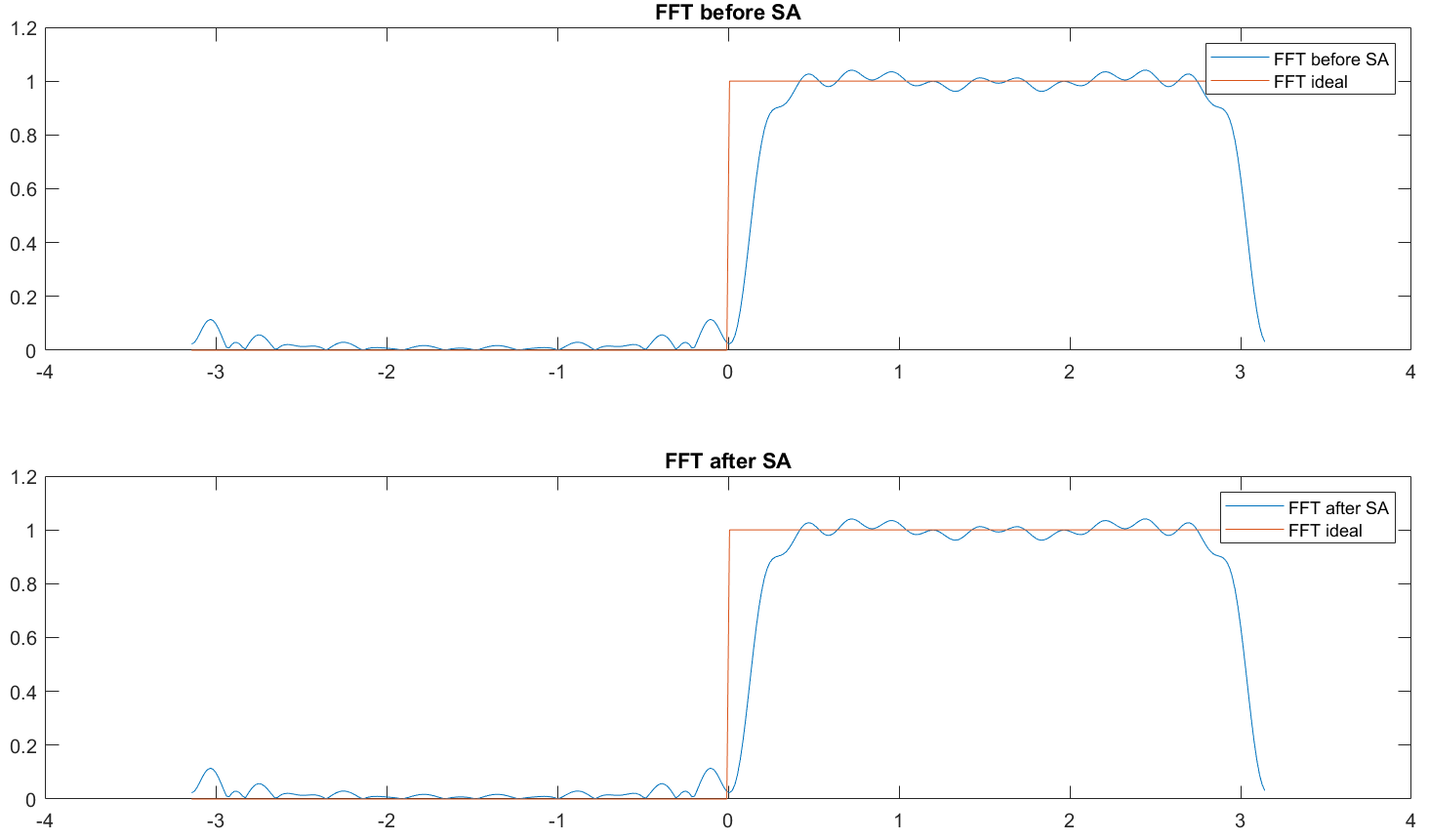}
	\caption{FFT Magnitude for 8 bit quantization. The top plot represents the frequency response of the input coefficients. The bottom plot represents the final design using transition of 0.2.}
	\label{fig:fft_b8f0}
\end{figure}

\begin{figure}[bt]
	\centering
	\includegraphics[width=\columnwidth]{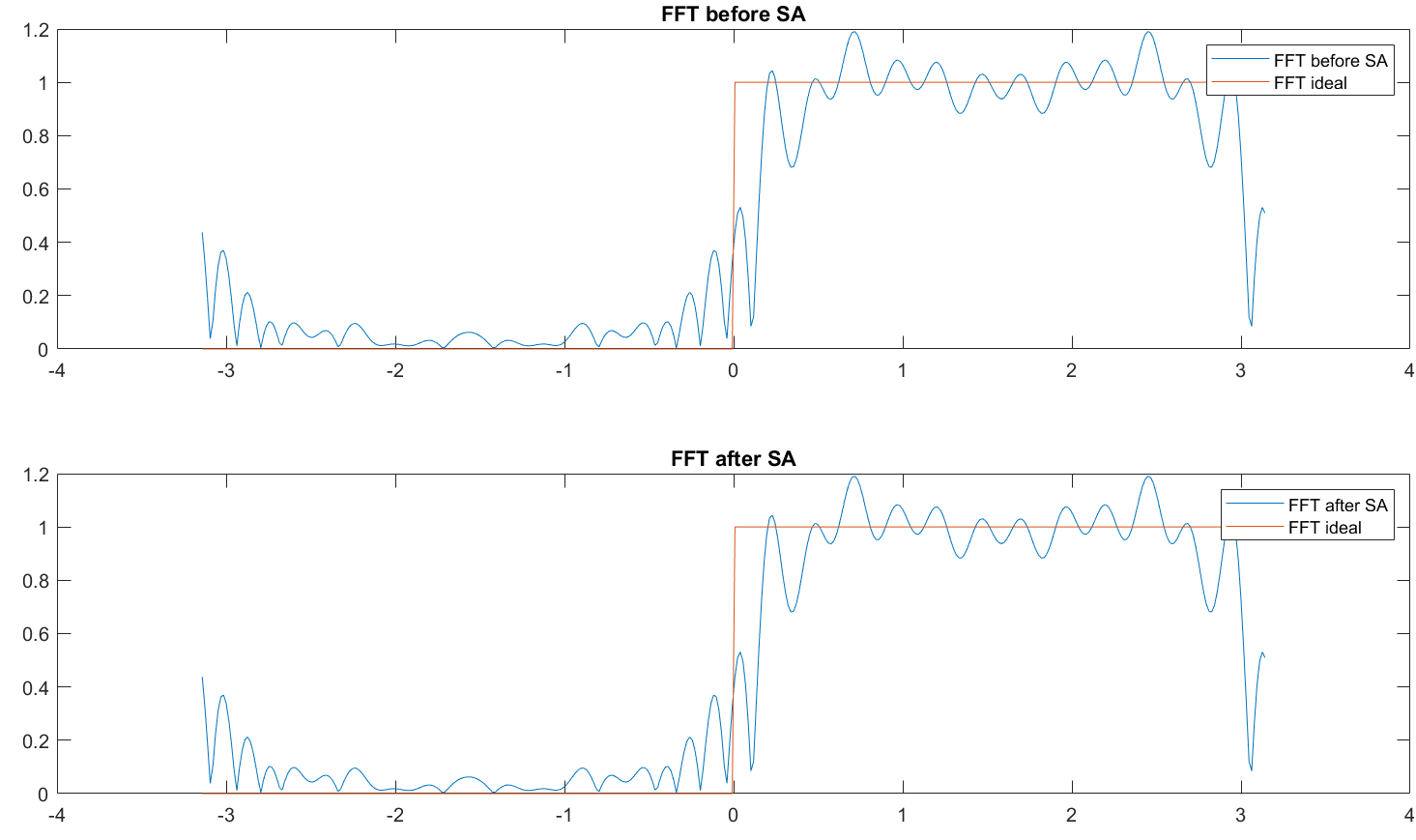}
	\caption{FFT Magnitude plots for 6 bit quantization. Top plot depicts the input coefficients without any modifications. The bottom plot represents the result after simulated annealing is applied.}
	\label{fig:fft_b6f0}
\end{figure}

A visual debugging is possible when comparing the plots of the FFT Magnitude from (a) the input coefficients without modification and (b) output coefficients of the Simulated Annealing. Figures \ref{fig:fft_b8f0} and \ref{fig:fft_b6f0} show the variations in the magnitude for the filter generated by quantizing with 8 and 6 bits (top subplot), and the filter after the Simulated Annealing correction was applied (bottom subplot).

\begin{figure}[bt]
\begin{lstlisting}[language=Python]
# Extra variables
sections = 1/length(x0);

for iter = 1 : MaxIterations
	# Select a neighbor from 4 possibilities:
	p = rand  # Uniform random variable from 0 to 1
	index_x = ceil(p/(sections))
	sign_flag = mod(floor(p*10),2)
	
	% Generate symmetry flags and index
	sign_flag = mod(floor(p*10),2) ;	
	if (sign_flag==0)
		sign_flag_symm = 1;
	else
		sign_flag_symm = 0;
	end	  
	middle_indx = (length(x0)+1)/2;
	index_x_symm = 2*middle_indx - index_x;	
	
	% Apply the selected choice with wrap-around:
	y=x;
	y(index_x) = x(index_x) + ((-1)^sign_flag)*step_x ;	
	y(index_x_symm) = x(index_x_symm) + ((-1)^sign_flag_symm)*step_x ;
	# New function evaluation:
	fy = obj_fun (y,band_pass_percentage,edges_percentage);

	# Evaluate the probability on whether you want to take the jump or not
	sub_c=(fy - fx)
	prob_SA =(1+iter)^(sub_c*C)
	p = min(1,prob_SA )
	value_random = rand
	
	if (value_random<p)
		x  = y;  # Jump with probability p. Else stay at x.
		fx = fy; # Update the previous values.
	end
end
\end{lstlisting}
	\caption{Simulated Annealing method for finding the optimal 8-bit filter coefficients in {\tt x }.}
\label{fig:SA_code}
\end{figure}

The increase, or decrease, on the coefficients depends on the number of bits used for the quantization. For numbers higher than 14 bits, $ 2e^{-11} $ was used for the step size. For fixed-point designs for fewer than 8 bits, the best results were accomplished by setting the step to the negative power equal to the number of bits used (i.e., $ 2e^{-8} $ for 8 bit quantization). A good practice for setting the parameter C is to check the mean between the error of the outputs of the objective function to be minimized,  $ fy-fx $. For all the cases presented here, the best results were achieved setting $ C=e^{Y} $, where Y is the power of the error. 

\section{Gabor Filterbank}

This research will use a Gabor filterbank along with a DCA to estimate AM-FM components. The Gabor equation from \cite{bovik} is presented in equation \eqref{eq:10}. Furthermore, the Gabor function is a Gaussian function shifted in frequency. Then, it is possible to generalize directional a Gaussian filter with $ \sigma $ variable in  $ x $ and $ y $ (see equation \eqref{eq:11}). In this equation, $ \theta $ defines the orientation of the filter with respect to the X-axis. 

\begin{figure}	[bt]
	\centering
	\includegraphics[width=0.7\columnwidth]{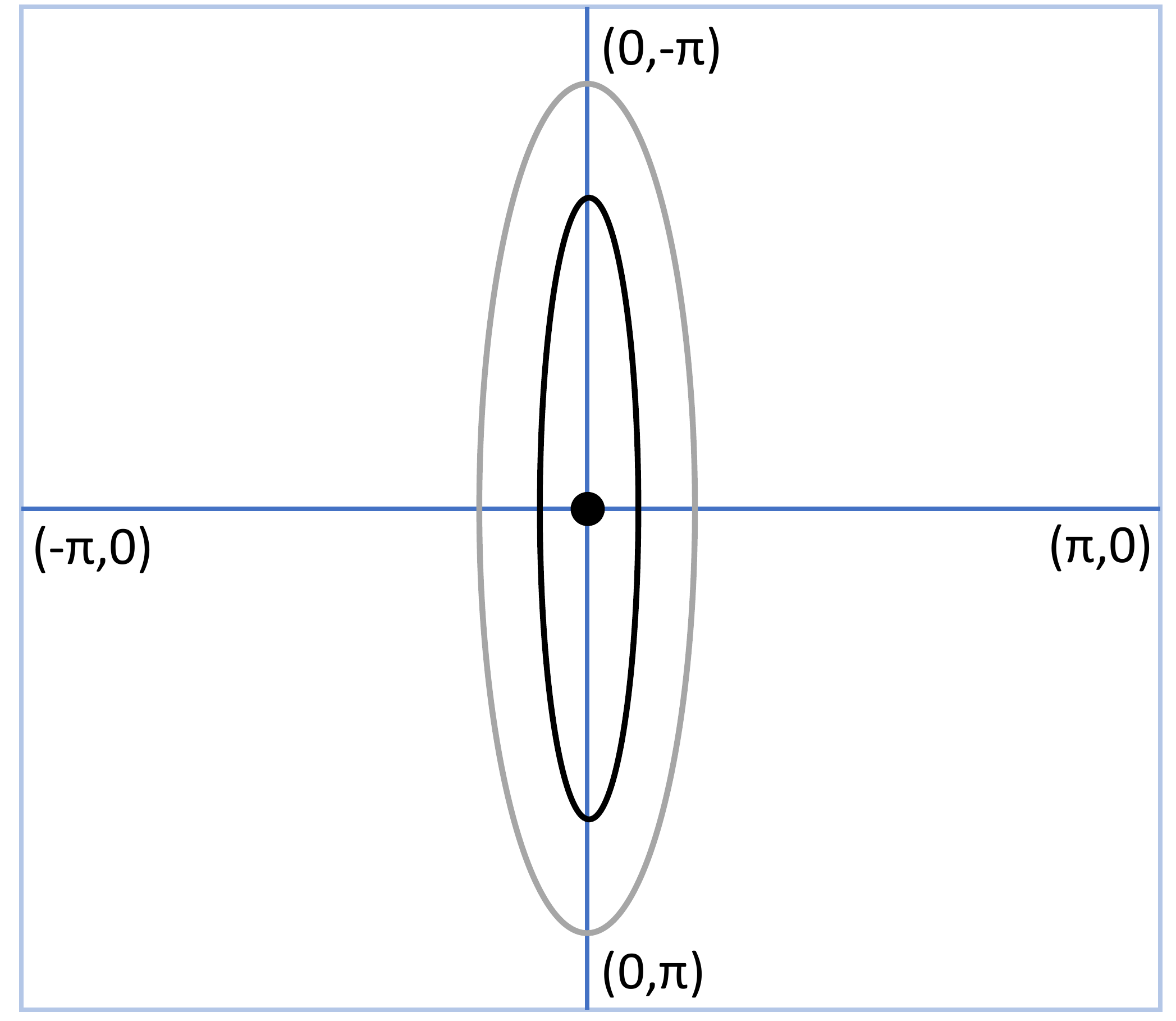}
	\caption{Single direction Gaussian filter design for different frequency-domain spreads.}
	\label{fig:SingleLobeAMFMfilter}
\end{figure}

\begin{figure}	[bt]
	\centering
	\includegraphics[width=0.7\columnwidth]{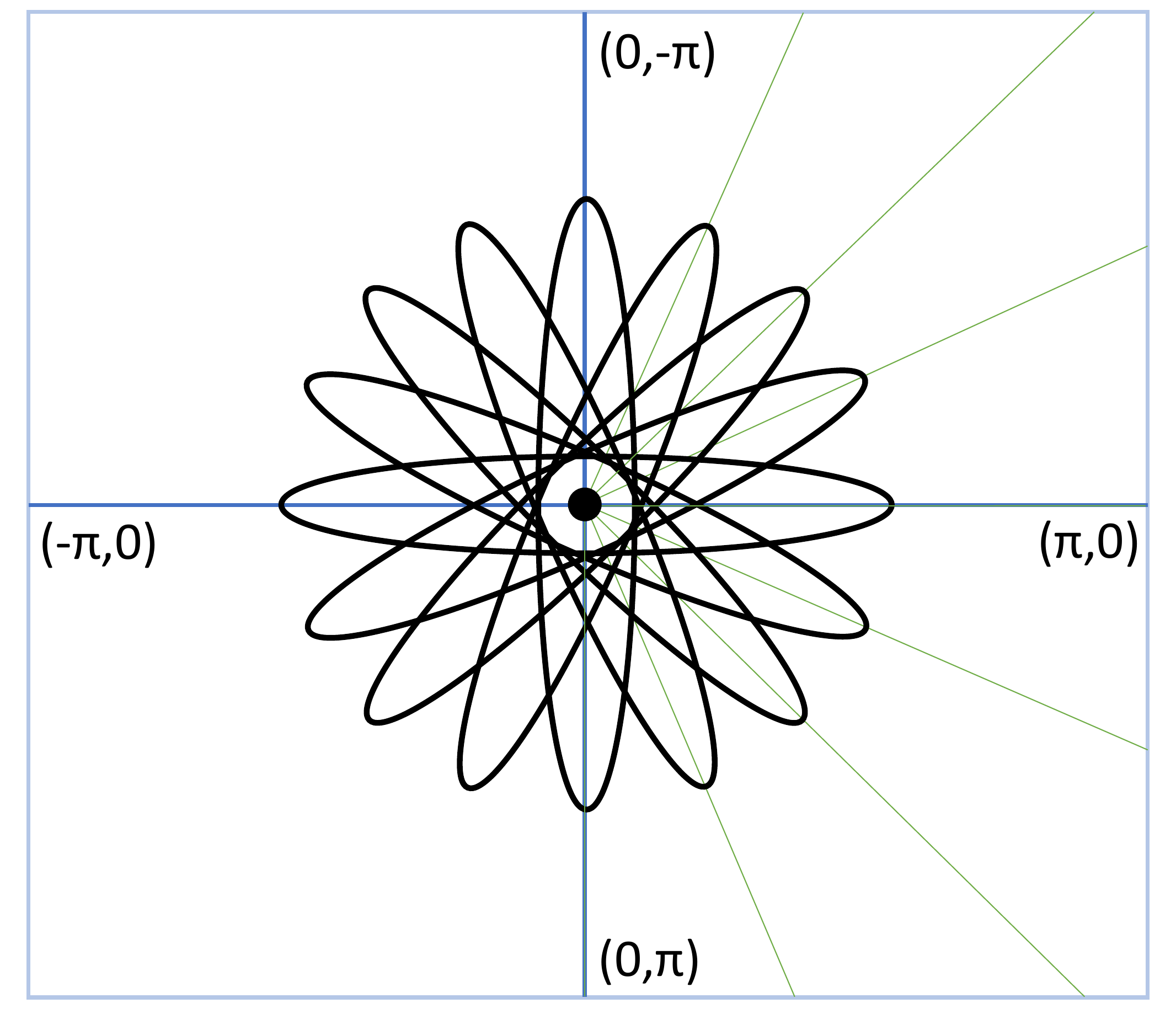}
	\caption{Directional Gaussian filter design. The example shows 8 orientations.}
	\label{fig:Level1AMFMfilter}
\end{figure}

We write:
\begin{align} 
\text{Gaussian} (x,y) &= \exp \left[ -(ax^{2} + 2bxy + cy^{2}) \right] \label{eq:11} \\
\text{Gabor} (x,y)    &= \frac{1}{2\pi \lambda \sigma^{2}} \exp \left[-(x^{2}+y^{2})/2\sigma^{2} \right] 
                          \exp \left[ \jmath 2\pi \left( \frac{u}{N}x + \frac{v}{N}y  \right) \right] \label{eq:10}	\\
\intertext{where:}
	a &= \frac{\cos^{2} \theta}{2\sigma^{2}_{x}} + \frac{\sin^{2} \theta}{2\sigma^{2}_{y}}  \\ 
	b &= -\frac{\sin^{2} \theta}{4\sigma^{2}_{x}} + \frac{\sin^{2} \theta}{4\sigma^{2}_{y}} \\ 
	c &= \frac{\sin^{2} \theta}{2\sigma^{2}_{x}} + \frac{\cos^{2} \theta}{2\sigma^{2}_{y}} 
\end{align} 	
The FFT magnitude of equation \eqref{eq:11} is shown in figure \ref{fig:SingleLobeAMFMfilter} with $ \sigma_{x} = \sigma_{y}/4 $. This directional Gaussian centered in the origin is then rotated 8 times using a step of $ \theta += 0.39 $ radians each time. The 8 orientations of the directional Gaussian are plotted together in figure \ref{fig:Level1AMFMfilter}.

\begin{figure}	[bt]
	\includegraphics[width=\columnwidth]{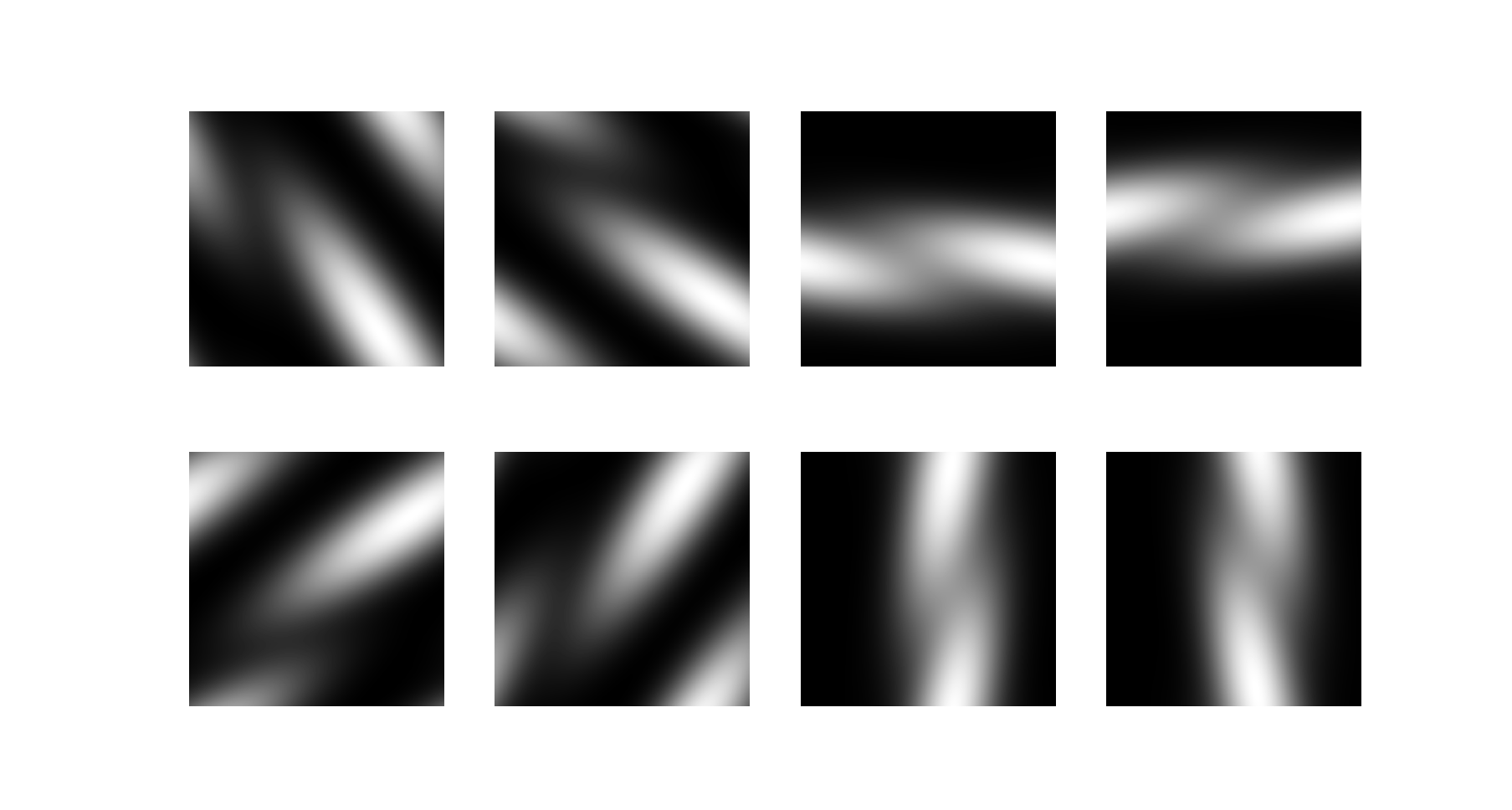}
	\caption{Low-parameter Directional Gabor implementations using $ 11 \times 11 $ filter coefficients.}
	\label{fig:FFT8kernelsLevel2}
\end{figure}

\begin{figure}	[bt]
	\includegraphics[width=\columnwidth]{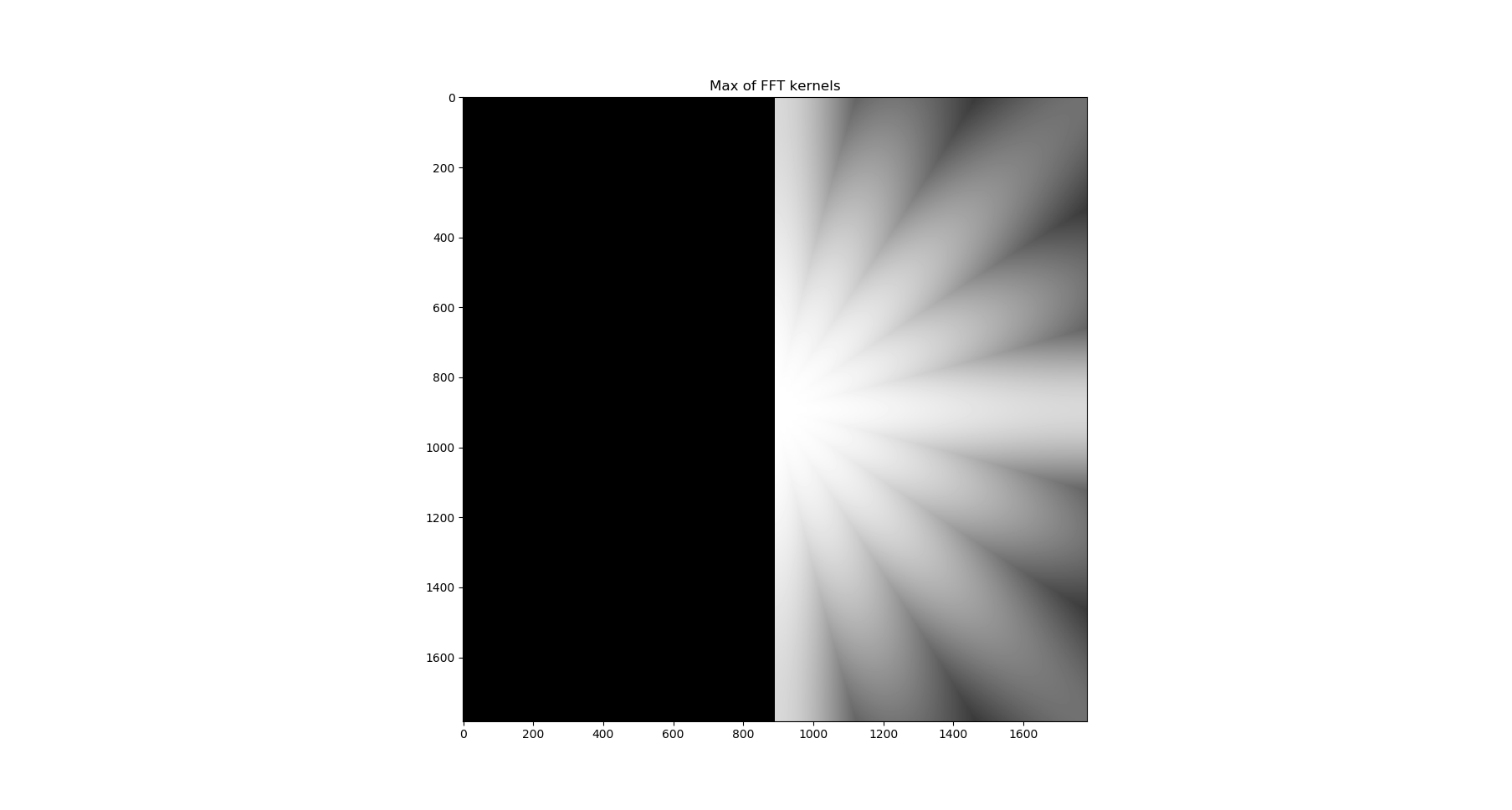}
	\caption{FFT magnitude plot from Low-parameter Directional Gaussian implementations using $ 11 \times 11 $ filter coefficients.}
	\label{fig:maxFFTlevel1}
\end{figure}

\begin{figure}	[bt]
	\includegraphics[width=\columnwidth]{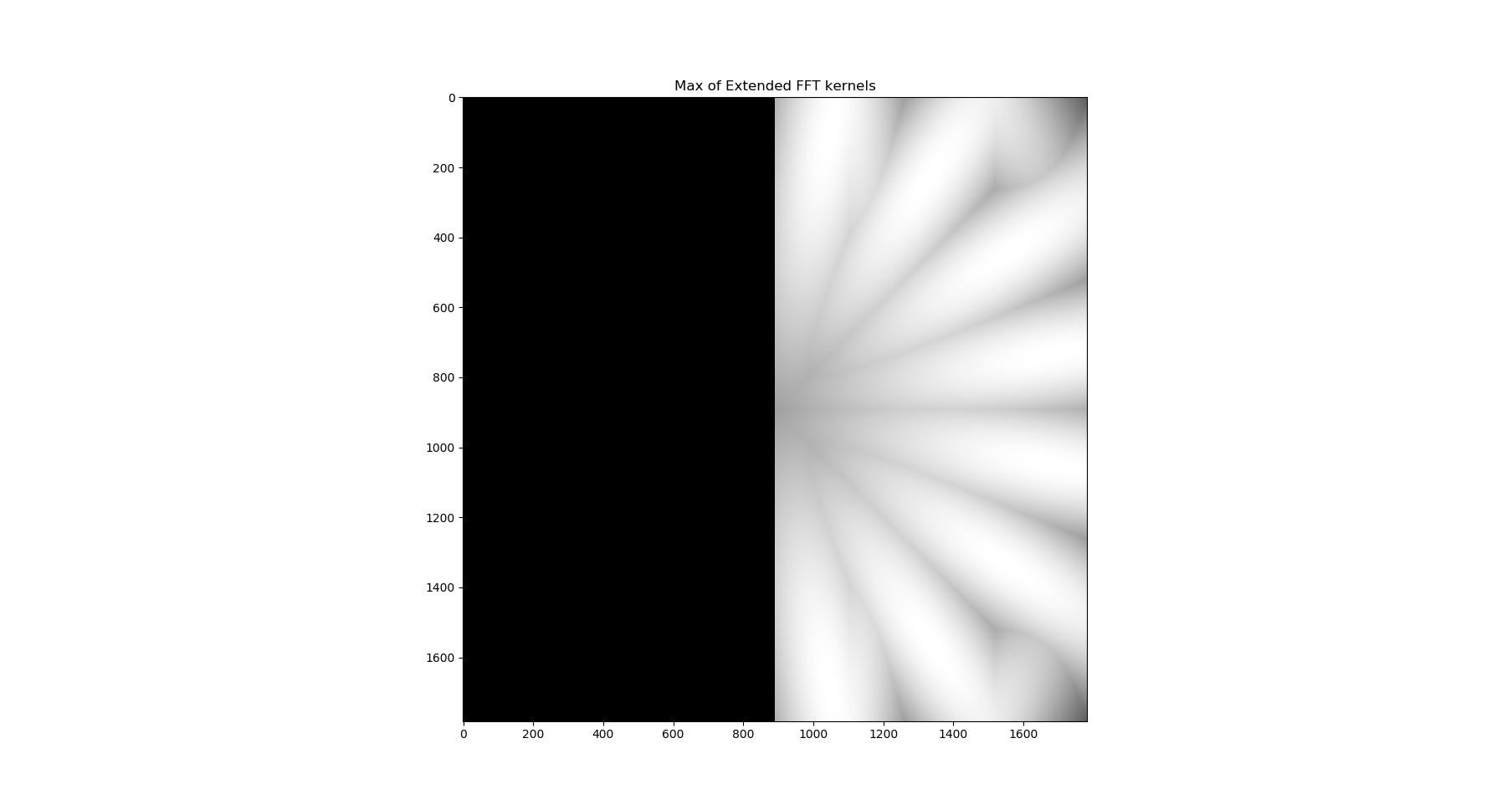}
	\caption{FFT magnitude plot from Low-parameters Directional Gabor implementations using $ 11 \times 11 $ filter coefficients. }
	\label{fig:maxFFTlevel2}
\end{figure}

The directional Gaussian is then shifted in frequency to produce the next scale in the filterbank. The number of rotations and the step of the angle are the same as the centered Gaussian, while the difference is the initial angle: $ \theta = 0.19 $. The shift corresponds to $ N/2 $, where $ N $ is the number of samples for the FFT representation. Figure \ref{fig:FFT8kernelsLevel2} shows the 8 orientations of the shifted Gaussian, the circle in the figure recalls the rotation nature, while the green lines indicate the x-axis of the previous centered Gaussians, while the purple lines depicts the current x-axis for the new shifted Gaussians. The FFT from the Directional Gaussian maximum magnitudes values are presented in \ref{fig:maxFFTlevel1}, similar representation from the Directional Gabor in \ref{fig:maxFFTlevel2}.

\begin{figure}	[bt]
	\centering
	\includegraphics[width=0.7\columnwidth]{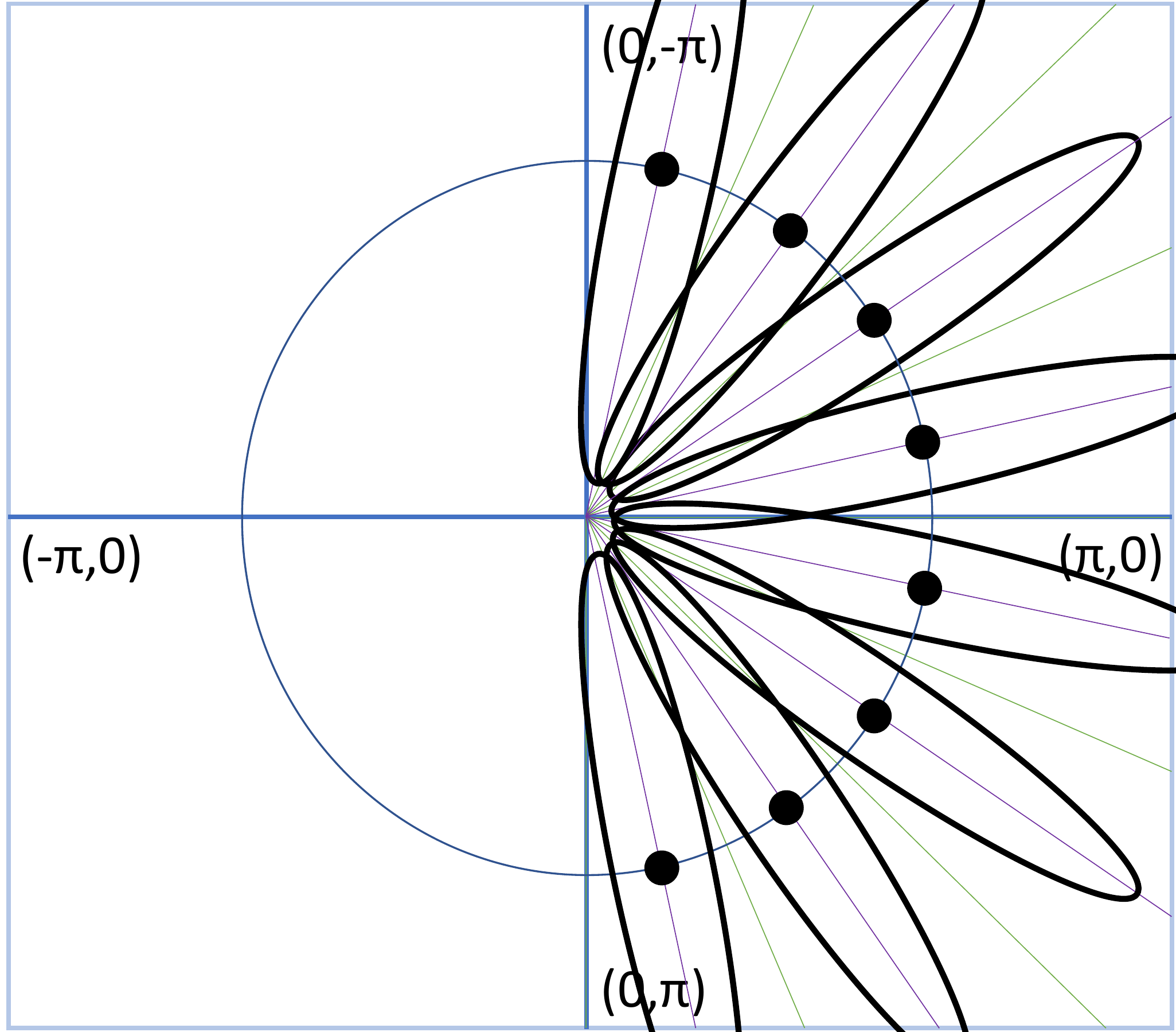}
	\caption{Directional Gabor filterbank design. The example does not show wrap-around artifacts. It also assumes the pre application of the 1D Hilbert Trasnform to suppress frequency magnitude components in the left two quadrants.}
	\label{fig:Level1Level2AMFMfilter}
\end{figure}

\begin{figure}	[bt]
	\includegraphics[width=\columnwidth]{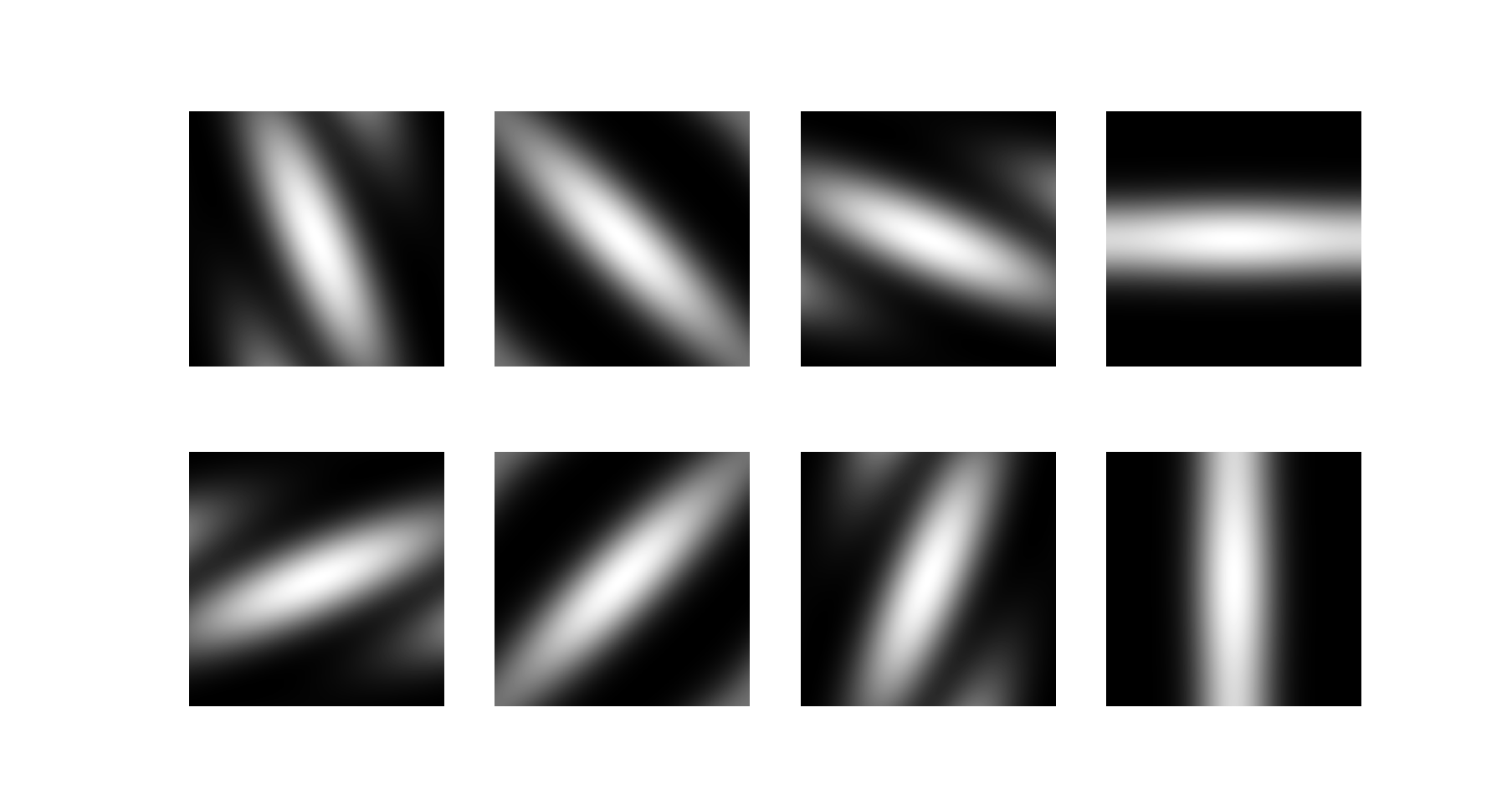}
	\caption{Low-parameter Directional Gaussian implementations using $ 11 \times 11 $ filter coefficients.}
	\label{fig:FFT8kernelsLevel1}
\end{figure}

The following figures depict the FFT magnitude plots of the low-parameter Directional Gabor filterbank. The filters were designed taking into account the implementation in fixed-point hardware. The low parameter implies a small kernel size. Here we used an $ 11 \times 11$ kernel size with $ \sigma_{x} = 1.5 $ and $ \sigma_{y} = 1.5/4 $. Figure \ref{fig:FFT8kernelsLevel1} shows the directional Gabor centered with the same orientations values as described in figure \ref{fig:Level1AMFMfilter}. On the other hand, figure \ref{fig:FFT8kernelsLevel2} shows the second scale of the Gabor filterbank, where wrap around artifacts that would be diminished when the filter is applied to the analytical image.

\begin{figure}	[bt]
	\includegraphics[width=\columnwidth]{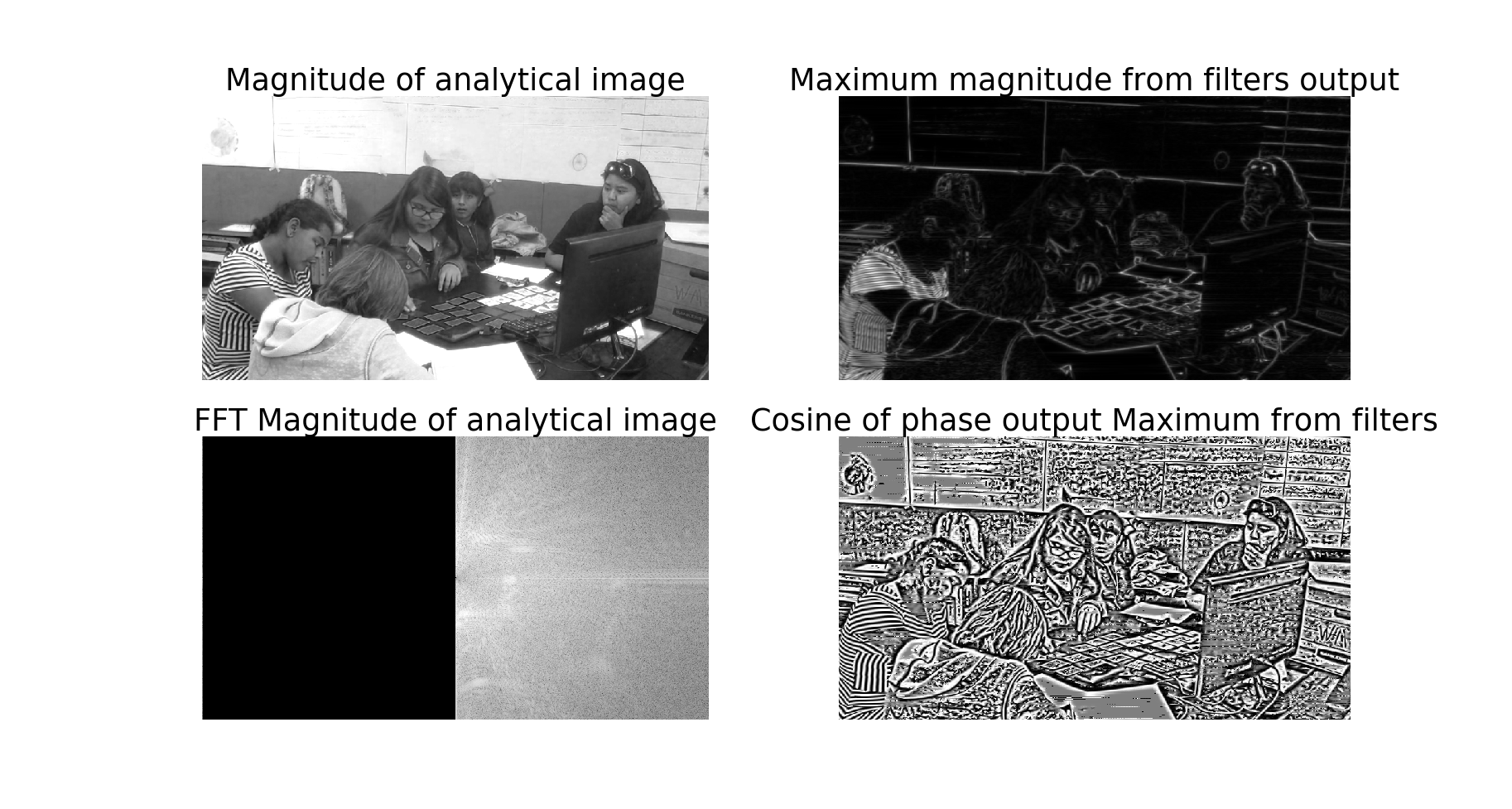}
	\caption{Top-left: Original image. Bottom-left: FFT Magnitude of analytical image. Top-right: IA from AM-FM filterbank using DCA. Bottom-right: FM component using DCA. Undesirable AM components can be seen at the lower-left corner of the IA image.}
	\label{fig:CosPhiHilbert}
\end{figure}

The analytical image is the input for the Gabor filterbanks described here, as it is shown in the lower left subplot of figure \ref{fig:CosPhiHilbert}. Thus, the upper and bottom left quadrants of the frequency spectrum are zero. The subplots on the right show the DCA output. The estimated phase was input to the cosine function to obtain the FM image.

\section{Low-Parameter Block-based Face Detection using AM-FM features}

This section describes the methodology for the Single-Block and Multi-Block regression architecture. The latter one is an extension of the Single-Block using their predicted values as input. Multi-Block regression aims to use the localization of the block to better generalize the blocks that contains faces.

\subsection{Block-based Regression Architecture}

The Single-Block architecture started as an exploration of the impact that the type of input data have over the effectiveness of a CNN architecture. The first architecture selected was the simple and well tested LeNet-5. LeNet-5, despite it's simplicity, has proven to be an effective baseline architecture for regression based on the dominant FM component.

Starting from the LeNet-5 as the baseline architecture, hyper-parameter tuning was used to reduce over-fitting. 
Without sacrificing performance, the number of convolutional layers was reduced from 3 to 1, followed by an increased Max Pooling size of $ 5 \times 5$, reduced-parameter fully connected layers, augmented by an additional fully connected layer. The final architecture is given in table \ref{tab:reducedlenetTable}. 

\begin{table}[H]
	\centering
	
	\begin{center}
		\begin{tabular}{||c c c c c c c||} 
			\hline
			Layer & Type & Maps & Size & Kernel Size & Stride & Activation \\ [0.5ex] 
			\hline\hline
			In & Input & 1 & 50x50 & - & - & - \\ 
			\hline
			C1 & Convolution & 6 & 46x46  & 5x5 & 1 & selu \\			
			\hline
			S2 & Max Pooling & 6 & 23x23  & 5x5 & 2 & selu \\
			\hline
			F3 & Fully Connected & - & 40  & - & - & selu \\
			\hline
			F4 & Fully Connected & - & 24  & - & - & selu \\
			\hline
			Output & Fully Connected & - & 1 & - & - & sigmoid \\ [1ex] 
			\hline			
		\end{tabular}
	\end{center}
	
	\caption{ Block-based regression architecture for face detection. The input block is of size $ 50 \times 50 $.}
	\label{tab:reducedlenetTable}
\end{table}

To compare against the proposed approach, we consider the MobileNet V2 for Single-Block regression. The model was imported to the Keras framework and then customized to meet the design approach used in the Reduced LeNet-5. 

The top layer was modified to be a sigmoid for percentage prediction. The input size was changed to $ 50 \times 50 $. Since the original size was $ 225 \times 225 \times 3$, $\alpha$ was set to 0.7 for the gradual reduction of the network. The weights were randomly initialized.

\subsection{Multi-Block Regression Architecture}

Is important to mention that the blocks where randomly shuffled during the training of the Single-Block regressor. 
Hence, no location information was used by the Single-Block regression architecture. To take advantage of block locations, a Multi-Block regression architecture was designed to process the outputs from each block.

The architecture of this Multi-Block Neural Network was inspired by the 2 fully connected layers of the Single-Block CNN. The input and output parameters were determined by the number of blocks in each frame. The architecture accepts and input of 45 Single-Block regression values and generates predictions for each block (also 45). The middle layers use roughly the same number of neurons as the previous CNN. The architecture is given in table \ref{tab:MLPTable}.

\begin{table}[H]
	\centering
	
	\begin{center}
		\begin{tabular}{||c c c c c c c||} 
			\hline
			Layer & Type & Maps & Size & Kernel Size & Stride & Activation \\ [0.5ex] 
			\hline\hline
			In & Input & 1 & 5x9x1 & - & - & - \\ [1ex] 
			\hline
			F1 & Fully Connected & - & 60  & - & - & selu \\
			\hline
			F2 & Fully Connected & - & 40  & - & - & selu \\
			\hline
			Output & Fully Connected & - & 45 & - & - & sigmoid \\ 
			\hline
		\end{tabular}
	\end{center}
	
	\caption{Multi-Block regression architecture.}
	\label{tab:MLPTable}
\end{table}

\begin{figure}	[H]
	\includegraphics[width=\columnwidth]{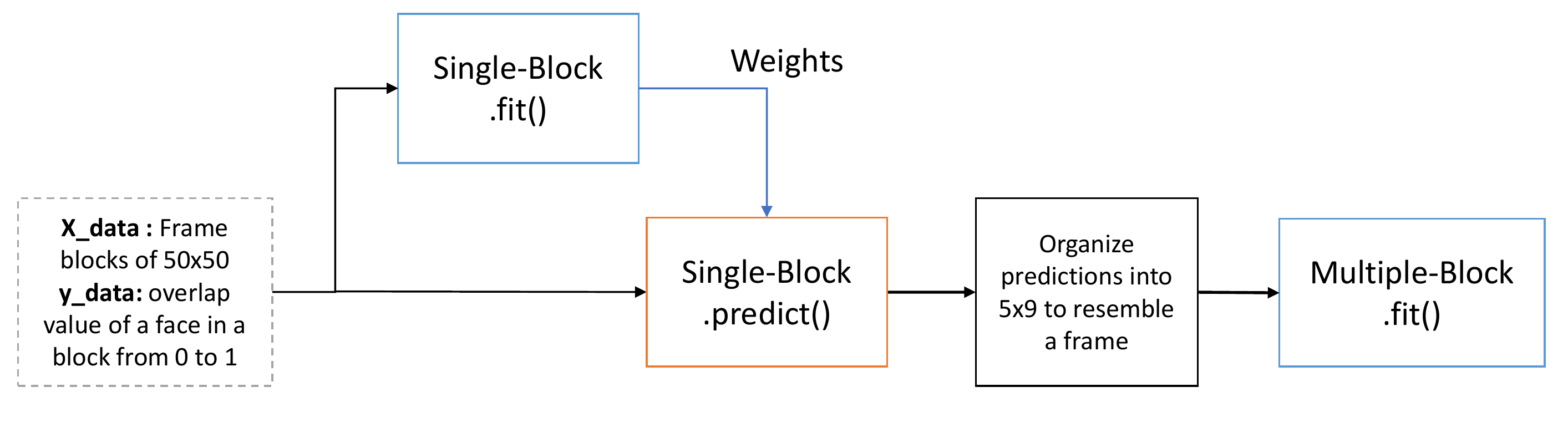}
	\caption{Training of Multi-Block regression architecture.}
	\label{fig:cnnBlockdiagram}
\end{figure}

The training of the Multi-Block Neural Network is based on the predicted values of the Single-Block CNN. The training process is shown in figure \ref{fig:cnnBlockdiagram}.

\chapter{ Results }
In this chapter, the results of running the Single-Block and Multi-Block regression architectures are presented for three different types of input: (i) Original gray scale image processed with LeNet-5, and the proposed architecture using DCA  estimates for: (ii) FM image, (iii) the IA image, (iv) IA and FM images. Refer to section \ref{aolmeDataset} for a description of the AOLME dataset, and the breakdown of the training, validation and testing sets.

To visualize the results, we present examples of: True Positives (TP), False Positives (FP) and False Negatives (FN) mark over the blocks. The results are color coded as follows: green dot for a True Positive, red dot for a False Positive and yellow dot for False Negative. To determine TP, FP, TN and FN, a threshold for the ground truth continuous values had to be defined. Several thresholds were tested in correspondence with the range values of the output predictions.

The Receiver Operating Characteristic (ROC) plots along with the AUC value are presented to compare the performance between to input data types. The FM image is compared to both the original and the AM-FM combined input. The AUC gives a reliable metric to compare both performances.

Loss and AUC plots are presented for the three different data inputs. The Loss function is the objective function to be minimized by the optimization algorithm in the CNN. The AUC is a metric that varies from 0.5 for a useless predictor and 1.0 for an ideal one. 
Also the MobileNet V2 results are shown for comparisons in performance and computation time.

\section{Face-Detection using Single-Block Classification}


Several frames were used for debugging the block-based face detection. In the following figures the selected frame will show the type of data as well the block markings for a threshold value of 0.15. Here, it is important to mention that a higher value will reduce the False Positive rates, but most often, the outcome will be the absence of any result, since the FP and TP regression values are close to each other. For this reason, the threshold was set to a point where enough TPs are displayed without predicting the majority of blocks as faces, having too many FPs.

\begin{figure}	[bt]
	\includegraphics[width=\columnwidth]{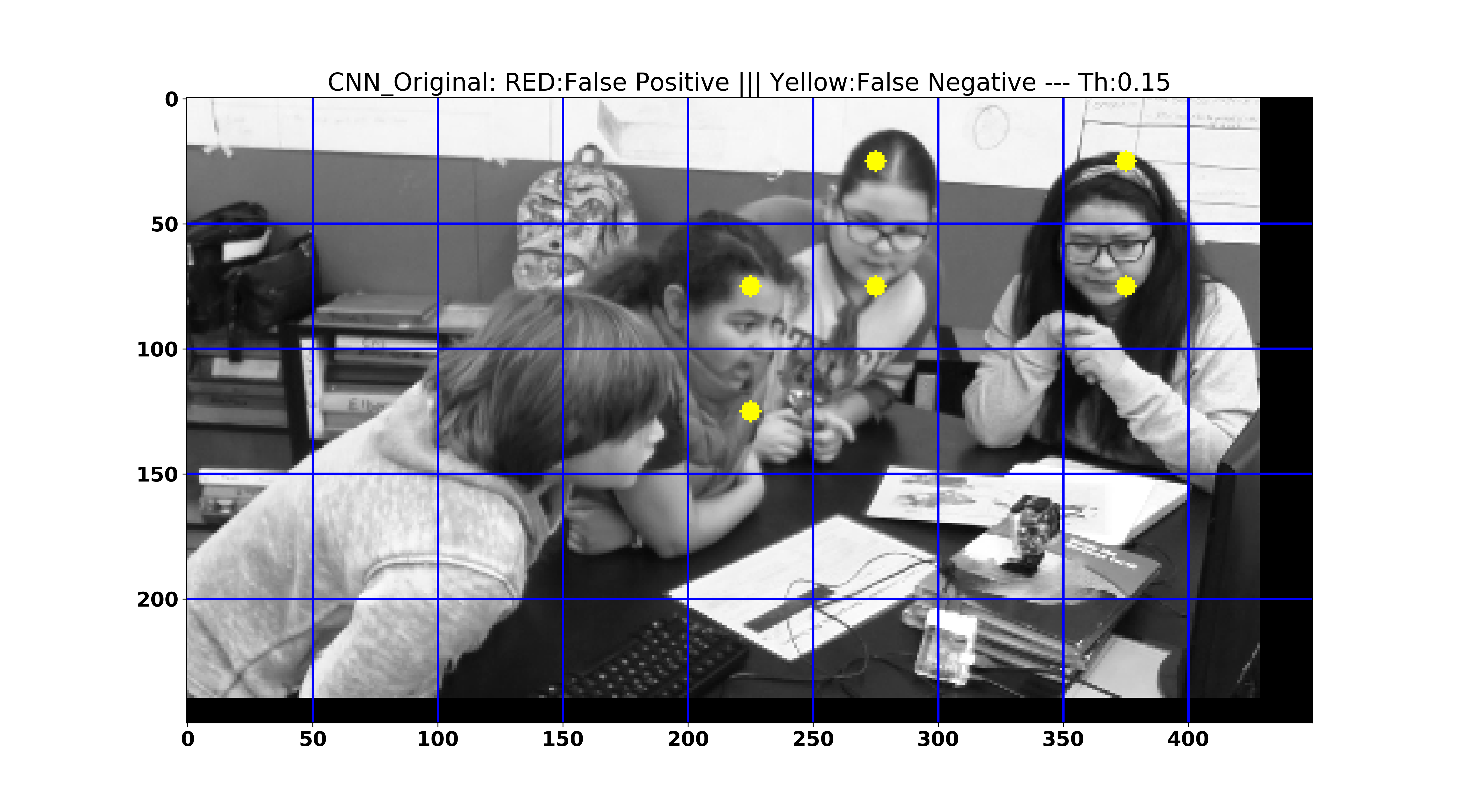}
	\caption{Single-Block regression: Original grayscale frame with marks for TP (green), FP (red) and FN (yellow). LeNet-5 did not work with the original grayscale image.}
	\label{fig:frame_lenet_original}
\end{figure}

\begin{figure}	[bt]
	\includegraphics[width=\columnwidth]{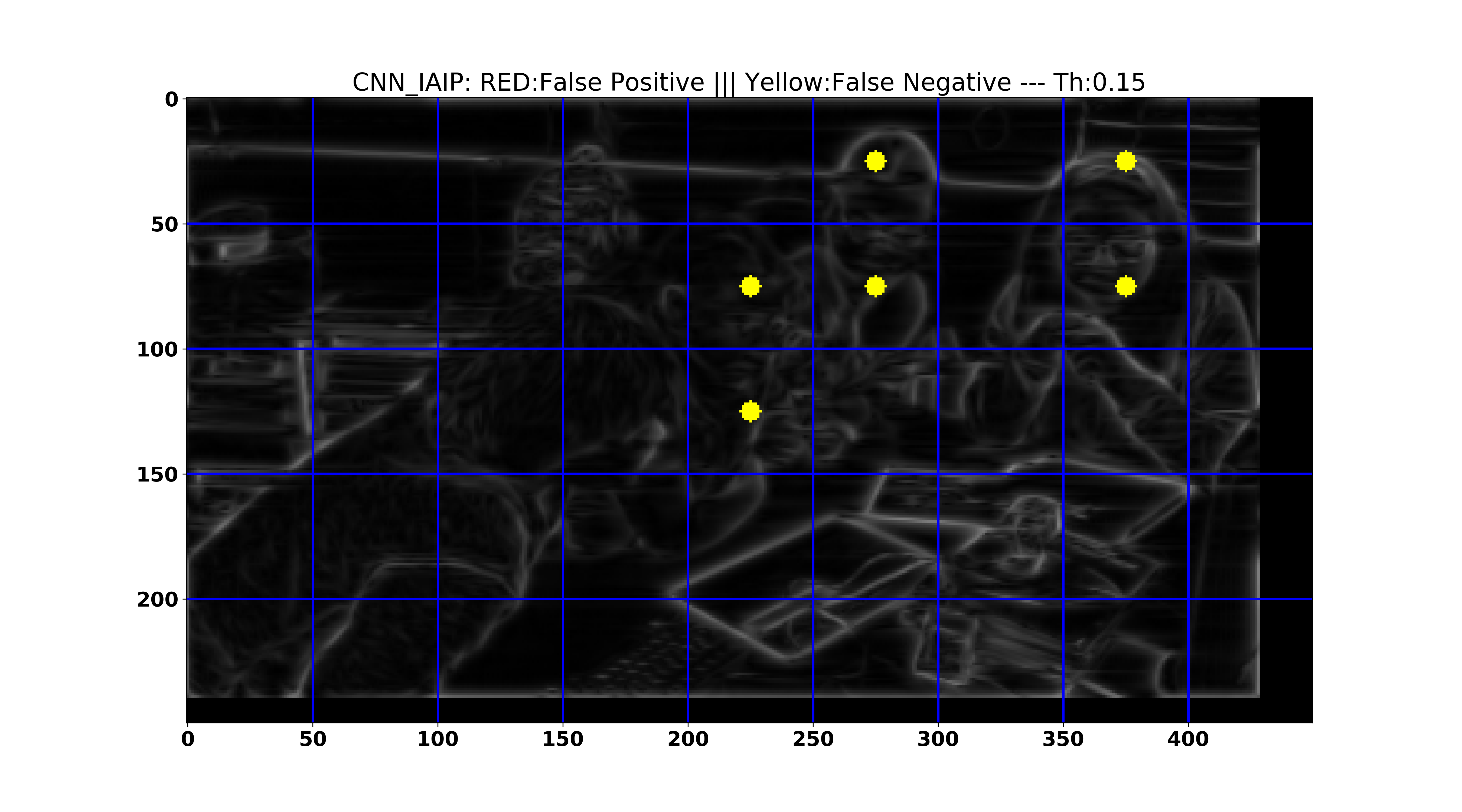}
	\caption{Single-Block regression: AM-FM frame with marks for TP (green), FP (red) and FN (yellow).}
	\label{fig:frame_lenet_IAIP}
\end{figure}

\begin{figure}	[bt]
	\includegraphics[width=\columnwidth]{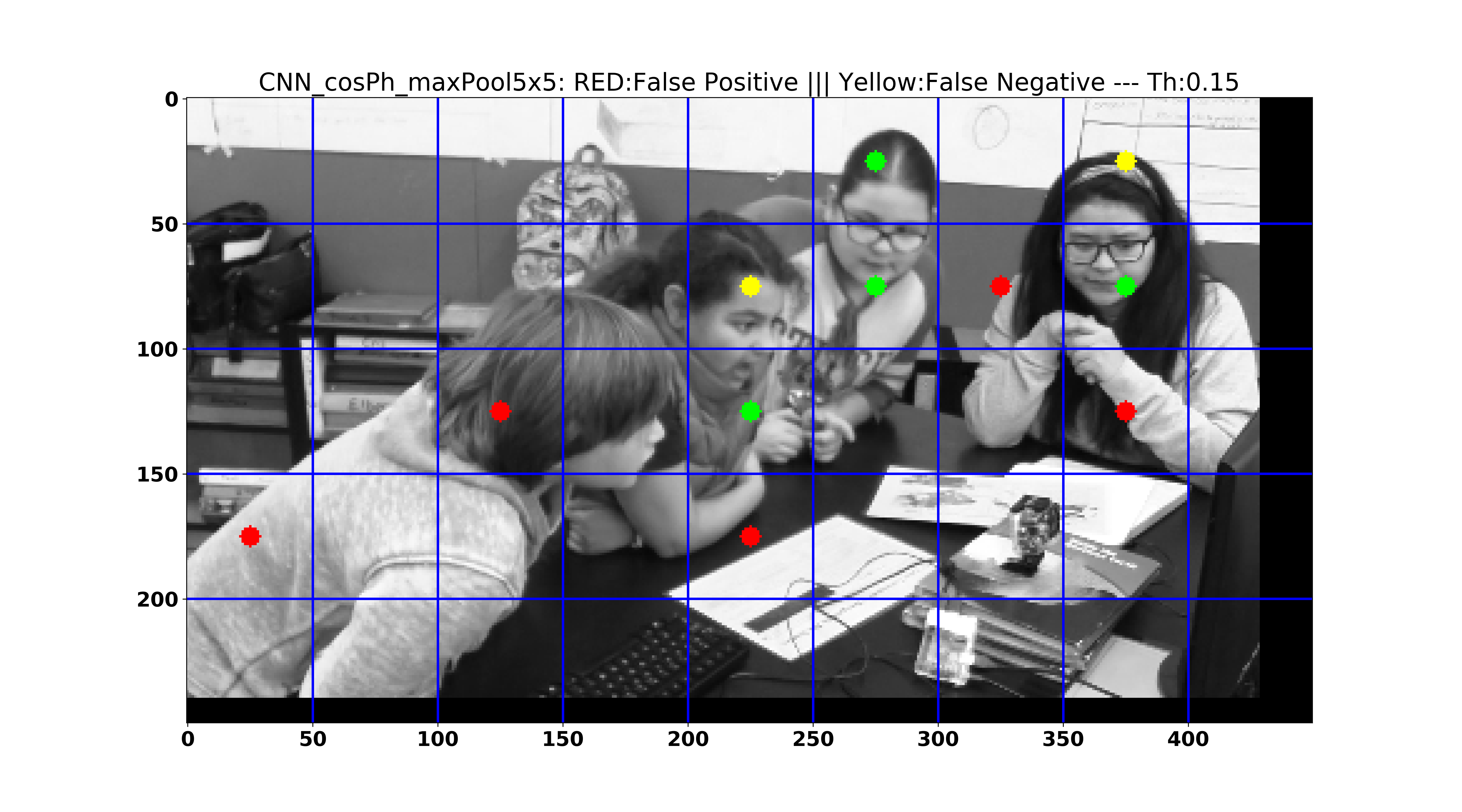}
	\caption{Single-Block regression: FM frame with marks for TP (green), FP (red) and FN (yellow).}
	\label{fig:frame_lenet_cosPhi}
\end{figure}

Starting with the original image in grayscale, figure \ref{fig:frame_lenet_original} shows no face predictions in any of the 45 blocks in the frame. Same behavior is encountered in figure \ref{fig:frame_lenet_IAIP}, regardless of the given threshold value. The results from the FM input image demonstrated that the proposed method managed to find 2 faces along with several False Positives (red marks), presented in figure \ref{fig:frame_lenet_cosPhi}.

\begin{figure}	[bt]
	\includegraphics[width=\columnwidth]{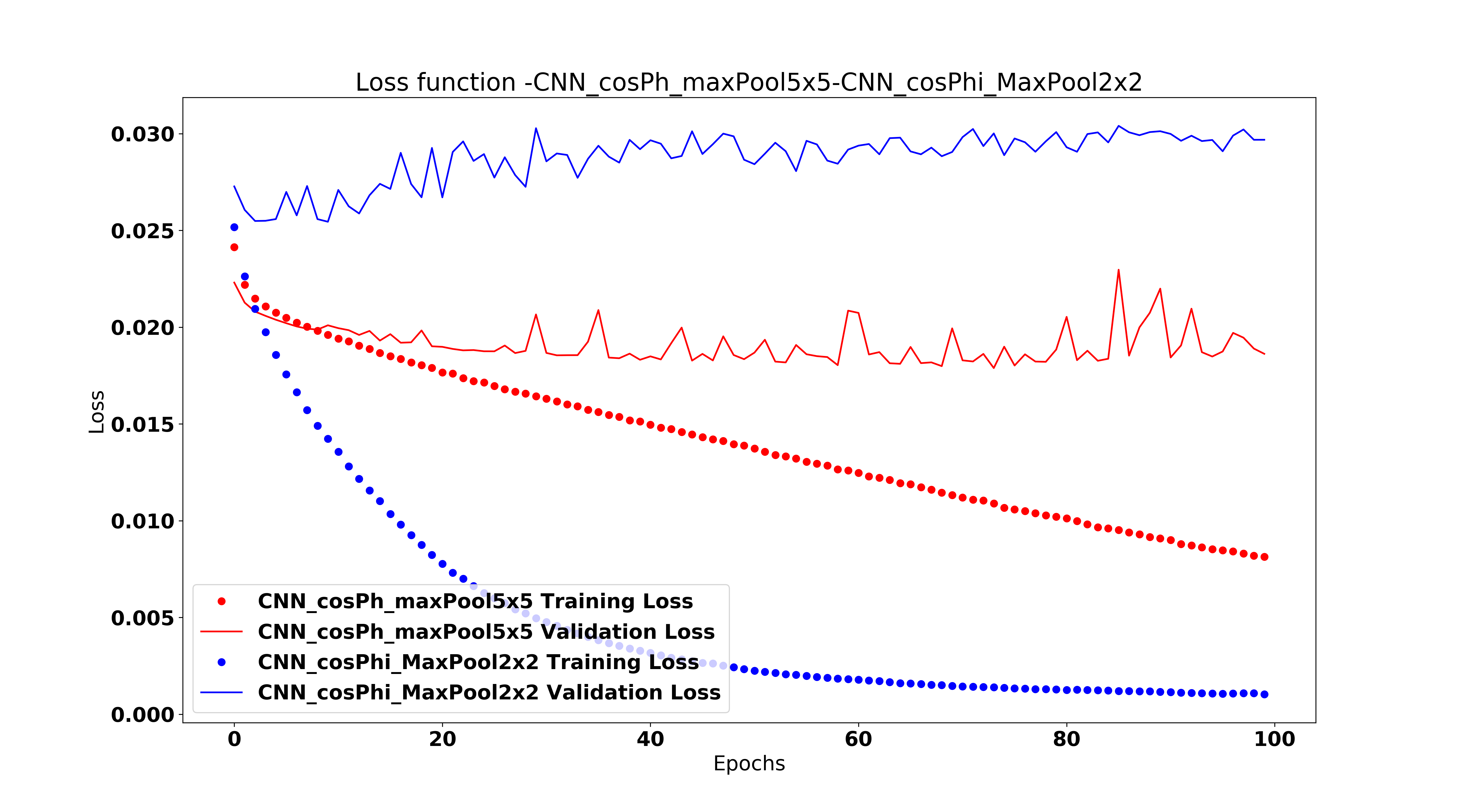}
	\caption{Loss function results during the Single-Block architecture tunning. Different values of stride at the max pooling layer are presented in blue and red for stride=2  and stride=5 respectively.}
	\label{fig:loss_overfitting_cos}
\end{figure}

The loss function plot per epoch is presented to facilitate a better understanding of the learning process. Figure \ref{fig:loss_overfitting_cos} summarizes the experiments conducted for reducing the over fitting by reducing the network complexity. The Loss values in blue corresponds to an architecture using a max pooling layer of $stride=2$ , while the red corresponds to $stride = 5$. There is a clear difference between the training loss versus the validation loss. This difference is greater with the $ stride=2 $, hence motivating an increase to the max pooling stride to the value used in the system: $stride=6$ .  

\begin{figure}	[bt]
	\includegraphics[width=\columnwidth]{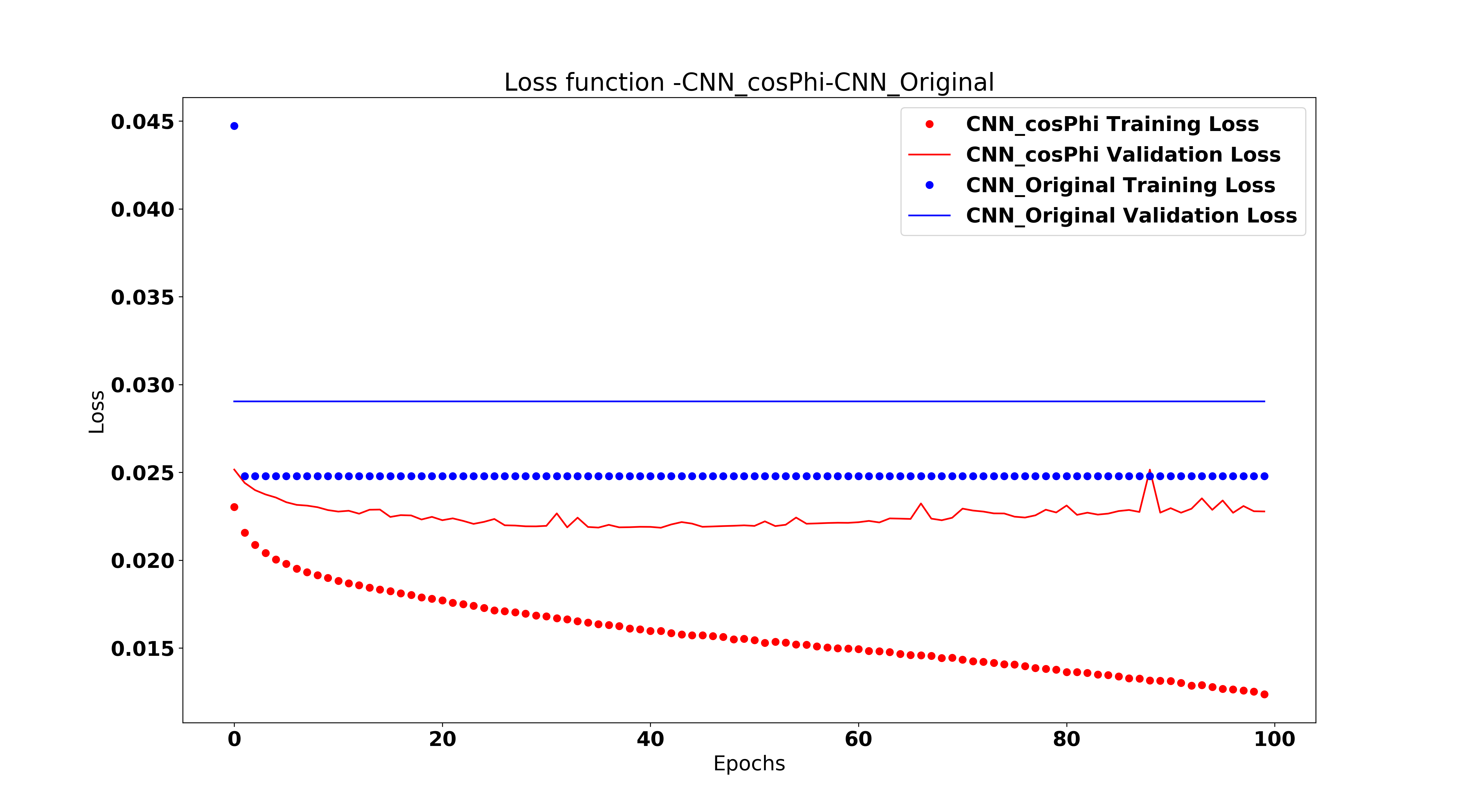}
	\caption{Single-Block training results: Loss function values for the original image and the FM component.}
	\label{fig:loss_lenet_original}
\end{figure}

\begin{figure}	[bt]
	\includegraphics[width=\columnwidth]{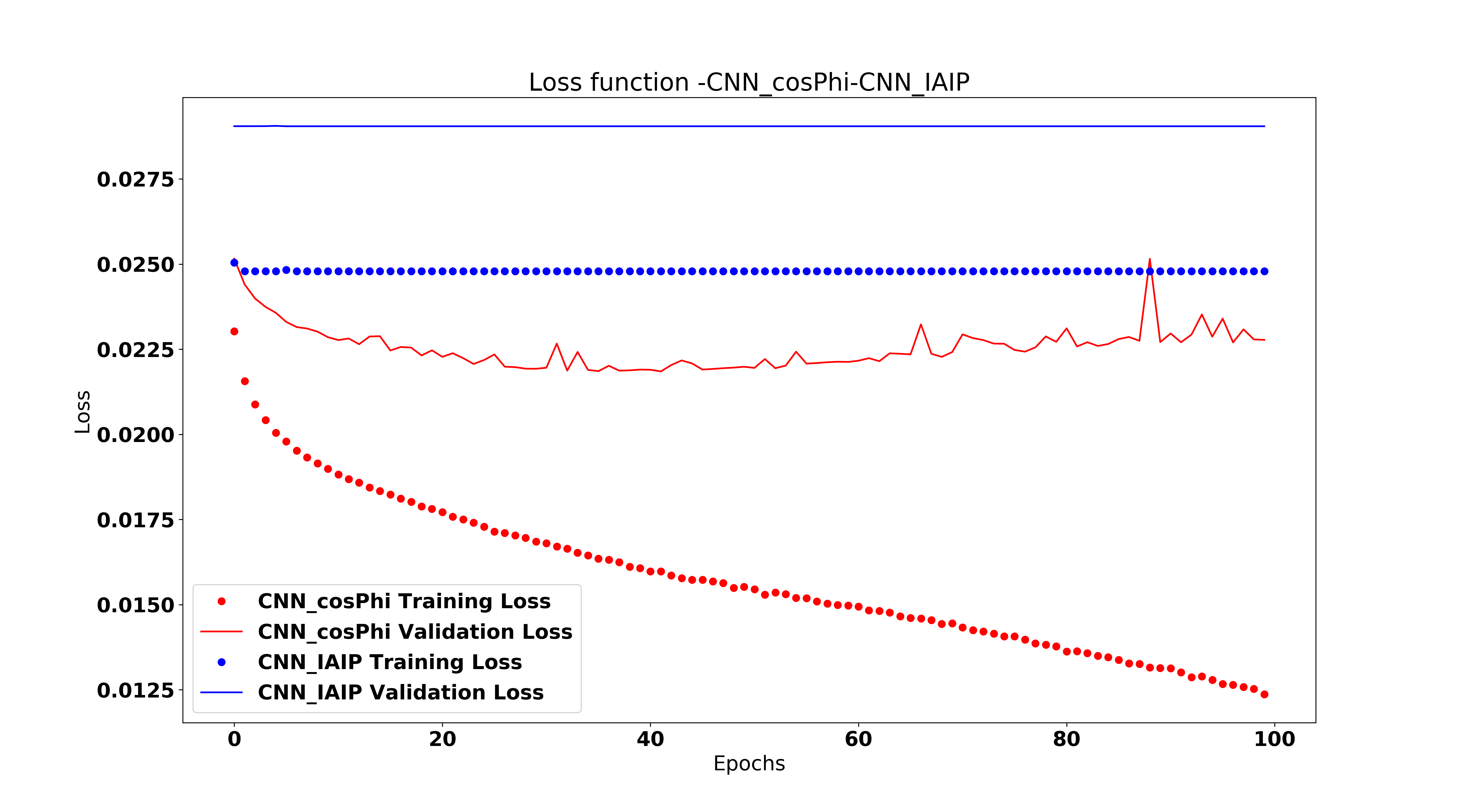}
	\caption{Single-Block training results: Loss  function values for the AM-FM component and the FM component. }
	\label{fig:loss_lenet_IA}
\end{figure}

Figures \ref{fig:loss_lenet_original} and \ref{fig:loss_lenet_IA} support the results from threshold experiment (figures \ref{fig:frame_lenet_original} and \ref{fig:frame_lenet_IAIP}). No learning is happening with the original grayscale image or the combination of the IA and FM image. In both figures, both of them having the same training and validation loss (in blue) of 0.025 and 0.028 respectively. In contrast, the FM image produced clear learning results at each epoch, with less difference between training and validation than in figure \ref{fig:loss_overfitting_cos}.

\begin{figure}	[bt]
	\includegraphics[width=\columnwidth]{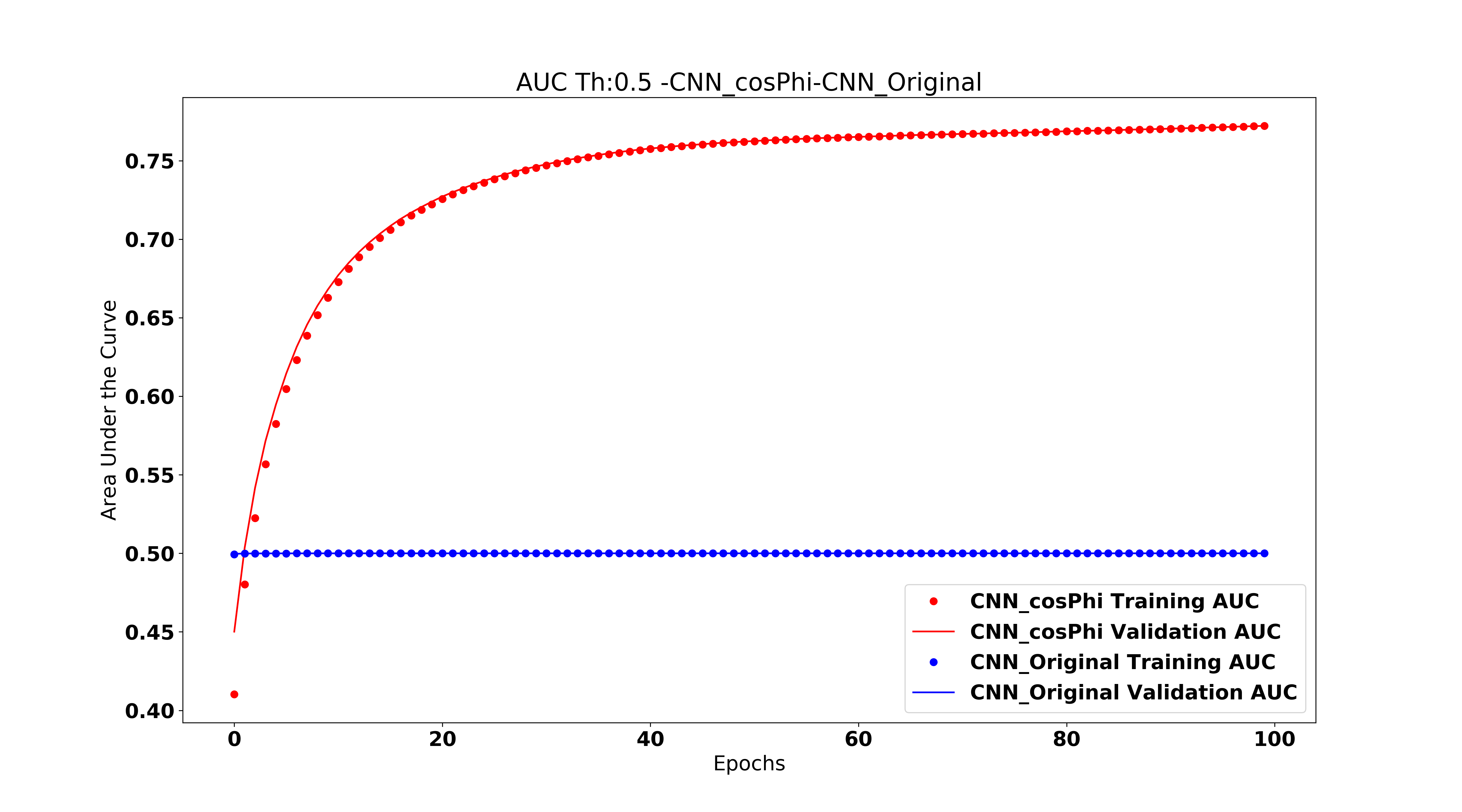}
	\caption{Single-Block: AUC at each epoch for FM component (in red) and original image (in blue).}
	\label{fig:auc_lenet_original}
\end{figure}

\begin{figure}	[bt]
	\includegraphics[width=\columnwidth]{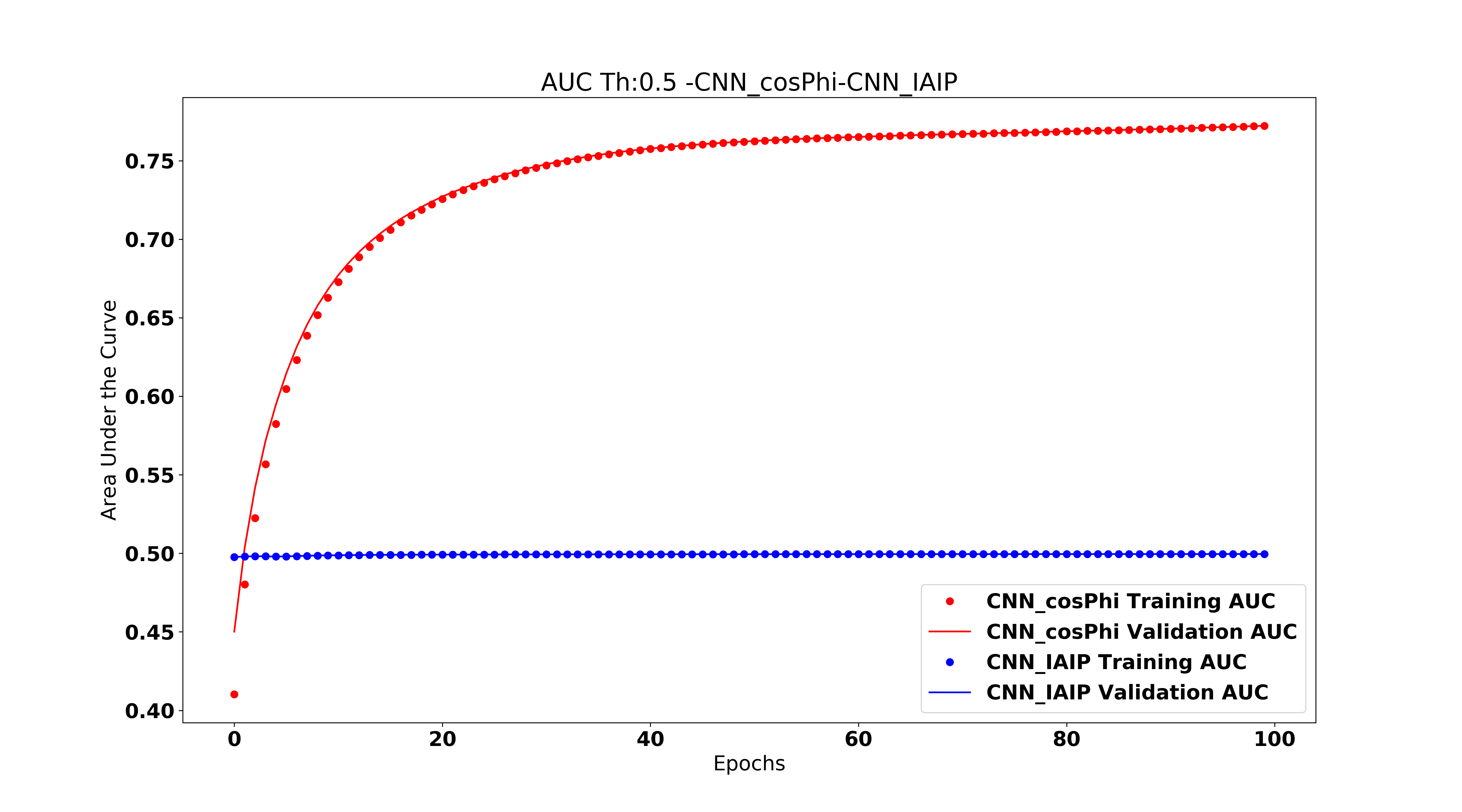}
	\caption{Single-Block: AUC at each epoch for FM component (in red) and AM-FM component (in blue).}
	\label{fig:auc_lenet_IA}
\end{figure}

The results for the Area Under the Curve (AUC) value at each epoch are shown in figures \ref{fig:auc_lenet_original} and \ref{fig:auc_lenet_IA}. The selection of this metric involves the unbalanced nature of the dataset, since the blocks with a face percentage comprises around 10\% of the total of blocks in the frame. Then, any dummy predictor set to "background" will encounter 90\% of accuracy. It was important to come up with a better metric to determine the success of the predictions. Both plots were included to show the similar behavior of random predictor when using the original image of the AM-FM. For the FM component, the AUC was constantly increasing, reaching 80\% of their final value at approximately 5 epochs.

The AUC plot was obtained using Keras callbacks. In this framework, the automatic threshold used to generate the AUC value was set to 0.5, because the ROC calculation requires the ground truth value to be boolean and the input to be continuous. In despite of this, the stranded AUC at 0.5 indicates no learning is happening for the original or AM-FM data.

The following figures \ref{fig:roc_cosPhi_original} and \ref{fig:roc_cosPhi_IAIP} demonstrate the importance of the phase component in block-based face detection using CNNs. The ROC for the original grayscale image describes a random predictor, not capable of differentiating between the features of the images. In contrast, the phase does give a functional predictor to start exploring the face detection methods. Similar results are presented in figure \ref{fig:roc_cosPhi_IAIP} the AM-FM components against the FM components alone.

\begin{figure}	[bt]
	\includegraphics[width=\columnwidth]{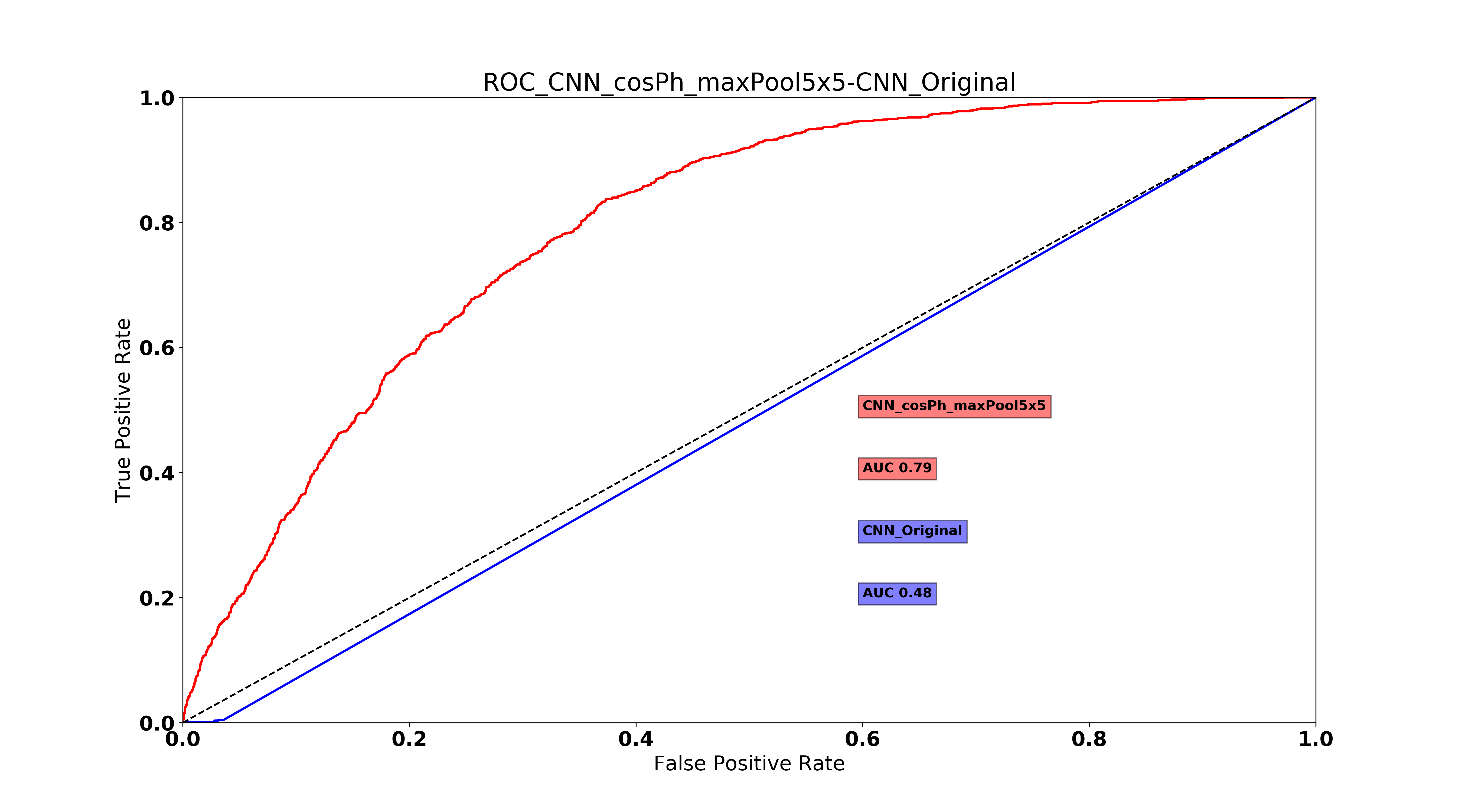}
	\caption{Single-Block: ROC curve for FM component (in red) and original image (in blue).}
	\label{fig:roc_cosPhi_original}
\end{figure}

\begin{figure}	[bt]
	\includegraphics[width=\columnwidth]{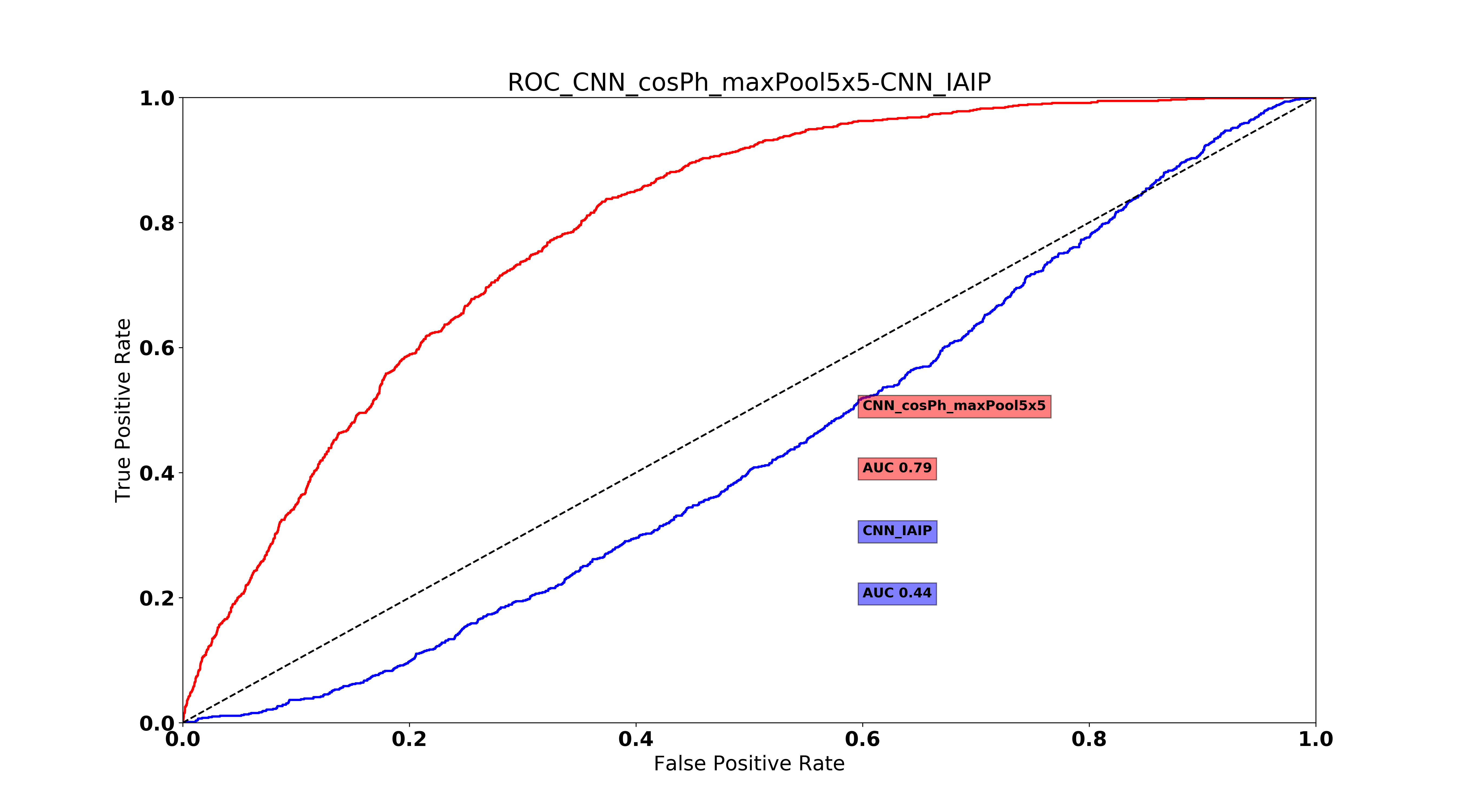}
	\caption{Single-Block: ROC curve for FM component (in red) and AM-FM component (in blue).}
	\label{fig:roc_cosPhi_IAIP}
\end{figure}

\section{Face Detection by Combining Regional Classifiers }

\begin{figure}	[bt]
	\includegraphics[width=\columnwidth]{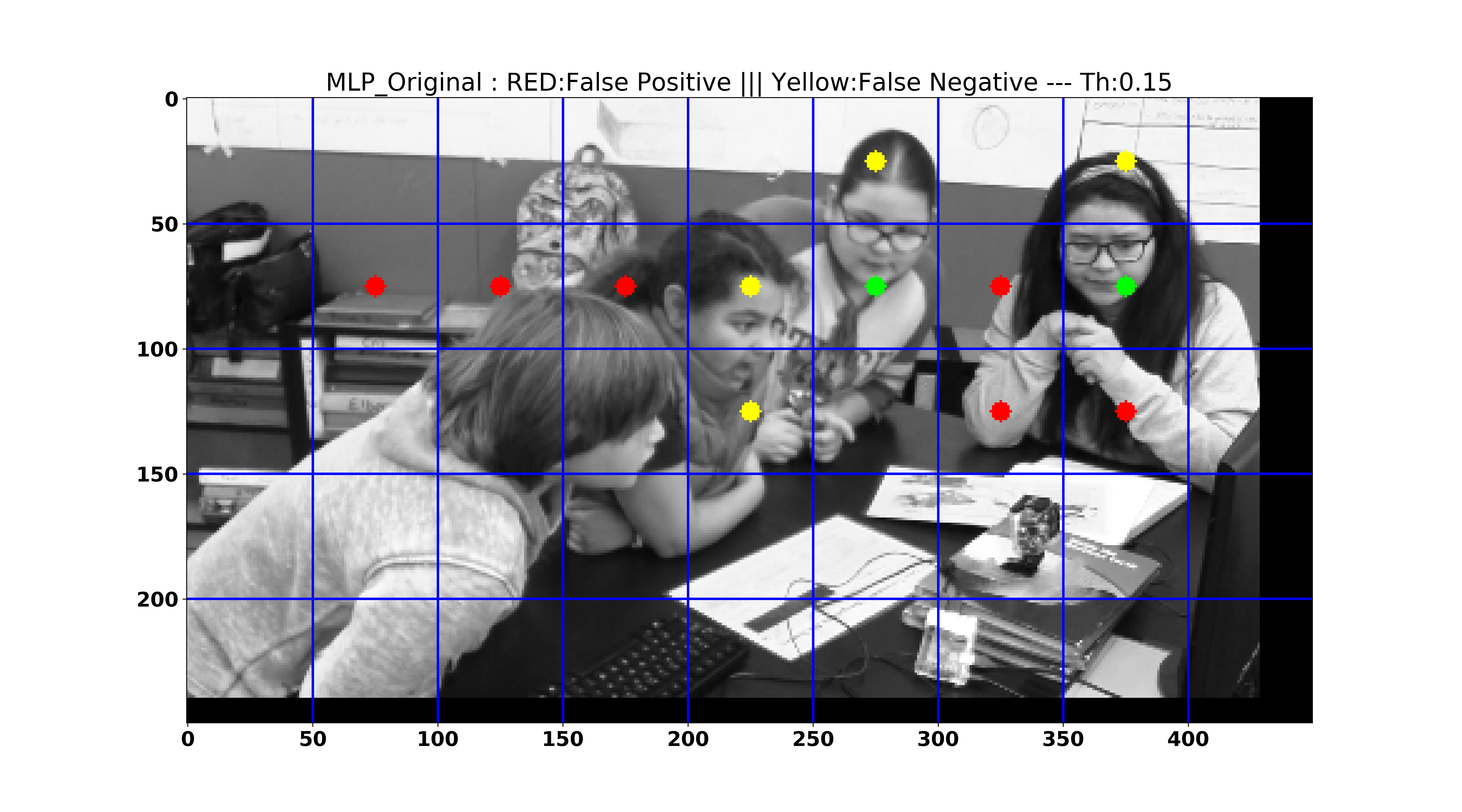}
	\caption{Multi-Block regression: Original grayscale frame with marks for TP (green), FP (red) and FN (yellow).}
	\label{fig:MLP_frame_lenet_original}
\end{figure}

\begin{figure}	[bt]
	\includegraphics[width=\columnwidth]{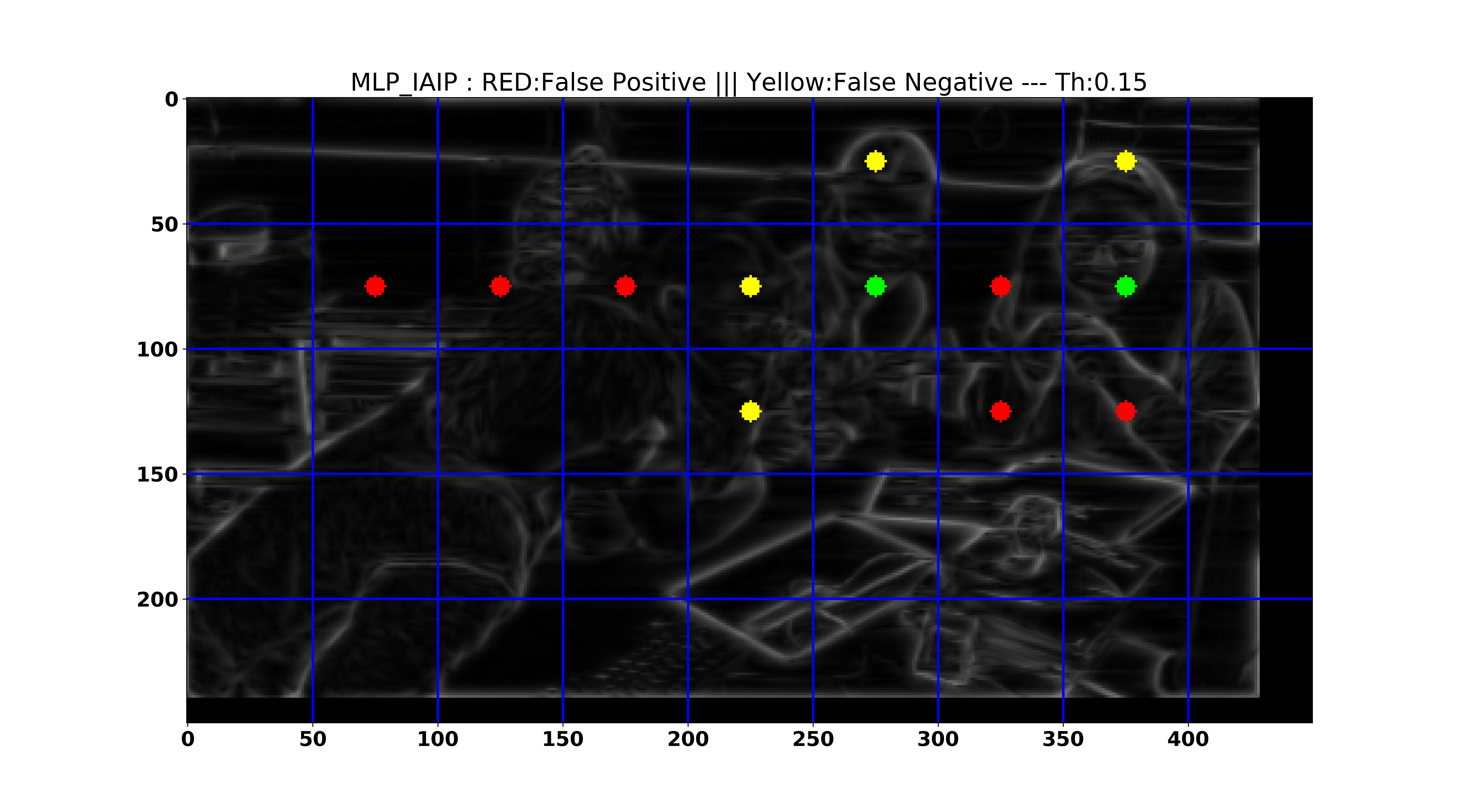}
	\caption{Multi-Block regression: AM-FM frame with marks for TP (green), FP (red) and FN (yellow).}
	\label{fig:MLP_frame_lenet_IAIP}
\end{figure}

\begin{figure}	[bt]
	\includegraphics[width=\columnwidth]{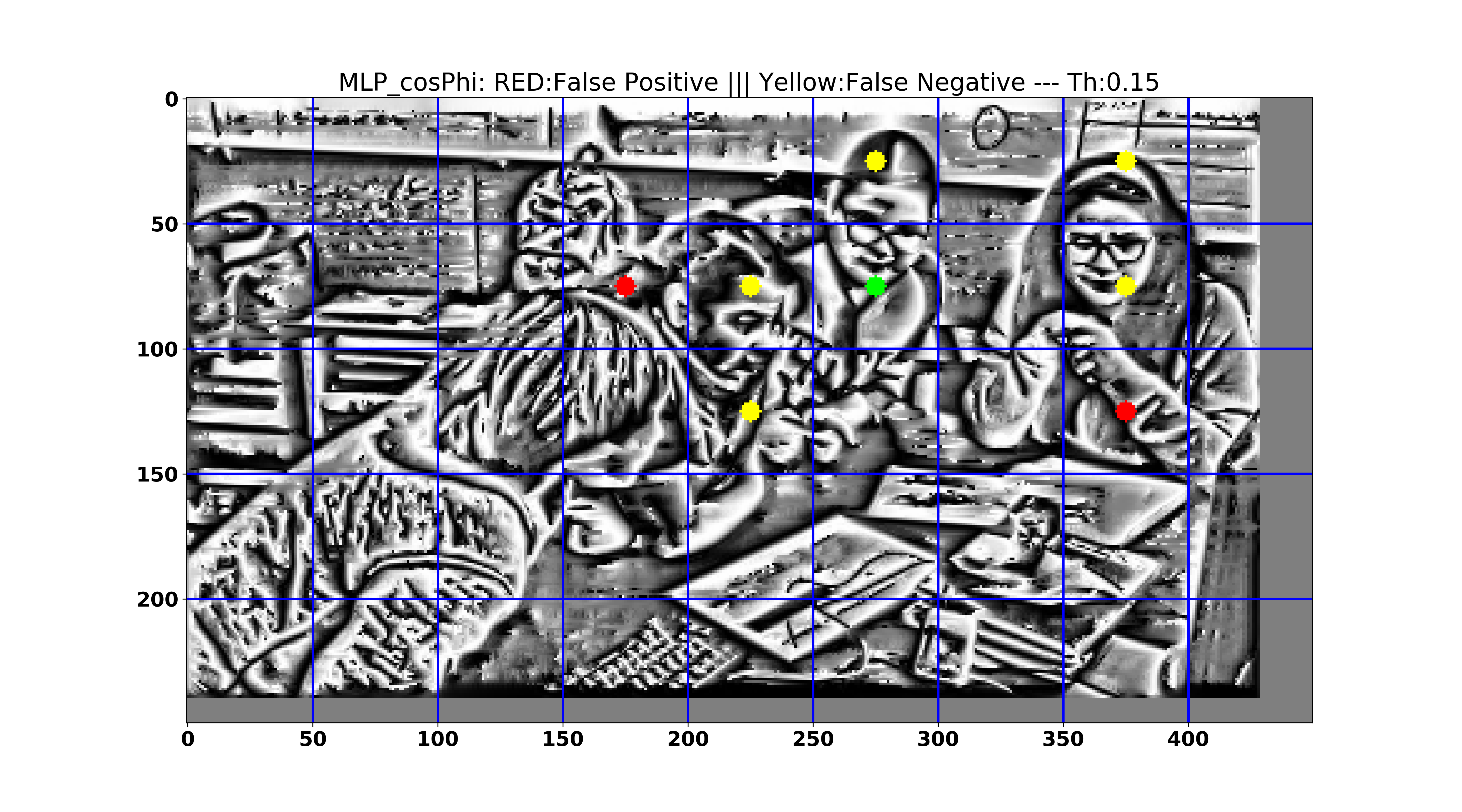}
	\caption{Multi-Block regression: FM frame with marks for TP (green), FP (red) and FN (yellow).}
	\label{fig:MLP_frame_lenet_cosPhi}
\end{figure}

This section summarize the results for Multi-block regression. The marks for True Positive, False Positive and False Negative are presented for the three input data types in figures \ref{fig:MLP_frame_lenet_original}, \ref{fig:MLP_frame_lenet_IAIP} and \ref{fig:MLP_frame_lenet_cosPhi}. 

From these results, the first 2 frames predict the same blocks. Furthermore, the same blocks were predicted as faces for several other frames. For results not based on the FM images, random-like predictions were generated by the Single-Block system, and were input to the training of the Multi-block regressor. The fully connected layers trained to generalize the data using random-like inputs, selected the blocks where the faces were most commonly found. In the case of the Multi-block regressor using the FM component data, the input predictions were far more accurate. Figure \ref{fig:MLP_frame_lenet_cosPhi} shows three predicted blocks as faces. From them, one was successfully found and the two others were close to the actual faces.

\begin{figure}	[bt]
	\includegraphics[width=\columnwidth]{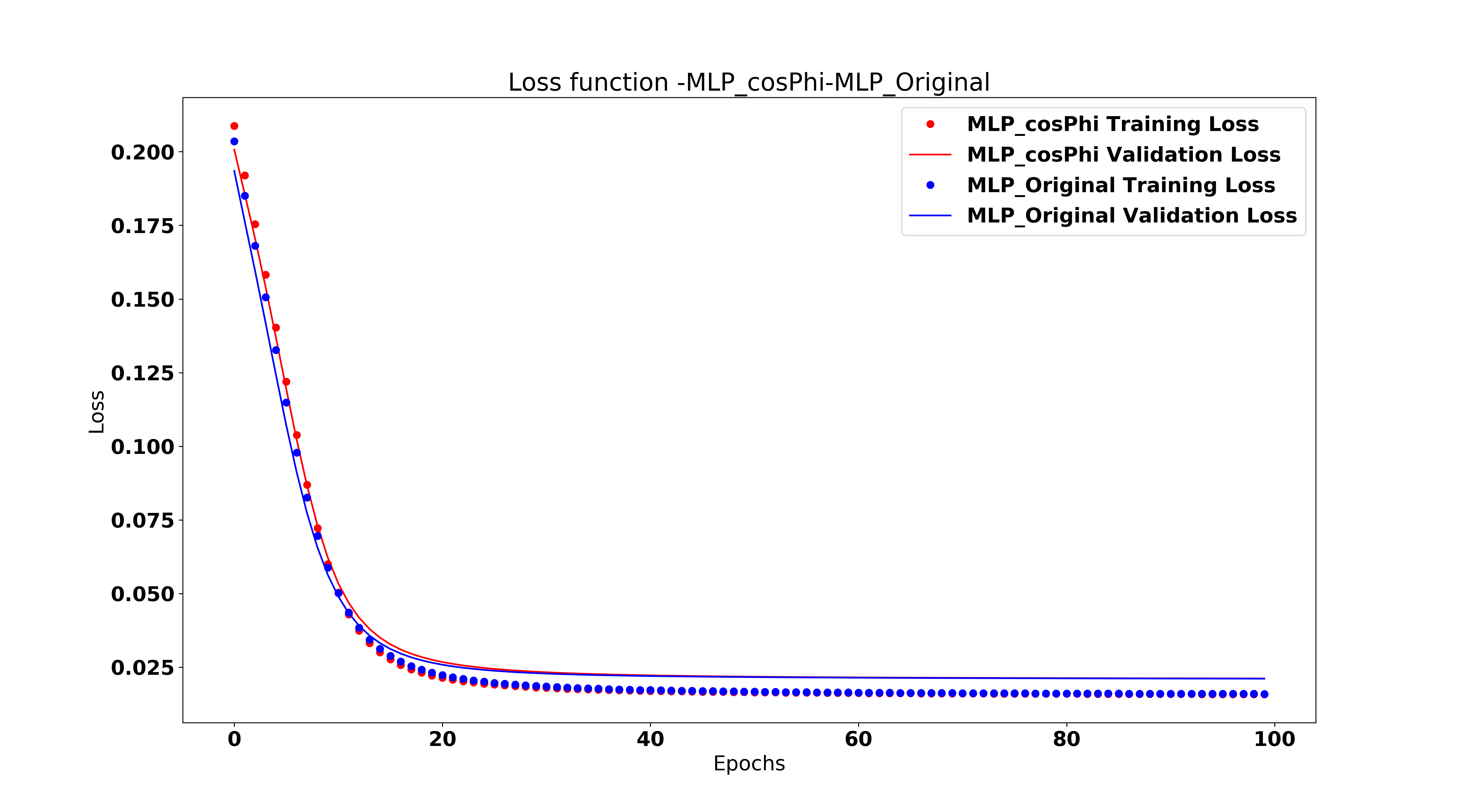}
	\caption{Multi-Block training results: Loss  function values for the original image and the FM component.}
	\label{fig:MLP_loss_lenet_original}
\end{figure}

\begin{figure}	[bt]
	\includegraphics[width=\columnwidth]{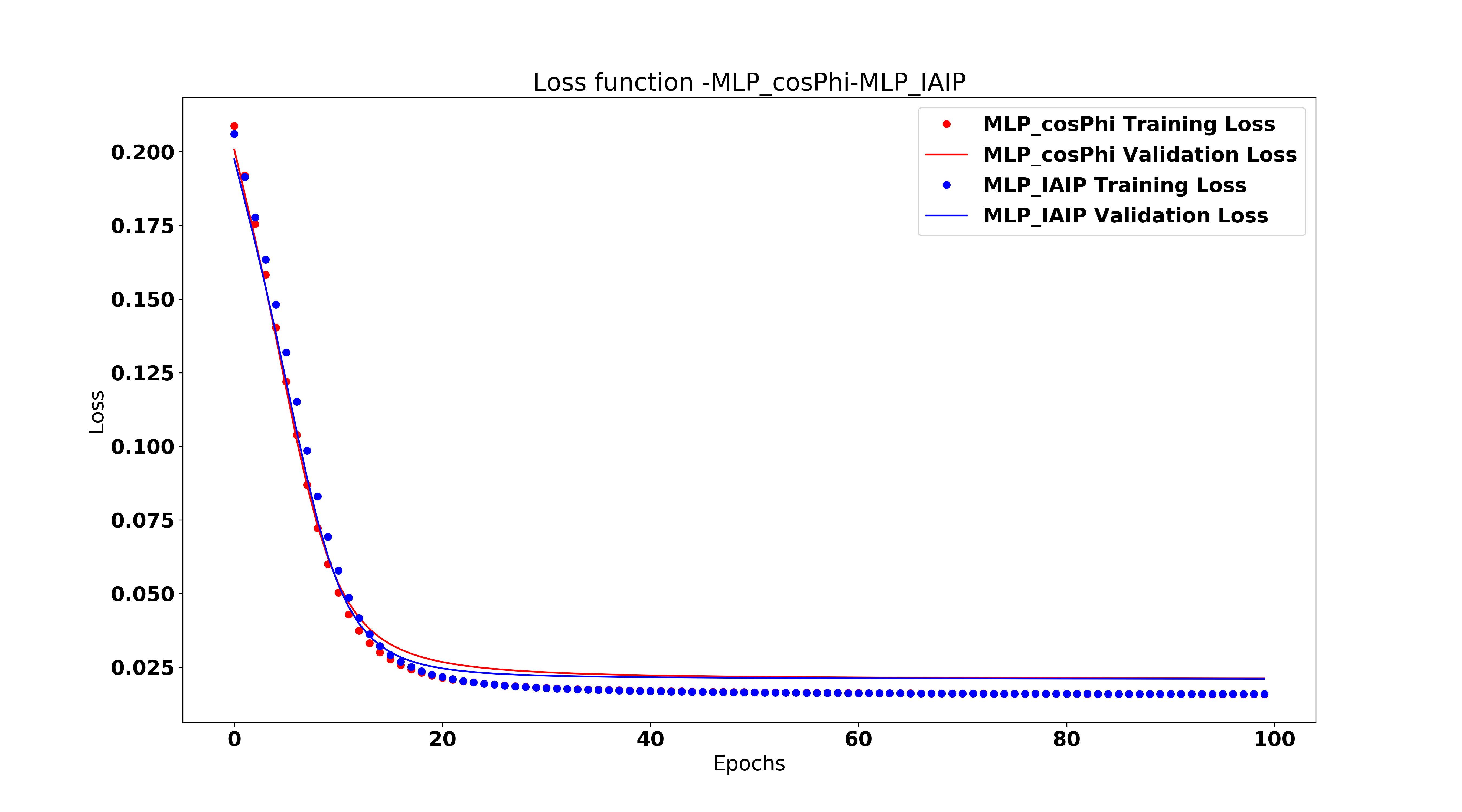}
	\caption{Multi-Block training results: Loss  function values for the AM-FM component and the FM component.}
	\label{fig:MLP_loss_lenet_IA}
\end{figure}

The generalization achieved with the original and AM-FM systems led to a loss function that resembles the loss function for the FM component (see figures \ref{fig:MLP_loss_lenet_original} and \ref{fig:MLP_loss_lenet_IA}). In the first case, there is a higher number of False Positives due to the broad selection of blocks as faces. While in the FM case, the prediction was done using the previous information, thus obtaining different results for each frame. 

\begin{figure}	[bt]
	\includegraphics[width=\columnwidth]{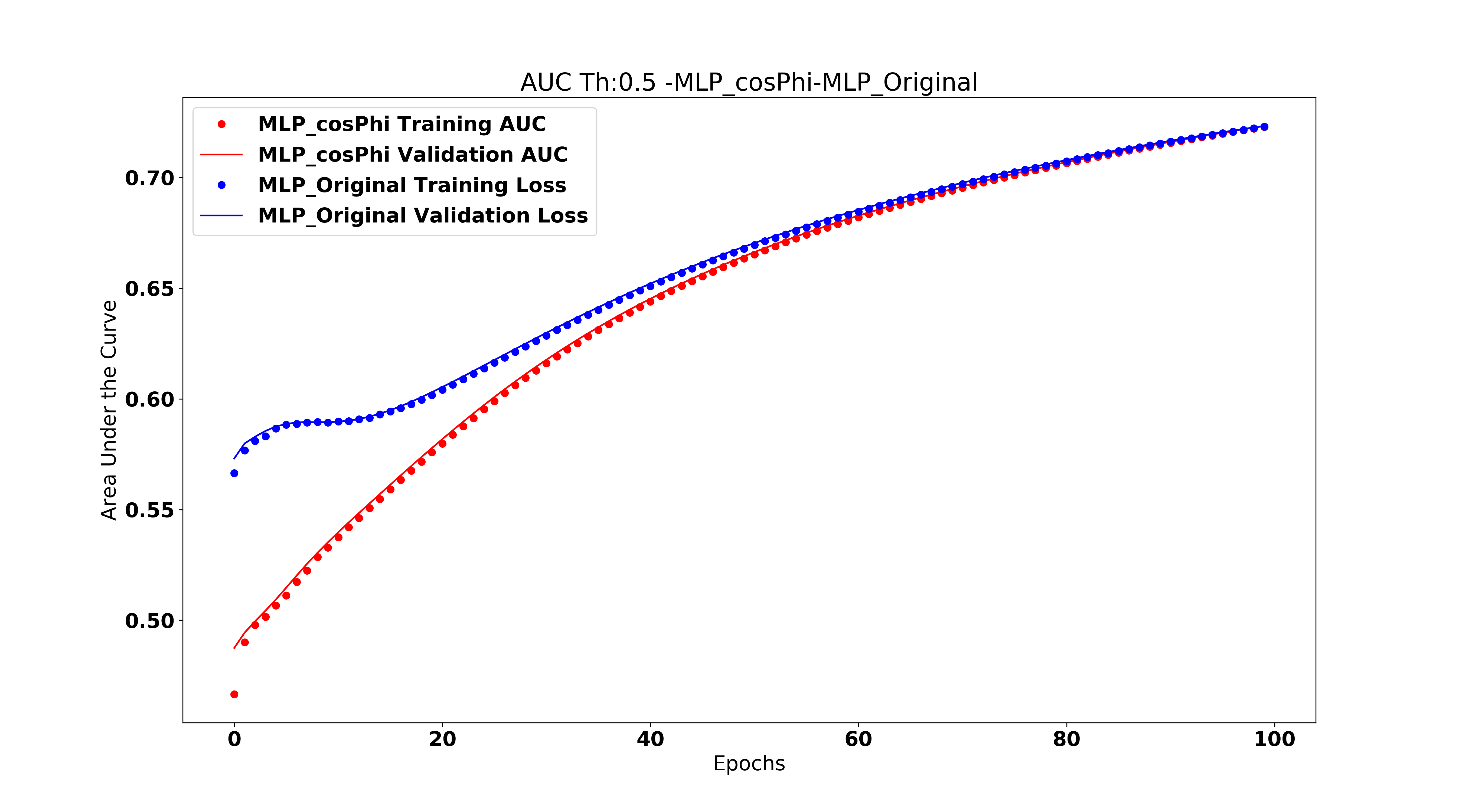}
	\caption{Multi-Block training results: AUC at each epoch for FM component (in red) and original image (in blue).}
	\label{fig:MLP_auc_lenet_original}
\end{figure}

\begin{figure}	[bt]
	\includegraphics[width=\columnwidth]{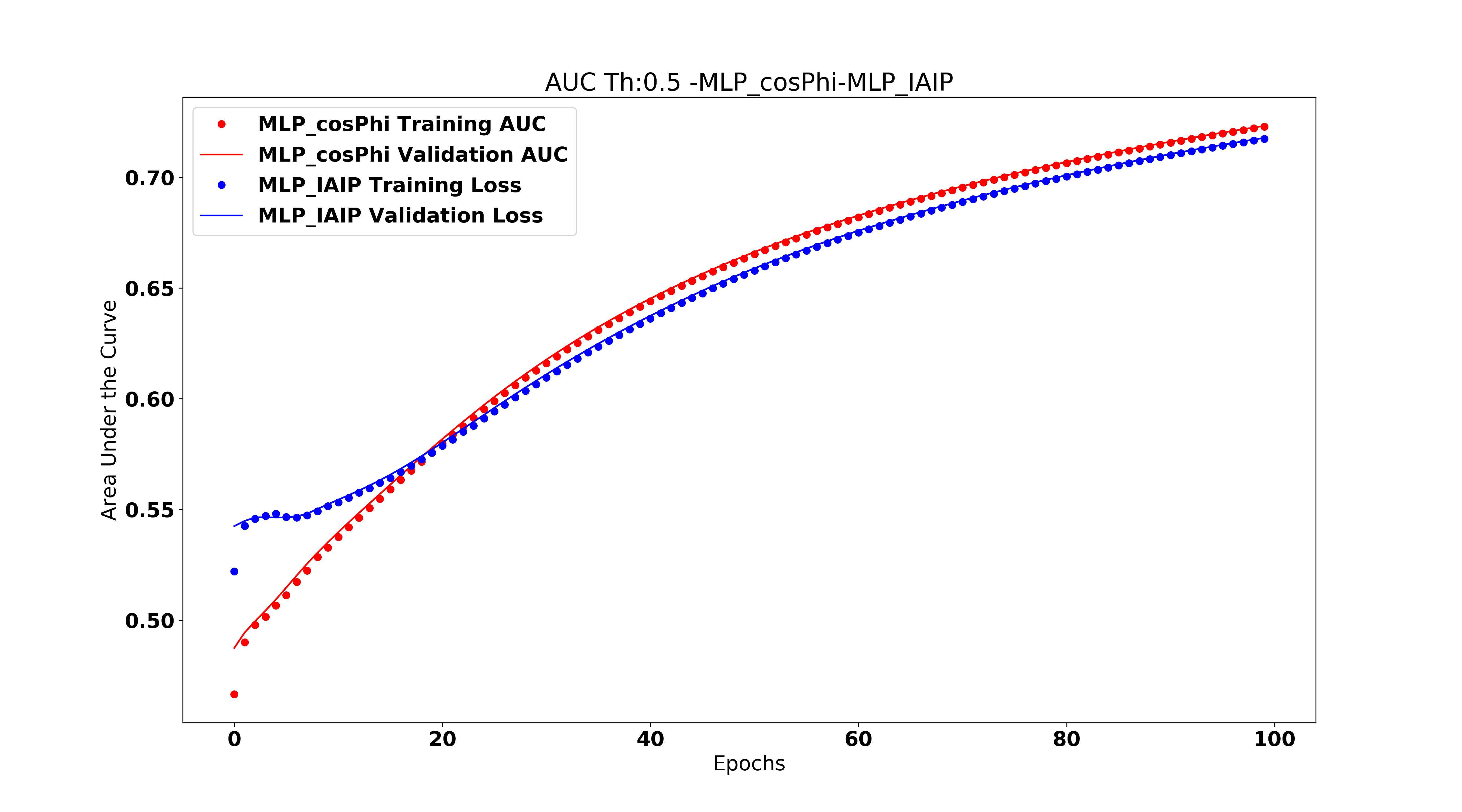}
	\caption{Multi-Block training results: AUC at each epoch for FM component (in red) and AM-FM component (in blue).}
	\label{fig:MLP_auc_lenet_IA}
\end{figure}

The following figures depict the evolution of the AUC per epoch. The original and AM-FM plots are similar in the evolution of the AUC values (see figures \ref{fig:MLP_auc_lenet_original} and \ref{fig:MLP_auc_lenet_IA}). The starting point is near 0.5 of a random predictor and later it increases as it generalizes for the whole dataset. They also have an irregular growth due to the several adjustments the network had to perform. Frames came from different groups and sessions, thus giving slight variations where the faces are often located. For the case of the FM component, the growth was uniform reaching the same AUC values.

\begin{figure}	[bt]
	\includegraphics[width=\columnwidth]{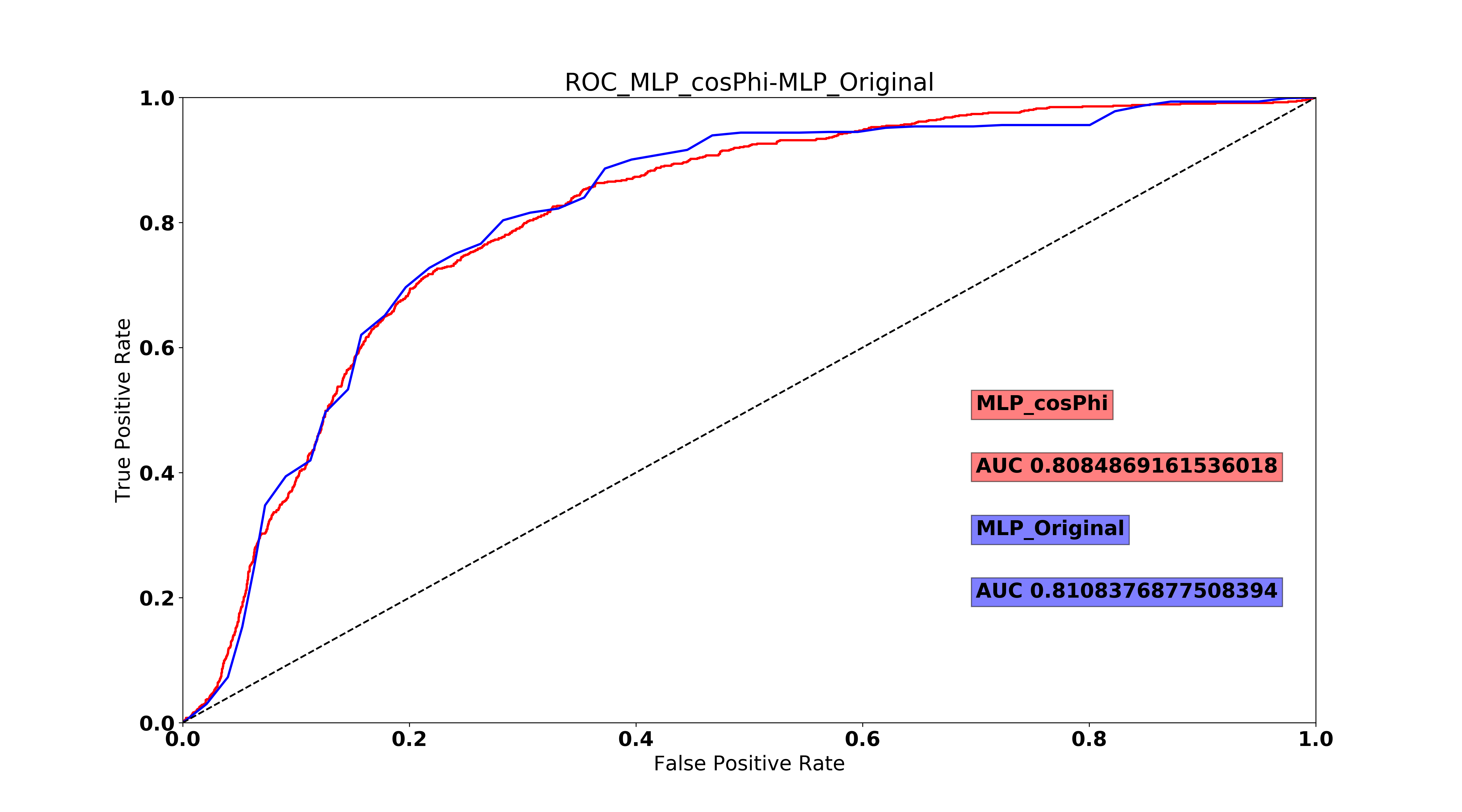}
	\caption{Multi-Block: ROC curve for FM component (in red) and original image (in blue). Unlike the popular belief that lowet layers learn ipmortant features, the plot illustrates that higher layer that learn spatial locations in stationary scenes can correct the results from lower layers.}
	\label{fig:MLP_roc_cosPhi_original}
\end{figure}

\begin{figure}	[bt]
	\includegraphics[width=\columnwidth]{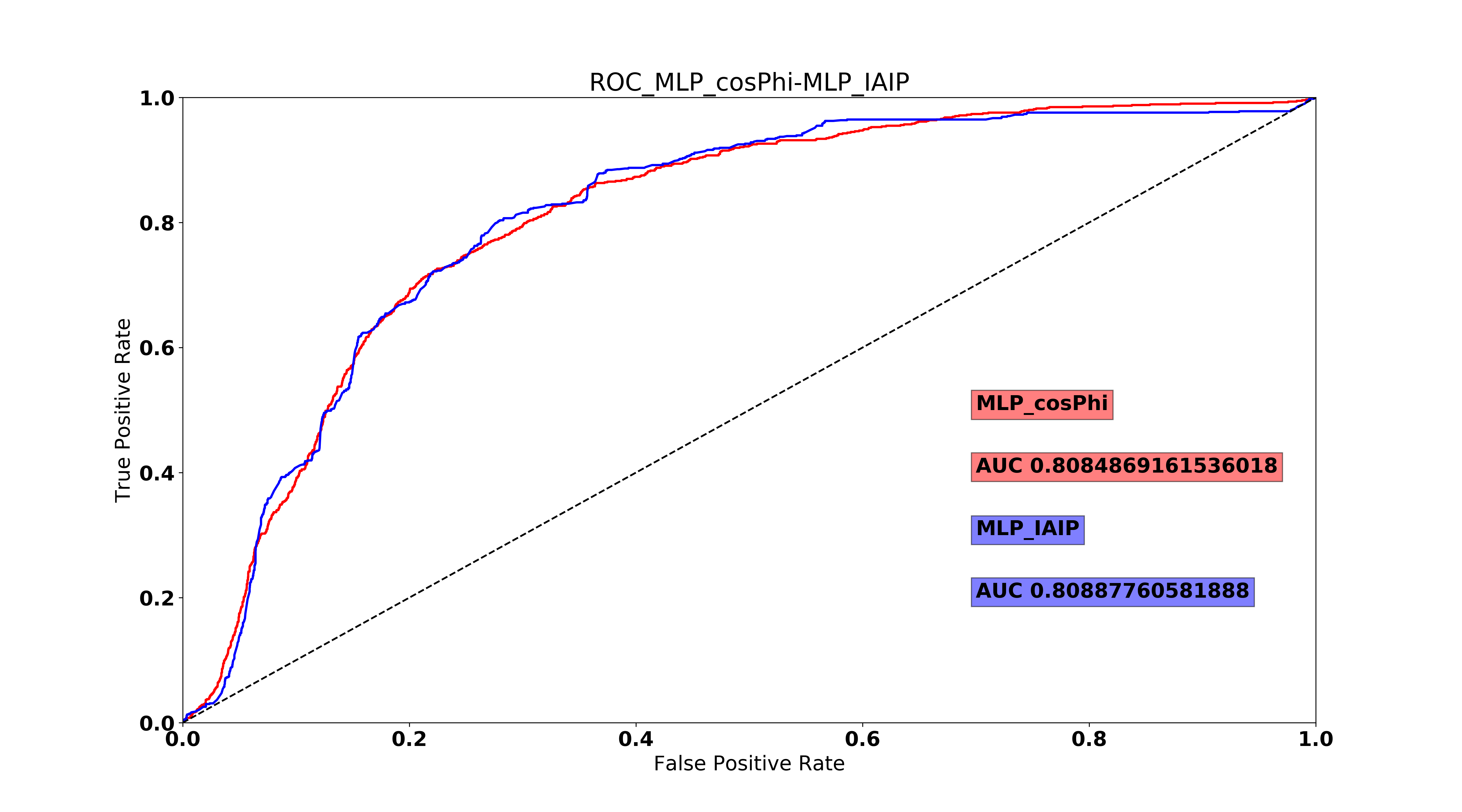}
	\caption{Multi-Block: ROC curve for FM component (in red) and AM-FM component (in blue).}
	\label{fig:MLP_roc_cosPhi_IAIP}
\end{figure}

The results in figures \ref{fig:MLP_roc_cosPhi_original} and \ref{fig:MLP_roc_cosPhi_IAIP} provide extensions of the previous figures. In terms of testing, the AUC values from the FM component and the original - AM-FM components are similar. 

\section{Results for MobileNetV2 }

MobileNetV2 network was tested for the single-block regression process. The following discussion compares the performance from the original image input and the FM component. Also, the training time for MobileNet V2 is compared against the training times for the proposed architectures. 

\begin{figure}	[bt]
	\includegraphics[width=\columnwidth]{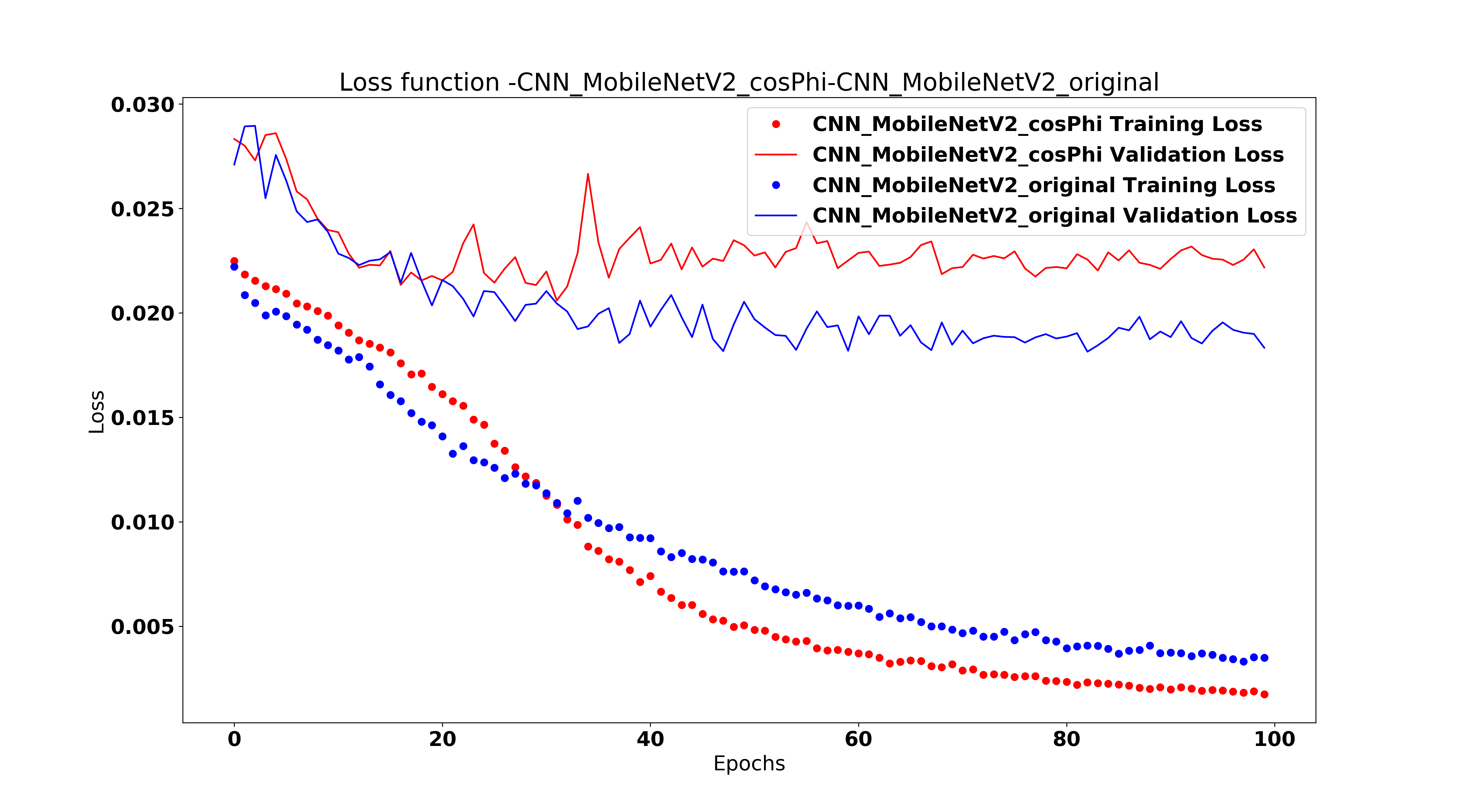}
	\caption{MobileNet V2 training results: Loss  function values for the original image and the FM component.}
	\label{fig:loss_mobile_original}
\end{figure}

In figure \ref{fig:loss_mobile_original}, the loss function values during the training are presented for the original and FM component input images. During validation the loss functions oscillate around 0.021 (original) and 0.024 (FM) (see figure \ref{fig:loss_lenet_original} ).

\begin{figure}	[bt]
	\includegraphics[width=\columnwidth]{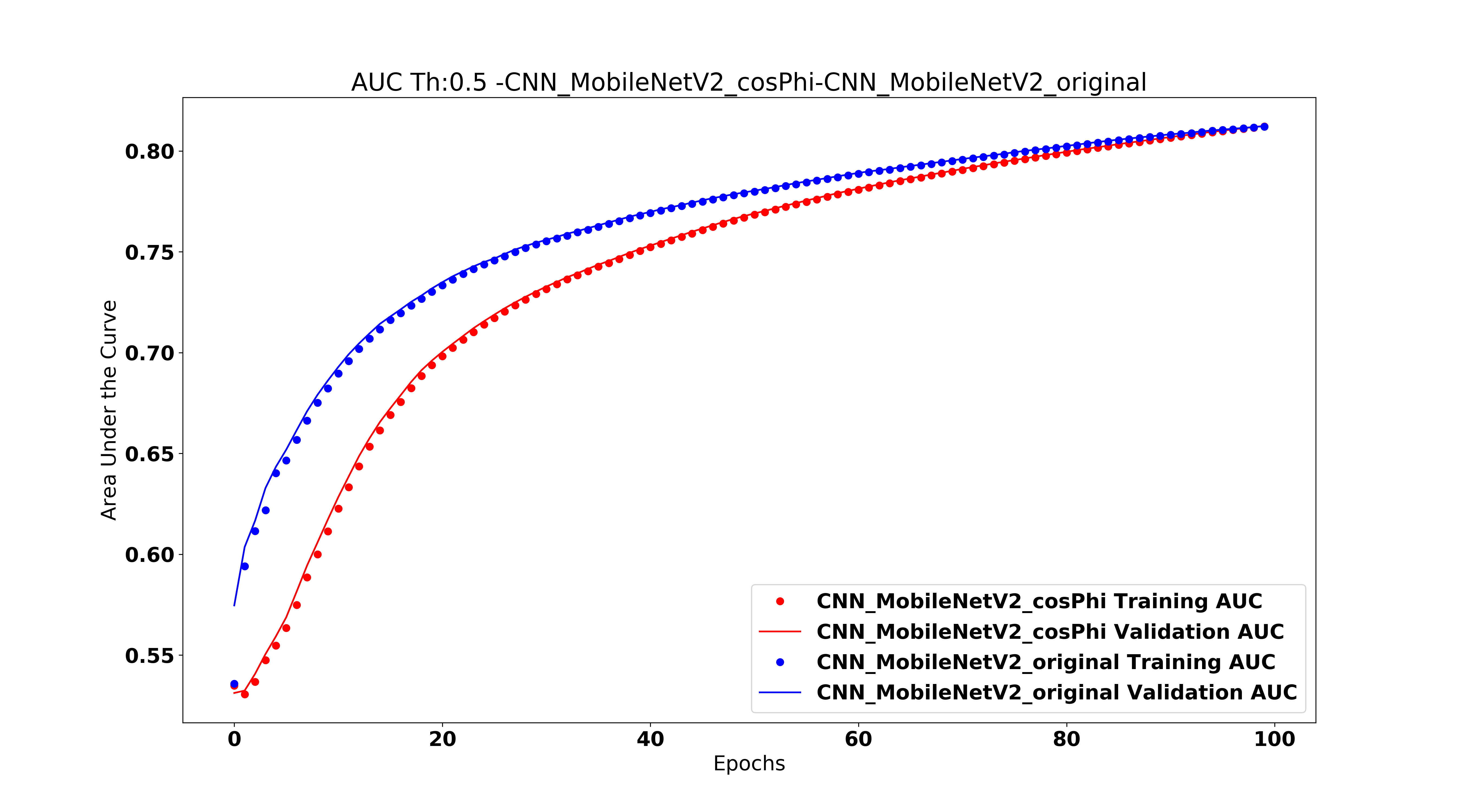}
	\caption{MobileNet V2 training results: AUC at each epoch for FM component (in red) and original image (in blue).}
	\label{fig:auc_mobile_cosPhi}
\end{figure}

Figure \ref{fig:auc_mobile_cosPhi} shows that both the FM and original input will converge to around the same AUC after 80 epochs. Furthermore, the proposed architecture of Single-block regression (see figure \ref{fig:auc_lenet_original}) achieves 0.6 AUC at the 5th epoch. MobileNet V2 achieves a 0.6 AUC at the 8th epoch.

The required training time for MobileNet V2 is significantly more than for the proposed methods (see table \ref{tab:mobileProposedTable}). Here, we noted that MobileNet V2 is considerably faster than other CNN architectures, an specifically targeted towards mobile applications. From table \ref{tab:mobileProposedTable}, it is clear that the proposed architectures are 7x to 11x faster to train. Furthermore, in terms of generalization, the proposed architecture use 123 times less trainable parameters than MobileNet V2.

\begin{table}[H]
	\centering
	
	\begin{center}
		\begin{tabular}{||c c c c||} 
			\hline
			Network      & Portable PC & Desktop PC & Parameters  \\ [0.5ex] 
			\hline\hline
			MobileNet V2 &220 sec          & 68 sec        & 1205073 \\ 
			\hline
			Single-block &20  sec         & 10  sec       & 9775  \\ 
			\hline
			Multi-block  &4   sec         & 2   sec       & 7045   \\ [1ex] 
			\hline			
		\end{tabular}
	\end{center}
	
	\caption{Comparison of MobileNet V2 vs the proposed architectures. The times correspond to a single epoch. For the laptop, we used Intel core i5-7300HQ CPU @ 2.50GHz, 8GB RAM and NVIDIA GeForce GTX 1050. For the desktop system, we used an Intel Xeon CPU ES-2630 v4 @ 2.20GHz, 32 GB RAM and two NVIDIA GeForce GTX 1080. The GPU was used during training. }
	\label{tab:mobileProposedTable}
\end{table}

\begin{figure}	[bt]
	\includegraphics[width=\columnwidth]{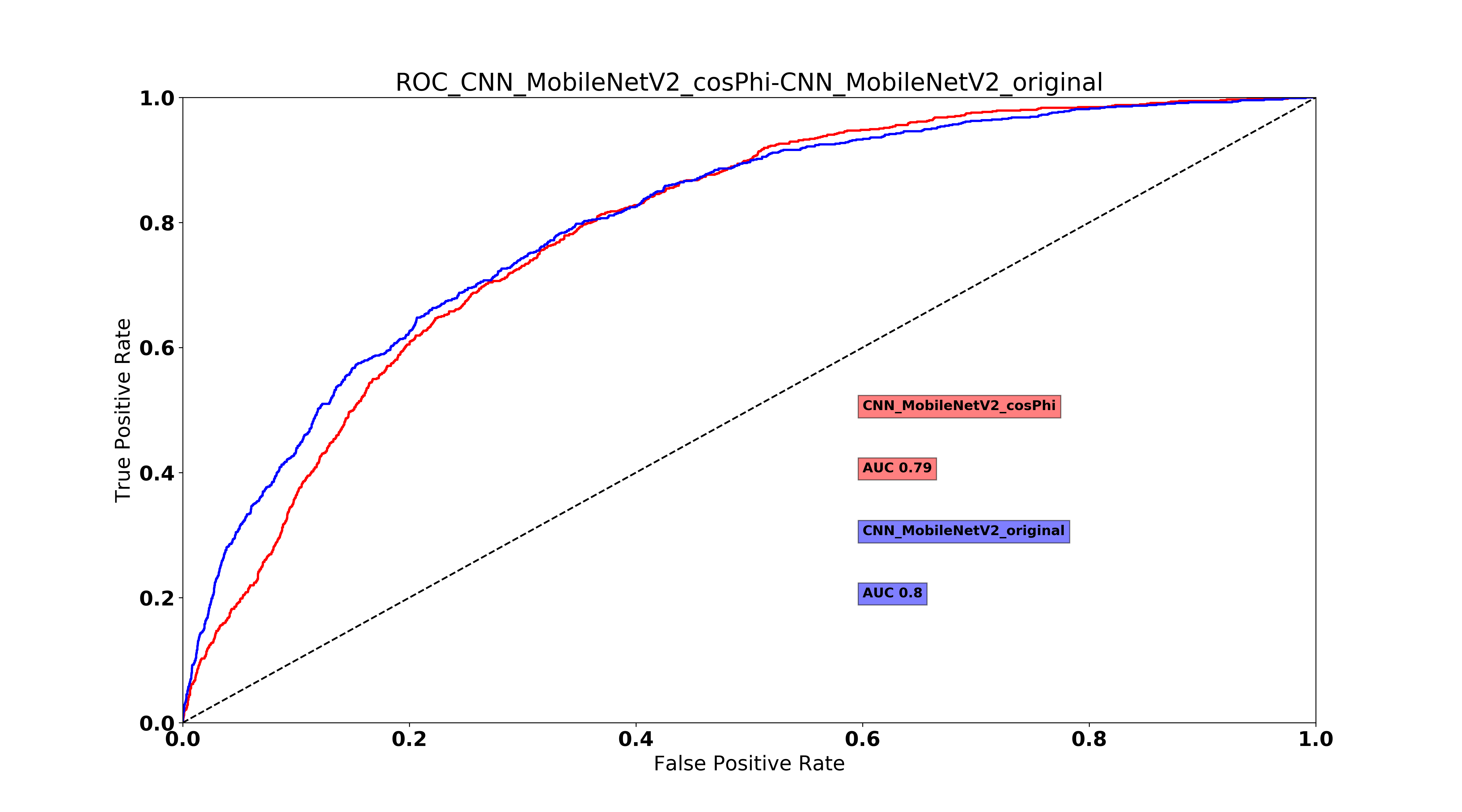}
	\caption{MobileNet V2 regression: ROC curve for FM component (in red) and original image (in blue).}
	\label{fig:roc_mobile_cosPhi}
\end{figure}

The ROC in figure \ref{fig:roc_mobile_cosPhi} demonstrates the similarities between training with the original image and the FM component. The AUC values are also similar.

The considerable increase in complexity when using the MobileNetV2 comes with a predictable improvement for results using any of the three input data types. This certainly helps the original gray scale image to achieves more accurate results.

\begin{figure}	[bt]
	\includegraphics[width=\columnwidth]{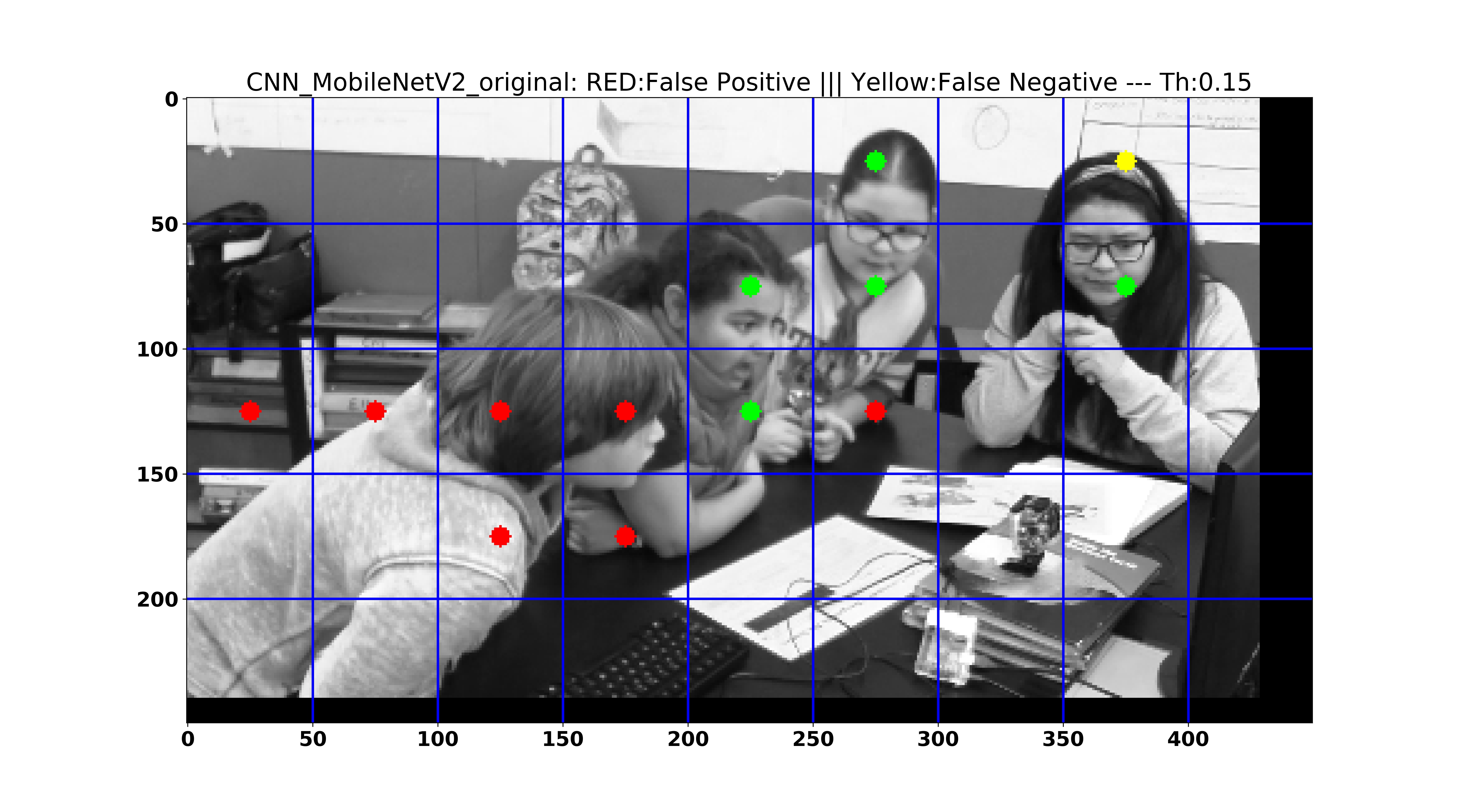}
	\caption{MobileNet V2 regression: Original grayscale frame with marks for TP (green), FP (red) and FN (yellow).}
	\label{fig:frame_mobile_original}
\end{figure}

\begin{figure}	[bt]
	\includegraphics[width=\columnwidth]{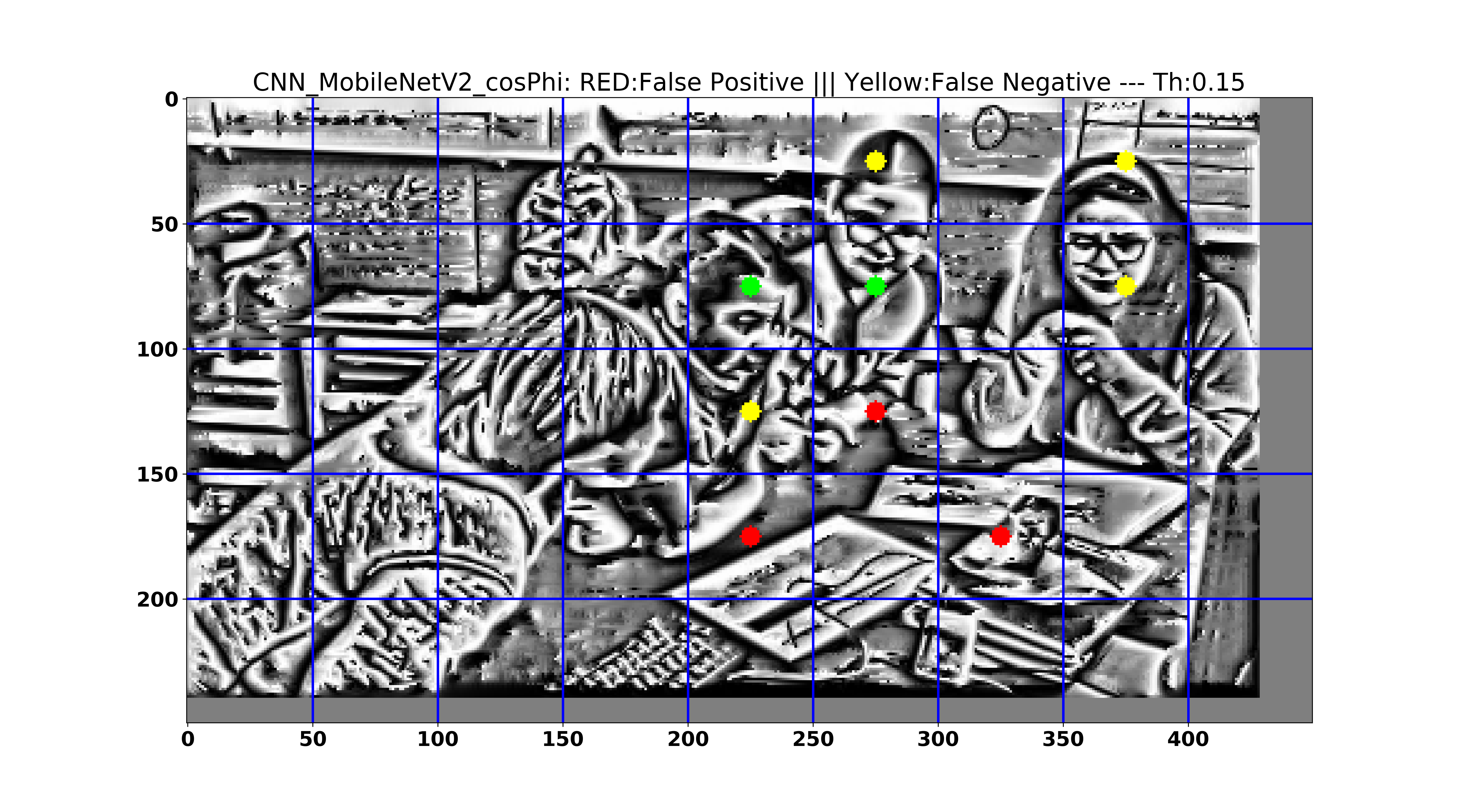}
	\caption{MobileNet V2 regression: FM component frame with marks for TP (green), FP (red) and FN (yellow).}
	\label{fig:frame_mobile_cosPhi}
\end{figure}

Face detection results are presented in figures \ref{fig:frame_mobile_original} and \ref{fig:frame_mobile_cosPhi}. The original image produced more blocks predicted as faces, having more False Positives and True Positives. Predicting every block as a face would decrease the AUC value, in this case there was a balanced in the prediction, corroborated by the AUC. For the FM component, the balance was similar but the number of blocks predicted as faces decreased to a more accurate number.

\chapter{Conclusions and Future Work}

\section{Conclusions}

This thesis has demonstrated the importance of using the dominant Frequency-Modulation components in the training of Convolutional Neural Networks in block-based face detection applications. The thesis has made the following contributions: (i) an approach for designing a Hilbert fixed-point filter using Simulated Annealing, (ii) the design of a low-parameter Gabor filterbank for implementing Dominant Component Analysis, (iii) a hybrid Neural Net using Single-block and Multi-block regression to detect faces in image blocks. 

The video dataset had a variety of student groups working in an unconstrained environment. From the AOLME dataset, 12960 blocks were extracted to train and validate the network, while 6480 blocks from different frames were used for testing. This data was processed by Dominant Component Analysis to obtain the Instantaneous Amplitude (IA) and the Frequency-Modulated (FM) components. The use of IA, FM and original images were also tested against MobileNet V2, that was designed for speed and low computational requirements associated with mobile devices.

The results from this study suggest that using the FM component of an image provides significantly better results and much faster training with low-parameter networks. The LeNet-5 network did not work for the original image or the IA. A vastly simplified version of LeNet-5 with a single convolutional layer produced much better results when used with FM components.

The Single-block regression with the FM image achieved and AUC of 0.78 compared to the AUC of 0.48 when using the original image as input. In comparison, the MobileNet V2 network achieved an AUC=0.8 with the original image input, while requiring 11x more time to train and 123x more parameters to train. The fast training results demonstrate the importance of the Instantaneous Phase. 

\section{Future Work}

Future work include the further exploration of methods that used FM components as input. Future research should consider the replacement of the block architecture by region of interest methods. Furthermore, there is a need to investigate the use of spatial statistics in face detection. Fixed-point implementation on FPGAs can also lead to significant speed-up in training.

\bibliographystyle{unsrt}
\bibliography{references_luis}

\end{document}